\documentclass{article}

\usepackage[final]{neurips_2022}

\usepackage[utf8]{inputenc} %
\usepackage[T1]{fontenc}    %
\usepackage{hyperref}       %
\usepackage{url}            %
\usepackage{booktabs}       %
\usepackage{amsfonts}       %
\usepackage{nicefrac}       %
\usepackage{microtype}      %
\usepackage{xcolor}         %
\usepackage{natbib}
\usepackage{algorithm}
\usepackage{algorithmic}
\usepackage{subfigure}

\usepackage{amsmath}
\usepackage{amssymb}
\usepackage{mathtools}
\usepackage{amsthm}
\usepackage{theoremref}
\usepackage{thmtools}
\usepackage{bbm}
\usepackage{thm-restate}

\usepackage{cleveref}
\crefname{figure}{figure}{figures}
\crefname{equation}{equation}{equations}
\crefname{table}{table}{tables}
\crefname{section}{section}{sections}
\crefname{rproposition}{proposition}{propositions}
\crefname{rlemma}{lemma}{lemmas}

\theoremstyle{plain}

\theoremstyle{definition}

\theoremstyle{remark}

\usepackage[textsize=tiny]{todonotes}

\newcommand{\dataset}{{\cal D}}

\newcommand{\pib}{{\pi_b}}
\newcommand{\btheta}{{\boldsymbol\theta}}

\newcommand{\pitheta}{{\pi_\btheta}}

\newcommand{\pidata}{{\pi_{D}}}

\newcommand{\pieval}{{\pi_e}}
\newcommand{\data}{{{D}}}

\newcommand{\msemath}{{\operatorname{MSE}}}
\newcommand{\loss}{{\mathcal{L}}}
\newcommand{\horizon}{{l}}
\newcommand{\tpieval}{{{\tilde \pi}_e}}
\newcommand{\tpiD}{{{\tilde \pi}_D}}
\newcommand{\RL}{\textsc{rl}}
\newcommand{\ROS}{\textsc{ros}}
\newcommand{\MSE}{\textsc{mse}}
\newcommand{\BPG}{\textsc{bpg}}
\newcommand{\MC}{\textsc{mc}}
\newcommand{\OPD}{\textsc{opd}}
\newcommand{\OIS}{\textsc{ois}}
\newcommand{\OS}{\textsc{os}}
\newcommand{\KL}{\textsc{kl}}
\newcommand{\WRIS}{\textsc{wris}}
\newcommand{\WIS}{\textsc{wis}}
\newcommand{\FQE}{\textsc{fqe}}

\title{Robust On-Policy Sampling for Data-Efficient Policy Evaluation in Reinforcement Learning}

\author{%
  Rujie Zhong$^1$, Duohan Zhang$^{2,\star}$, Lukas Sch\"afer$^1$, Stefano V. Albrecht$^1$, Josiah P. Hanna$^{3,\star}$ \\
  $^1$ School of Informatics, University of Edinburgh \\
  $^2$ Statistics Department, University of Wisconsin -- Madison \\
  $^{3}$ Computer Sciences Department, University of Wisconsin -- Madison \\
  $^\star$ Correspondence to \{dzhang357, jphanna\}@wisc.edu
}

\begin{document}

\maketitle

\begin{abstract}

Reinforcement learning (\RL{}) algorithms are often categorized as either on-policy or off-policy  depending on whether they use data from a target policy of interest or from a different behavior policy.
In this paper, we study a subtle distinction between on-policy \textit{data} and on-policy \textit{sampling} in the context of the \RL{} sub-problem of policy evaluation.
We observe that on-policy sampling may fail to match the expected distribution of on-policy data after observing only a finite number of trajectories and this failure hinders data-efficient policy evaluation.
Towards improved data-efficiency, we show how non-i.i.d., off-policy sampling can produce data that more closely matches the expected on-policy data distribution and consequently increases the accuracy of the Monte Carlo estimator for policy evaluation.
We introduce a method called \textit{Robust On-Policy Sampling} and demonstrate theoretically and empirically that it produces data that converges faster to the expected on-policy distribution compared to on-policy sampling.
Empirically, we show that this faster convergence leads to lower mean squared error policy value estimates.

\end{abstract}

\section{Introduction}

Reinforcement learning (\RL{}) algorithms are often categorized using the dichotomy of on-policy versus off-policy.
On-policy algorithms learn about a particular target policy using data collected by behaving according to the target policy.
Off-policy algorithms use data collected by behaving according to a different \textit{behavior} policy.
We study a subtle distinction between on-policy \textit{data} versus on-policy \textit{sampling} as a step towards more data-efficient \RL{} algorithms.
To better understand this distinction, consider a simple example.
In this example, a certain target policy repeatedly visits a state in which it takes action A with probability $0.2$ and action B with probability $0.8$.
Under on-policy sampling, after five visits to this state, we might actually observe action A 2 times and action B 3 times instead of the expected 1 and 4 times.
Alternatively, we could collect data off-policy by deterministically tracking the expected target policy action proportions; doing so results in observing the exact expected action frequencies.
Though the latter case uses off-policy sampling, it produces data that is arguably more on-policy than the data produced by on-policy sampling. %

In this paper, we study the distinction between on-policy sampling and on-policy data in the context of the \RL{} sub-problem of policy evaluation \citep{Zinkevich2006}.
In policy evaluation, we are given an \textit{evaluation policy} and asked to estimate the expected return that would be accrued when running the evaluation policy on a task of interest.
This problem is important for high confidence deployment of \RL{}-trained policies.
In \RL{} applications, such as robotics, data-efficient policy evaluation is of the utmost importance -- we desire the most accurate estimate with minimal collected data.
While much research has gone into how to most efficiently use a set of already collected data, i.e., the off-policy policy evaluation problem \citep{jiang2016doubly,thomas2016data-efficient}, an implicit assumption in the \RL{} community is that on-policy data is preferred to off-policy data when available.
When data can be collected on-policy, we can use the Monte Carlo estimator which computes a mean return estimate using trajectories sampled i.i.d.\ by running the evaluation policy.
In the limit, with infinite trajectories, the empirical proportion of each trajectory will converge to its true probability under the evaluation policy and the estimate will converge to the true expected return.
However, for any \textit{finite} sample-size, the empirical proportion of each trajectory will likely fail to match the true probability and the estimate will have error.
Such \textit{sampling error} is an inevitable feature of i.i.d.\ sampling.
The probability of each new trajectory is unaffected by the trajectories occurring in the past and thus the only way to ensure the empirical distribution matches the true probability is to sample a large enough data set.
That is, it is only in the limit that on-policy sampling produces exactly on-policy data.

The observations made so far raise the question: ``can non-i.i.d., off-policy trajectory sampling cause the empirical distribution of trajectories to converge to the expected on-policy distribution faster?"
We answer this question affirmatively by introducing a method that adapts the data collecting behavior policy to  consider what data has already been collected when selecting future actions.
We call this method \textit{Robust On-Policy Sampling} (\ROS{})\footnote{We provide an open-source implementation of \ROS{} and all experimental data at \url{https://github.com/uoe-agents/robust_onpolicy_data_collection}.} since the empirical distribution of data it produces converges faster to the expected on-policy trajectory distribution compared to standard on-policy sampling.
We give a theoretical result supporting this claim and then confirm our theory with policy evaluation experiments in finite and continuous-valued state- and action-space domains showing 1) \ROS{} reduces sampling error in finite datasets and 2) consequently lowers the \MSE{} of policy value estimates compared to i.i.d.\ on-policy sampling.

Our paper contributes to the field of \RL{} on two fronts.
On one front, we introduce a practical method for data collection and demonstrate empirically that it leads to more accurate policy evaluation compared to on-policy sampling.
Simultaneously, our work examines nuance in the on-policy versus off-policy dichotomy.
A better understanding of this nuance opens up the possibility of designing new data collection procedures to improve the data efficiency of any \RL{} algorithm that relies upon on-policy data.

\section{Related Work}

Data collection is a fundamental part of the \RL{} problem.
The most widely studied data collection problem is the question of how an agent should explore its environment to learn an optimal policy~\citep{schaefer2022derl,ostrovski2017count,tang2017exploration}.
In contrast to these approaches, our work focuses on the question of how an agent should collect data to \textit{evaluate} a fixed policy.
When given a choice of how to collect data for policy evaluation, on-policy data collection is generally preferable to off-policy data collection \citep{sutton1998reinforcement}.
Notable exceptions are adaptive importance sampling (\textsc{ais}) methods \citep{oosterhuis2020taking,hanna2017data-efficient,ciosek2017offer,bouchard2016online,frank2008reinforcement} and quasi-Monte Carlo methods \citep{arnold2022policy}.
Both these \textsc{ais} methods and the Quasi-Monte Carlo method of \citet{arnold2022policy} lower variance in estimates computed with future samples while our method lowers the total error in the estimate computed from both past and future samples.

In one of our experiments, we consider a setting where we already have some data (collected off-policy) and must decide how to collect additional data for policy evaluation.
This problem has been previously studied in the bandit literature \citep{tucker2022variance-optimal} or when there are only a finite number of policies that could be ran \citep{konyushkova2021active}.
These prior works also show that on-policy data collection is a sub-optimal choice.
They differ from (and are complementary to) our work in that they still use i.i.d. sampling for data collection whereas we show how non-independent sampling can be used to produce data that more closely matches a desired distribution.

The method we introduce in this paper is motivated by the idea of decreasing sampling error in all collected data.
Previous work has considered how sampling error can be reduced \textit{after} data collection by re-weighting the obtained samples.
For example, \citet{hanna2021importance} show how importance sampling with an estimated behavior policy can lower sampling error and lead to more accurate policy evaluation.
Similar methods have also been studied for policy evaluation in multi-armed bandits \citep{narita2019efficient,li2015toward} and temporal-difference learning \citep{pavse2020reducing}.
These prior works assume  data is available a priori and ignore the question of how to collect it when unavailable.

Finally, the idea of adapting the \textit{sampling distribution}, (i.e., behavior policy) has analogs outside of policy evaluation in Markov decision processes.
\citet{ohagan1987monte} identifies flaws with i.i.d.\ sampling for Monte Carlo estimation that motivate taking past samples into account.
\citet{rasmussen2003bayesian} use Gaussian processes to represent uncertainty in an expectation to be evaluated and use this uncertainty to guide future sample generation.
Concurrent to this work, \citet{mukherjee_revar_2022} introduced an uncertainty aware method for data collection in policy evaluation that can be seen as an adaptation of these ideas to the \RL{}.

\section{Preliminaries}

In this section, we introduce notation, formalize the policy evaluation problem, and introduce the Monte Carlo estimator for policy evaluation.

\subsection{Notation}\label{sec:prelims}

We assume the environment is a finite-horizon, episodic \emph{Markov decision process} (\textsc{mdp}) with state set $\mathcal{S}$, action set $\mathcal{A}$, transition function, $P: \mathcal{S} \times \mathcal{A} \times \mathcal{S} \rightarrow [0,1]$,  reward function $R: \mathcal{S} \times \mathcal{A} \rightarrow \mathbb{R}$, discount factor $\gamma$, maximum horizon $l$, and initial state distribution $d_0$ \citep{puterman2014markov}.
We assume that $\mathcal{S}$ and $\mathcal{A}$ are finite though our empirical analysis considers both settings.
We assume that the transition and reward functions are unknown.
A policy, $\pi: \mathcal{S} \times \mathcal{A} \rightarrow [0,1]$, is a function mapping states and actions to probabilities.
We use $\pi(a|s) \coloneqq \pi(s,a)$ to denote the conditional probability of action $a$ given state $s$ and $P(s^\prime | s,a) \coloneqq P(s, a, s^\prime)$ to denote the conditional probability of state $s^\prime$ given state $s$ and action $a$.
Since we assume a finite-horizon, we assume the state definition implicitly includes temporal information \citep{agarwal2022reinforcement}.

Let $h \coloneqq (s_0,a_0,r_0,s_1, \dotsc, s_{\horizon - 1},a_{\horizon - 1},r_{\horizon - 1})$ be a  \textit{trajectory} and $g(h) \coloneqq \sum_{t=0}^{\horizon - 1} \gamma^t r_t$ be the \textit{discounted return} of $h$.
Any policy induces a distribution over trajectories, $\Pr(h | \pi)$.
We define the \textit{value} of a policy, $v(\pi)$, as the expected discounted return when sampling a trajectory by following policy $\pi$: $v(\pi) \coloneqq \mathbf{E}[g(H) | H \sim \pi] = \sum_{h} \Pr(h|\pi) g(h)$ where $H$ is a random variable representing a trajectory and $H \sim \pi$ denotes sampling $H$ by running $\pi$ in the given environment.

\subsection{Policy Evaluation}

In the policy evaluation problem, we are given an \textit{evaluation policy}, $\pieval$, for which we would like to estimate $v(\pieval)$.
Conceptually, algorithms for policy evaluation involve two steps: collecting data (or receiving previously collected data) and computing an estimate from that data.
We assume that data is collected by running a policy which we call the \textit{behavior policy}.
If the behavior policy is the same as the evaluation policy data collection is \textit{on-policy}; otherwise it is \textit{off-policy}.
Whether on-policy or off-policy, we assume the data collection process produces a set of trajectories, $D \coloneqq \{H_i\}_{i=1}^n$ and write $D \sim \pi$ to denote collecting these trajectories by running policy $\pi$.
The final value estimate is then computed by a policy evaluation estimator (\textsc{pe}) that maps the set of trajectories to a scalar-valued estimate of $v(\pieval)$.
Following earlier work in policy evaluation (e.g., \citep{thomas2016data-efficient,jiang2016doubly}), we set our goal to be policy evaluation with low \textit{mean squared error} (\MSE{}):
\begin{equation}
    \msemath \biggl [\operatorname{PE} \biggr] \coloneqq \mathbf{E}\biggl [\biggl(\operatorname{PE}(D) - v(\pi_e)\biggr)^2 \biggm | D \sim \pib \biggr],
\end{equation}
where $\pib$ is the behavior policy that is run to collect $D$ and $\operatorname{PE}$ is a generic policy evaluation estimator.

\subsection{Monte Carlo Policy Evaluation}

Perhaps the most fundamental, model-free policy evaluation method is the \textit{Monte-Carlo} (\MC{}) estimator.
Given a data set, $D$, of $n$ trajectories, the Monte Carlo estimate, $\operatorname{MC}(D)$, is the mean return over $D$:
\begin{equation}
    \operatorname{MC}(D) \coloneqq
\frac{1}{n}\sum_{i=1}^n g(H_i)
= \sum_{h} \Pr(h|D) g(h),
\end{equation}
where $\Pr(h|D)$ denotes the empirical probability of $h$, i.e.\ how often $h$ appears in $D$.

If trajectories in $D$ are collected i.i.d.\ by running $\pieval$ (i.e., \textit{on-policy} sampling), the Monte Carlo estimator is unbiased and consistent assuming $g(h)$ is bounded \citep{sen1993large}.
However, this method can have high variance as on-policy sampling may require many trajectories for the empirical trajectory distribution $\Pr(h|D)$ to accurately approximate $\Pr(h|\pieval)$.
Since on-policy sampling collects each trajectory i.i.d., it relies on the law of large numbers for an accurate weighting on each possible return.
We call error between $\Pr(h|D)$ and $\Pr(h|\pieval)$ \textit{sampling error}.

\section{Data-Conditioned Monte Carlo Estimates}

\looseness=-1
In this section, we motivate how an estimator that uses on-policy data can benefit from off-policy sampling.
Specifically, we consider the Monte Carlo estimator and suppose that we have already collected a data set, $\mathcal{D}_1$, of trajectories.
We now wish to collect an additional set of trajectories, $D_2$, and compute the Monte Carlo estimate with the set $\dataset_1 \cup D_2$.
Note that $\dataset_1$ is a fixed set (the trajectories already observed) while $D_2$ is a random variable (the trajectories yet to be observed).
How should $D_2$ be collected for minimal \MSE{} policy evaluation with the Monte Carlo estimator?
Our analysis in this section suggests that i.i.d.\ sampling of trajectories with $\pieval$ may be a sub-optimal choice.

In this setting, the Monte Carlo estimator using $\dataset_1 \cup \data_2$ can be written as:
\begin{equation}\label{eq:dc-estimator}
    \operatorname{MC}(\dataset_1 \cup D_2) \coloneqq \underbrace{\frac{1}{n} \sum_{i=1}^{n_{\dataset_1}} g(h_i)}_\text{fixed value} + \underbrace{\frac{1}{n} \sum_{i=1}^{n_{D_2}} g(H_i)}_\text{random variable},
\end{equation}
where $n_{\dataset_1}$ and $n_{D_2}$ are the number of trajectories in $\dataset_1$ and $\data_2$, respectively and $n = n_{\dataset_1} + n_{\data_2}$.
We refer to \eqref{eq:dc-estimator} as the \textit{data-conditioned Monte Carlo estimator}.

Viewing the Monte Carlo estimator as a sum between a fixed quantity and a random quantity changes how we view the statistical properties of the estimator.
For instance, while the Monte Carlo estimator is known to be unbiased under on-policy sampling, its data-conditioned estimate is biased as shown in the following proposition.
\begin{restatable}[]{rproposition}{propositionbias}
The data conditioned Monte Carlo estimator is biased under on-policy sampling of $D_2$ unless $\operatorname{MC}(\dataset_1) = v(\pieval)$ or $\dataset_1 = \emptyset$. That is:
\[
\mathbf{E} \biggl[\operatorname{MC}(\dataset_1 \cup D_2) \biggm | D_2 \sim \pieval \biggr] \neq v(\pieval).
\]\label{proposition:bias}
\end{restatable}
\begin{proof}
See Appendix \ref{app:prop-bias}.
\end{proof}
\begin{restatable}[]{rremark}{remarkbias}
\Cref{proposition:bias} holds even if $\dataset_1$ was collected under on-policy sampling as well.
When $\dataset_1$ was collected under on-policy sampling then the Monte Carlo estimator is unbiased \textit{considering all possible realizations of $\dataset_1$}.
However, once the trajectories in $\dataset_1$ are fixed, it no longer matters what others values they could have taken.
\end{restatable}

Can we reduce the bias of the data-conditioned Monte Carlo estimator by collecting $D_2$ with a policy that is different than $\pieval$?
We conclude this section with an example showing that we can.
Consider a one-step \textsc{mdp} with one state, $s$, and two actions, $a_0$ and $a_1$. The return following $a_0$ is $2$ and the return following $a_1$ is $4$. The evaluation policy is $\pieval(a_0|s) = \pieval(a_1|s) = 0.5$. Suppose that, after sampling 3 trajectories, $\dataset_1$ contains two of $\{s, a_0,2\}$ and one occurrence of $\{s, a_1,4\}$.
Note that action $a_0$ is over-sampled relative to its true probability in $s$ and $a_1$ is under-sampled.
If we collect an additional trajectory with $\pieval$ the expected value of the Monte Carlo estimate is: $\frac{1}{4} (2 + 2 + 4 + 2\pieval(a_0) + 4\pieval(a_1)) = \frac{11}{4} = 2.75$.
The true value, $v(\pieval) = 3$ and thus, \textit{conditioned on prior data}, the Monte Carlo estimate is biased in expectation as shown in \Cref{proposition:bias}.
If instead we choose the behavior policy such that $\pib(a_1)=1$ then neither action is over- or under-sampled and the expected value of the Monte Carlo estimate is the exact true value: $\frac{1}{4} (2 + 2 + 4 + 4) = \frac{12}{4} = 3$.

This example highlights that adapting the behavior policy to consider previously collected data can lower the expected finite-sample error of policy evaluation.
In the next section, we introduce an adaptive data collection method that adjusts the behavior policy based on what data has already been observed so as to lower the \MSE{} of a Monte Carlo estimate using all observed data.

\section{Robust On-Policy Data Collection}

In this section, we introduce a method that adapts the data-collecting behavior policy online to minimize sampling error in the data used by the Monte Carlo estimator.
Specifically, let $\dataset_t$ denote all trajectories observed  up to time-step $t$ of the current trajectory (including the partial current trajectory).
At time-step $t$, our method sets the behavior policy so as to reduce the current sampling error, i.e., divergence between $\Pr(h | \pieval)$ and $\Pr(h | \dataset_t)$.
Our method can be run starting with $\dataset_t=\emptyset$ or already containing trajectories in a setting like that described in the preceding section.

\looseness=-1
To reduce sampling error when collecting future trajectories, we want to adjust the behavior policy to increase the probability of  under-sampled trajectories, i.e., $h$ for which $\Pr(h|\dataset_t) < \Pr(h | \pieval)$.
Unfortunately, the trajectory distributions are unknown because the transition function, $P$, is also unknown.
Instead, we will increase the probability of under-sampled actions.
Let $\pidata: \mathcal{S} \times \mathcal{A} \rightarrow [0,1]$ denote the \textit{empirical policy} which gives the proportion of times that each action was taken in each state in $\dataset_t$.
If $\pidata(a|s) < \pieval(a|s)$, then $a$ has appeared less often in the data than it would in expectation under $\pieval$. Thus, we should increase the probability of $a$ in $s$ for future data collection.

\looseness=-1
When the state and action spaces are finite, $\pidata$ can be computed exactly as the maximum likelihood policy under $\dataset_t$: 
\begin{align}
    \pidata \coloneqq \arg\max_\pi \loss (\pi), &&& \loss(\pi) \coloneqq \sum_{h \in \dataset_t} \sum_{t'=0}^{l-1}\log \pi (a_{t'} | s_{t'}),
\end{align}
where the argmax is taken with respect to all policies.
In larger \textsc{mdp}s, we require function approximation which may make $\pidata$ hard to compute and update online as new data is collected.
Fortunately, with an additional assumption we can determine the direction to adjust action probabilities without explicitly computing $\pidata$.
This assumption is that $\pieval$ belongs to a class of differentiable, parameterized policies and is parameterized by vector $\btheta \in \mathbb{R}^d$.
This assumption is mild for many \RL{} applications as it permits tabular, linear, and neural network policy representations.
We use $\btheta_e$ to represent the parameter values for $\pieval$.
We show in the next subsection that the gradient of the log-likelihood at $\btheta_e$, $\nabla_\btheta \loss(\pitheta) \vert_{\btheta = \btheta_e}$, can be used to make sampling-error-reducing changes to the behavior policy.

\subsection{Robust On-Policy Sampling}

Our primary algorithmic contribution -- \textbf{R}obust \textbf{O}n-Policy \textbf{S}ampling (\ROS{}) -- reduces sampling error by adapting the behavior policy with a single step of gradient descent on the log-likelihood at each time-step.
From here on, we use $\nabla_\btheta \loss$ to denote the gradient of the log-likelihood evaluated at $\btheta_e$.
Observe that $\nabla_\btheta \loss$ provides a direction to adjust $\btheta_e$ to \textit{increase} the probability of actions that were over-sampled relative to their probability under $\pieval$.
Thus, $-\nabla_\btheta \loss$ provides a direction to adjust $\btheta_e$ to \textit{decrease} the probability of over-sampled actions for which $\pidata(a|s) > \pieval(a|s)$.
With this insight, \ROS{} is able to adapt $\btheta_e$ so that $\pidata$ tracks $\pieval$ without ever computing $\pidata$.
At each time-step, \ROS{} computes $\nabla_\btheta \loss$ with all state-action pairs previously observed and then changes the evaluation policy parameters with a single step of gradient descent so that under-sampled actions have greater probability than they would have under $\pieval$.

Pseudocode for \ROS{} is given in Algorithm \ref{alg:ros2}.
\ROS{} first computes $\nabla_\btheta \loss$ with previously collected trajectories if any are provided (Line 4).
\ROS{} then collects $n$ additional trajectories by interacting with the given \textsc{mdp} (Lines 6-14).
For each action selection, \ROS{} sets the behavior policy parameters as $\btheta_e - \alpha \nabla_\btheta \loss(\pitheta) \vert_{\btheta = \btheta_e}$ (Lines 9 and 10).
It then computes $\nabla_\btheta \log \pitheta (A | s) \vert_{\btheta = \btheta_e}$ and updates $\nabla_\btheta \loss(\pitheta) \vert_{\btheta = \btheta_e}$ (Lines 11 and 12).
Finally, the chosen action is executed in the environment, a reward received, and the agent moves to the next state (Line 13).
Importantly, note that updating  $\nabla_\btheta \loss$ requires per-timestep computation that is linear in the number of policy parameters and remains constant as the size of $\dataset$ grows.

\begin{algorithm}[tb]
    \begin{algorithmic}[1]
        \STATE {\bfseries Input:} Evaluation policy $\pieval$ with parameters $\btheta_e$, step size $\alpha$, previously collected trajectories to be used for policy evaluation, $\dataset_1$ (possibly empty), number of trajectories to collect, $n$.
        \STATE \textbf{Output:} Data set of trajectories.
        \STATE $k \leftarrow $ number of state-action tuples in $\dataset_1$
        \STATE $\nabla_\btheta \loss \leftarrow \frac{1}{k} \sum_{(s,a) \in \dataset_1} \nabla_\btheta \log \pitheta(a|s) \vert_{\btheta=\btheta_e}$
        \STATE $\mathcal{D} \leftarrow \dataset_1$
        \FOR{$0 \leq i < n$}
            \STATE $s_0 \sim d_0$
            \FOR{$0 \leq t < \horizon$}
                \STATE $\btheta_b \leftarrow \btheta_e - \alpha \nabla_\btheta \loss$
                \STATE $a_t \leftarrow A \sim \pi_{\btheta_b}(\cdot|s_t)$
                \STATE $\nabla_\btheta \loss \leftarrow \frac{k}{k+1}\nabla_\btheta \loss + \frac{1}{k+1}\nabla_\btheta \log\pitheta(a_t|s_t) \vert_{\btheta=\btheta_e}$
                \STATE $k \leftarrow k + 1$
                \STATE $s_{t+1} \sim P(\cdot | s_t, a_t)$, $r_t \leftarrow R(s_t, a_t)$
            \ENDFOR
            \STATE $\dataset \leftarrow \dataset \cup \{(s_0,a_0,r_0,...,s_{\horizon-1},a_{\horizon-1}, r_{\horizon-1})\}$
        \ENDFOR
        \STATE \textbf{Return} $\mathcal{D}$
    \end{algorithmic}
    \caption{Robust On-Policy Sampling.}
    \label{alg:ros2}
    
\end{algorithm}

\subsection{ROS Convergence}\label{sec:theory}

This section develops our theoretical understanding of \ROS{}.
Due to space constraints, we defer all proofs to Appendix \ref{app:theory}.
First, we show that \ROS{} converges to the expected state visitation frequencies under $\pieval$.
Second, we show that, for a fixed state, $\pi_D(\cdot |s)$ converges to $\pieval(\cdot | s)$ faster under \ROS{} compared to on-policy sampling.
Finally, we introduce an upper bound on the squared error between the Monte Carlo estimate and $v(\pieval)$ in terms of sampling error which shows how \ROS's faster convergence affects the MSE of policy evaluation.
These results use the following assumption:
\begin{restatable}[]{rassumption}{assumptiondag}
The discrete state-space of the \textsc{mdp} has a directed acyclic graph (\textsc{dag}) structure. Specifically, states in $\mathcal{S}$ can be partitioned into $\horizon$ disjoint sets $\mathcal{S}_t$ indexed by episode step. The transition function is such that $P(s' | s,a) > 0$ implies that $s \in \mathcal{S}_t$ and $s' \in \mathcal{S}_{t+1}$.\label{ass:dag}
\end{restatable}
Note that Assumption \ref{ass:dag} is mild as any finite-horizon \textsc{mdp} can be made a \textsc{dag} by including the current time-step as part of the state (as we have already assumed in Section \ref{sec:prelims}).
\begin{restatable}[]{rassumption}{assumptionparams}
    \ROS{} uses a step-size of $\alpha \rightarrow \infty$ and
    the behavior policy is parameterized as a softmax function, i.e.,
    $
        \pitheta(a|s) \propto e^{\theta_{s,a}}, %
    $ where for each state, $s$, and action, $a$, we have a parameter $\theta_{s,a}$. As we formally show in Appendix \ref{app:theory}, this assumption implies that \ROS{} always takes the most under-sampled action in each state.\label{ass:step}
    \label{ass:param}
\end{restatable}
We also introduce the notation of $d^t_{\pi}(s)$ as the probability of visiting state $s$ at episode time $t$ while following policy $\pi$ and $d^t_n(s)$ as the empirical frequency of visitations to state $s$ at episode time $t$ after observing $n$ trajectories.

\begin{restatable}[]{rtheorem}{theoremmdpgeneral}
Under Assumptions \ref{ass:dag} and \ref{ass:step} and \ROS{} action selection, $d^t_n(s)$ converges to $d^t_\pi(s)$ with probability 1 for all $s \in \mathcal{S}$ and $0 < t < \horizon$:
\[
 \lim_{n\rightarrow\infty} d^t_n(s) = d^t_\pi(s), \ \forall s \in \mathcal{S}, \ 0 \leq t < \horizon.
\]\label{theorem:converge}
\end{restatable}
\begin{restatable}[]{rtheorem}{theoremfaster}
Let $s$ be a particular state that is visited $m$ times during data collection and assume that $|\mathcal{A}| \geq 2.$
Under Assumption \ref{ass:step}, $D_\mathtt{KL}(\pi_D(\cdot | s) || \pi(\cdot | s)) = O_p(\frac{1}{m^2})$ under \ROS{} sampling while $D_\mathtt{KL}(\pi_D(\cdot | s) || \pi(\cdot | s)) = O_p(\frac{1}{m})$ under on-policy sampling, where $O_p$ denotes stochastic boundedness. 
\label{theorem:rate}
\end{restatable}

\begin{restatable}[]{rtheorem}{theoremerror}
Assume $\forall s \in \mathcal{S}, a \in \mathcal{A}$ that $R(s,a) \leq R_\mathtt{max}$.
The squared error in the Monte Carlo estimate using $\dataset$ can be upper-bounded by:
\[
\left(v(\pieval) - \operatorname{MC}(\dataset) \right)^2 \leq \sum_{t=0}^{\horizon-1} \gamma^{2t} R_\mathtt{max}^2 \sqrt{2D_\mathtt{KL}(d_n^t||d_\pieval^t) + 2\mathbf{E}_{S \sim d_n^t} [D_\mathtt{KL}(\pi_D(\cdot|S)||\pieval(\cdot|S)]}.
\]
\label{theorem:error}
\end{restatable}

\begin{restatable}[]{rremark}{remarkerror}
The second term in the bound in Theorem \ref{theorem:error} is the KL-divergence between $\pidata$ and $\pieval$ which Theorem \ref{theorem:rate} tells us will decrease faster under \ROS{} action selection. The first term is the KL-divergence between the empirical and true state distributions which depends both on sampling error in action selection as well as sampling error in the transition and initial state distributions. While the former decreases faster under \ROS{}, the latter will decrease the same for both \ROS{} and on-policy sampling. Hence, the theoretical faster rate of \ROS{} for reducing sampling error in action selection may be muted by high environment stochasticity. The experimental results given in Figure \ref{fig:imp_env_gw} complement this theoretical observation.
\end{restatable}

\section{Empirical Study}\label{sec:empirical}

We next conduct an empirical study of \ROS{} in policy evaluation problems. %
Our primary goal is to answer the following questions:
\begin{enumerate}%
    \setlength\itemsep{0.01em}
    \item Does \ROS{} reduce sampling error compared to on-policy sampling?
    \item Does \ROS{} lower policy evaluation \MSE{} when starting with and without off-policy data?
\end{enumerate}

We conduct policy evaluation experiments in four domains covering discrete and continuous state and action spaces: a multi-armed bandit problem~\citep{sutton1998reinforcement}, Gridworld~\citep{thomas2016data-efficient}, CartPole, and Continuous CartPole~\citep{gym}.
Since these domains are widely used, we defer their descriptions to Appendix~\ref{sec:env}.
Our primary baseline for comparison is on-policy sampling (\OS{}) of i.i.d.\ trajectories with the Monte Carlo estimator used to compute the final policy value estimate (denoted \textbf{\OS{}-\MC{}}).
We also compare to \BPG{} which finds a minimum variance behavior policy for the ordinary importance sampling (\OIS{}) policy value estimator~\citep{hanna2017data-efficient} (denoted \textbf{\BPG-\OIS{}}).
We provide full experimental details concerning how $\pieval$ and $v(\pieval)$ were determined in Appendix \ref{app:pi_e}.

\subsection{Policy Evaluation without Initial Data}

\begin{figure}
\centering
\subfigure[Without initial data]{\includegraphics[width=0.33\textwidth]{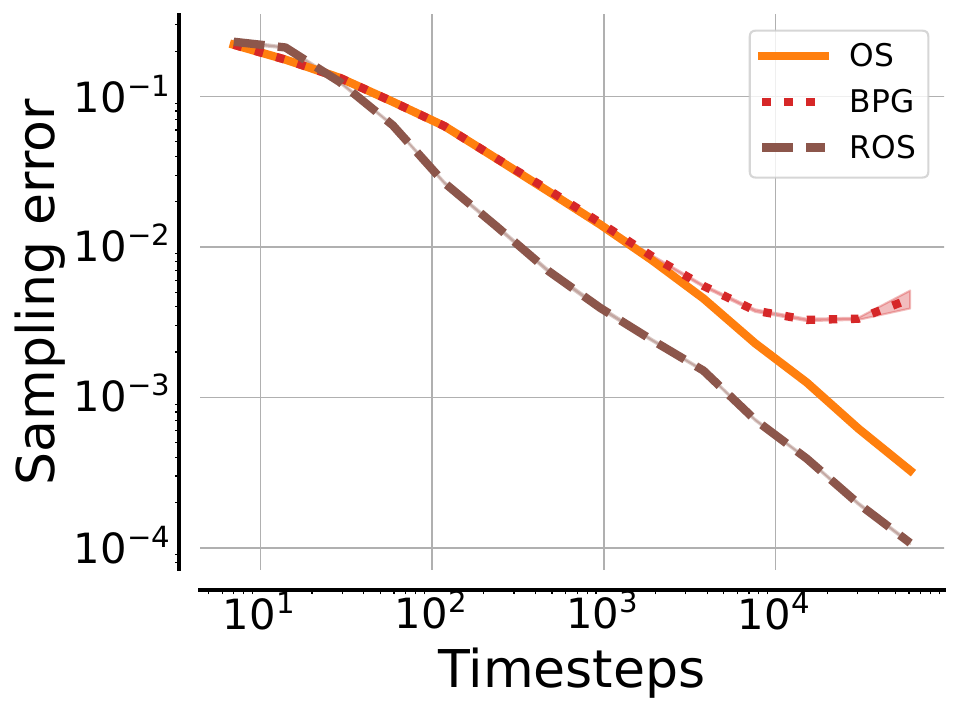}\label{fig:_kl_basic}}
\hspace{3em}
\subfigure[With initial data]{\includegraphics[width=0.33\textwidth]{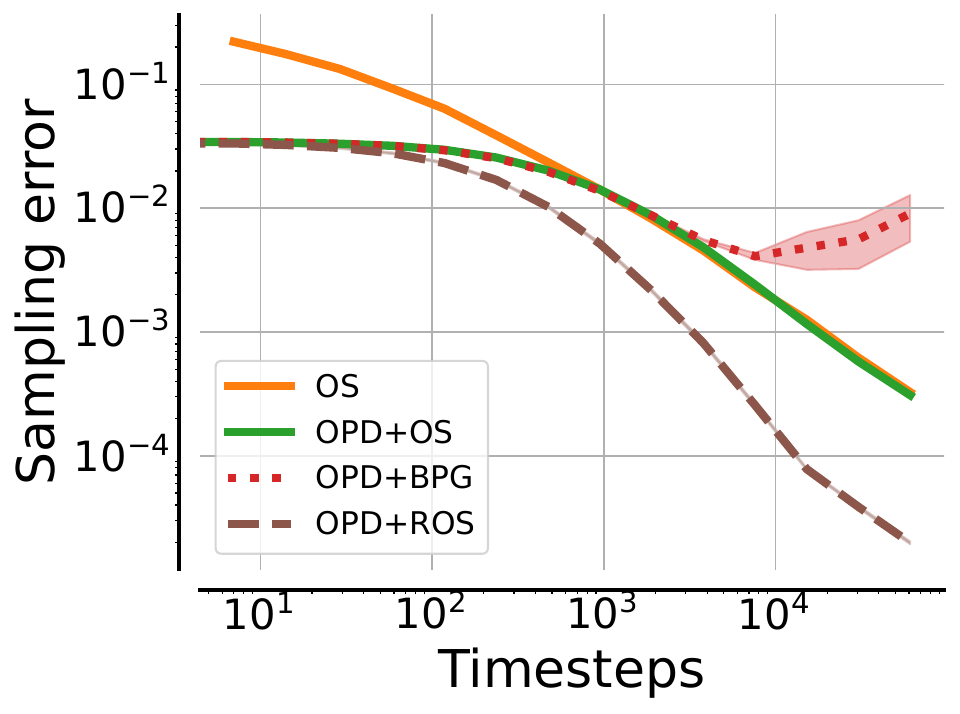}\label{fig:_kl_comb}}
\caption{Sampling error (\KL{}) curves of data collection in the GridWorld domain. Each strategy is followed to collect data with $2^{13}\overline{T}$ steps, and all results are averaged over $200$ trials with shading indicating one standard error intervals. Figures \ref{fig:_kl_basic} and \ref{fig:_kl_comb} show the sampling error curves of data collection \textbf{without} and \textbf{with initial data}, respectively. Axes in these figures are log-scaled.
}
\label{fig:_kl}
\end{figure}

We first run experiments in a setting \textbf{without initial data}, in which all  data is collected from scratch.
Letting $\overline{T}$ denote the average length of a trajectory, in each domain, we collect a total of $2^{13}\overline{T}$ environment steps with each method and compute metrics every $2^1, 2^2, ..., 2^{13}$ trajectories.
Note that we specify the number of environment steps rather than number of trajectories in our empirical results. 
For Bandit $\overline{T}=1$, for GridWorld $\overline{T}=7.43$, CartPole $\overline{T}=48.48$, and for CartPoleContinuous $\overline{T}=49.56$. 
The hyper-parameter settings for all experiments are presented in Appendix~\ref{app:hparams}.

We first verify that \ROS{} reduces sampling error compared to on-policy sampling.
We measure sampling error with the \KL{}-divergence (\KL{}) between $\pieval$ and a parametric maximum likelihood estimate of $\pidata$ from the observe data.
In Appendix \ref{app:measure-sampling-error}, we give a complete definition of the measure as well as an alternative measure that leads to qualitatively similar results.
Due to space constraints, we only show this result for the GridWorld domain (Figure~\ref{fig:_kl_basic}); results for other domains are qualitatively the same and can be found in Appendix~\ref{sec:kl}.
Figure~\ref{fig:_kl_basic} shows that with \ROS{} sampling error decreases faster than \OS{}.
Unsurprisingly, \BPG{} increases sampling error as it is an off-policy method which adapts the behavior policy away from $\pieval$.
These results answer our first empirical question and confirm our theoretical claim that non-i.i.d.\ off-policy sampling can cause the empirical distribution of data to converge to the expected on-policy distribution faster.

\begin{figure}
\centering

\subfigure[Bandit]{\includegraphics[height=7.3em]{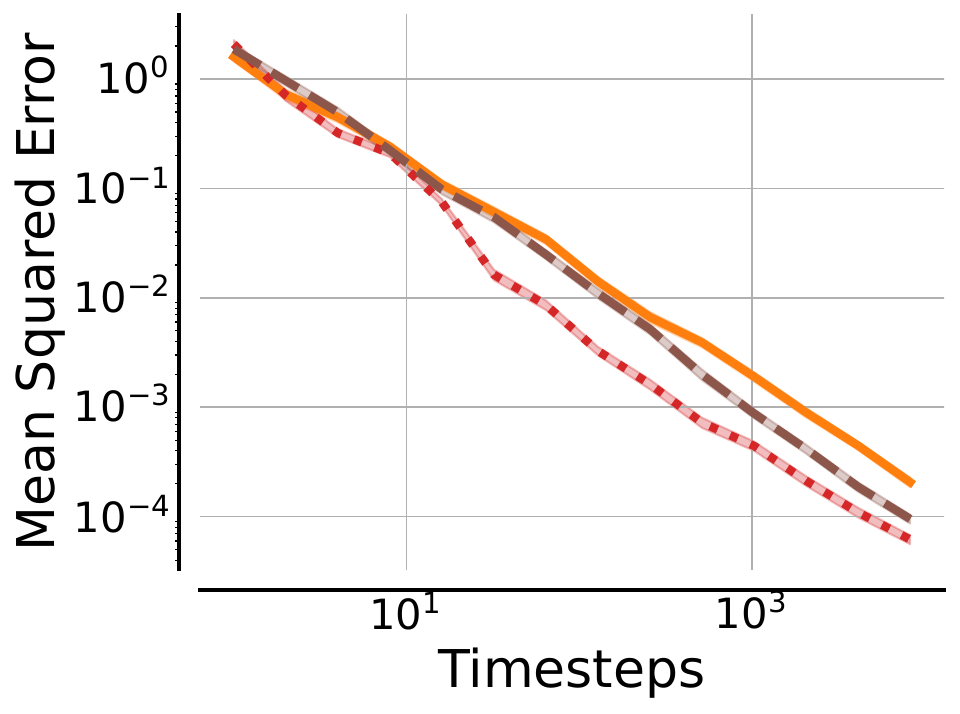}}
\subfigure[GridWorld]{\includegraphics[height=7.3em]{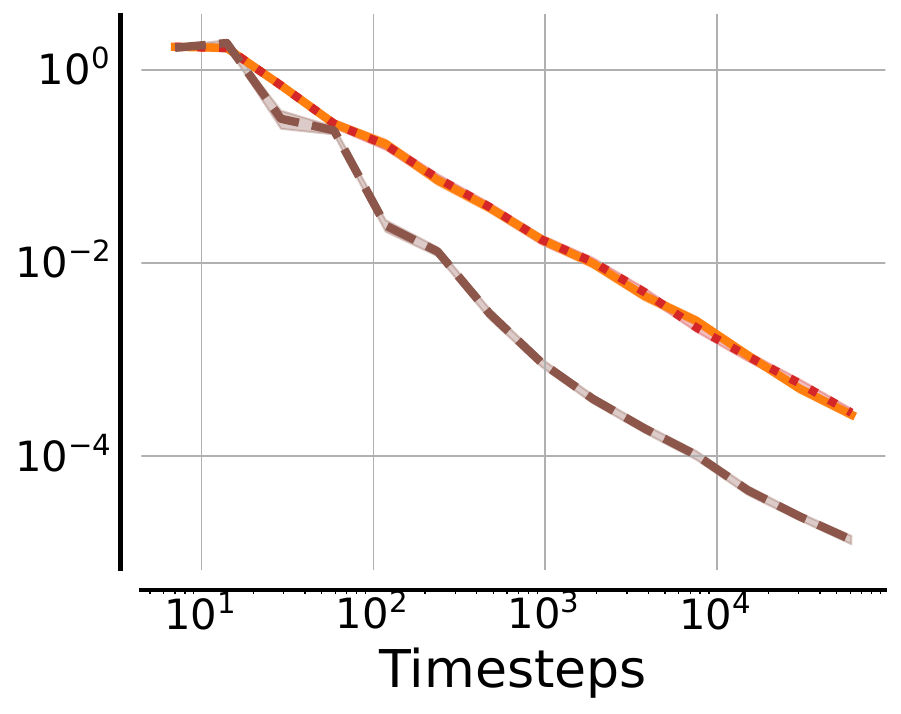}}
\subfigure[CartPole]{\includegraphics[height=7.3em]{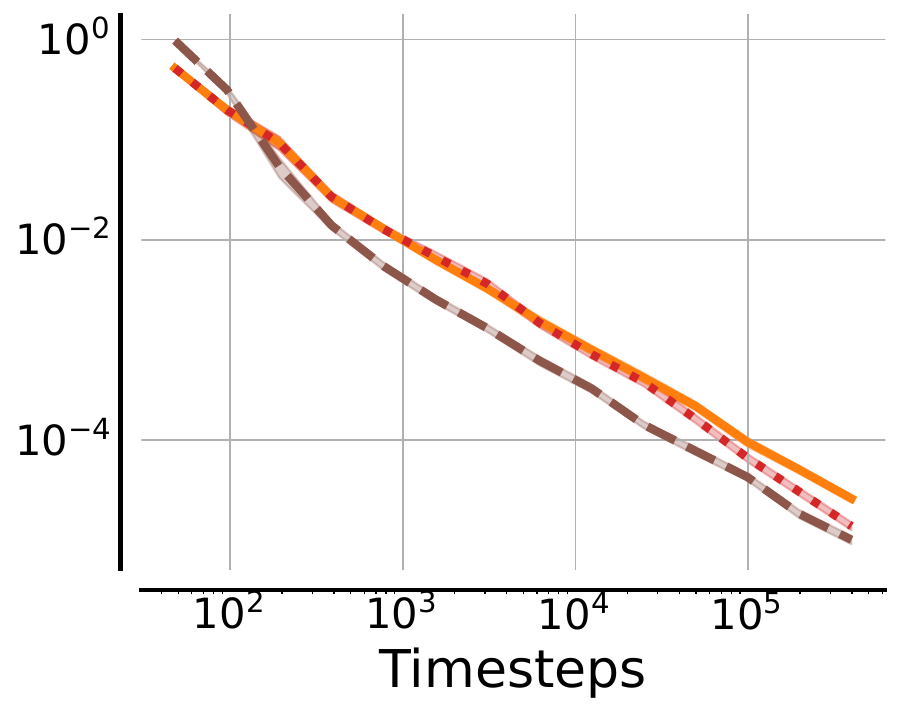}}
\subfigure[CartPoleContinuous]{\includegraphics[height=7.3em]{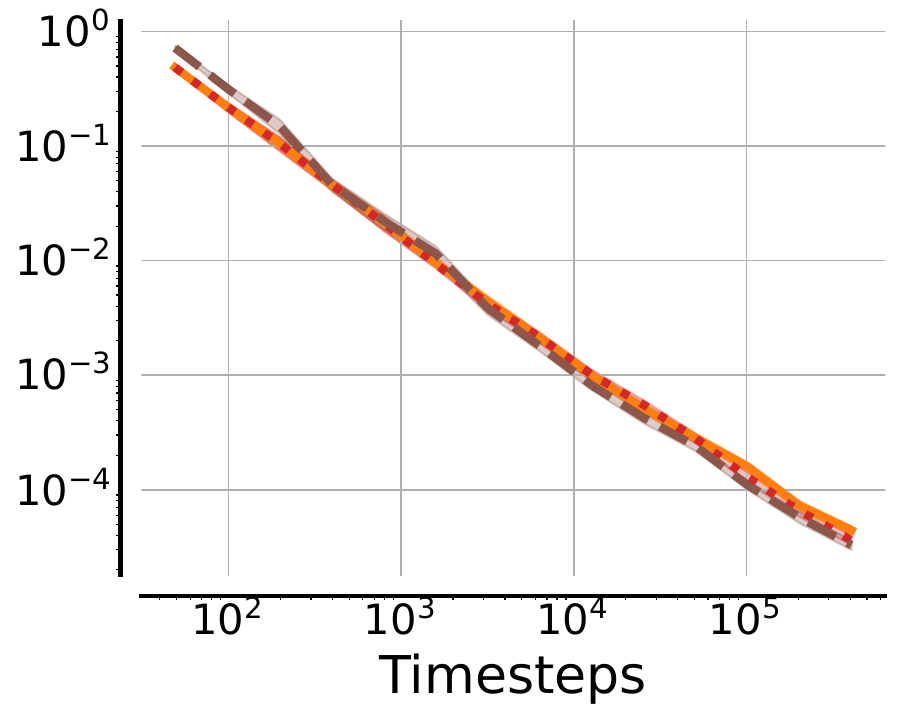}}
\subfigure{\includegraphics[width=0.31\textwidth]{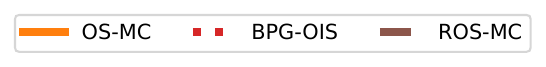}}

\vspace{-5pt}
\caption{Mean squared error (\MSE{}) of policy evaluation in the \textbf{without initial data} setting. Policy evaluation is conducted on the data collected from each strategy, and these curves show the \MSE{} of the estimates (lower is better). %
The vertical axis gives \MSE{} and the horizontal axis is the amount of environment steps taken (both are log-scaled). Shading indicates one standard error.
}
\label{fig:mse}
\end{figure}

Ultimately, this paper focuses on reducing sampling error for lower \MSE{} policy evaluation.
Figure~\ref{fig:mse} shows that \ROS{} lowers \MSE{} compared to both \OS{} and \BPG{} across all domains.\footnote{Numeric values for the final \MSE{} of each method can be found in Appendix \ref{app:stat}. We also report median and interquartile ranges of the error of each method in Appendix \ref{app:median}.}
These results address our second empirical question and support the claim that reducing sampling error decreases the \MSE{} of the Monte Carlo estimator for policy evaluation.

\subsection{Policy Evaluation with Initial Data}

\begin{figure}
\centering
\subfigure[Bandit]{\includegraphics[height=7.3em]{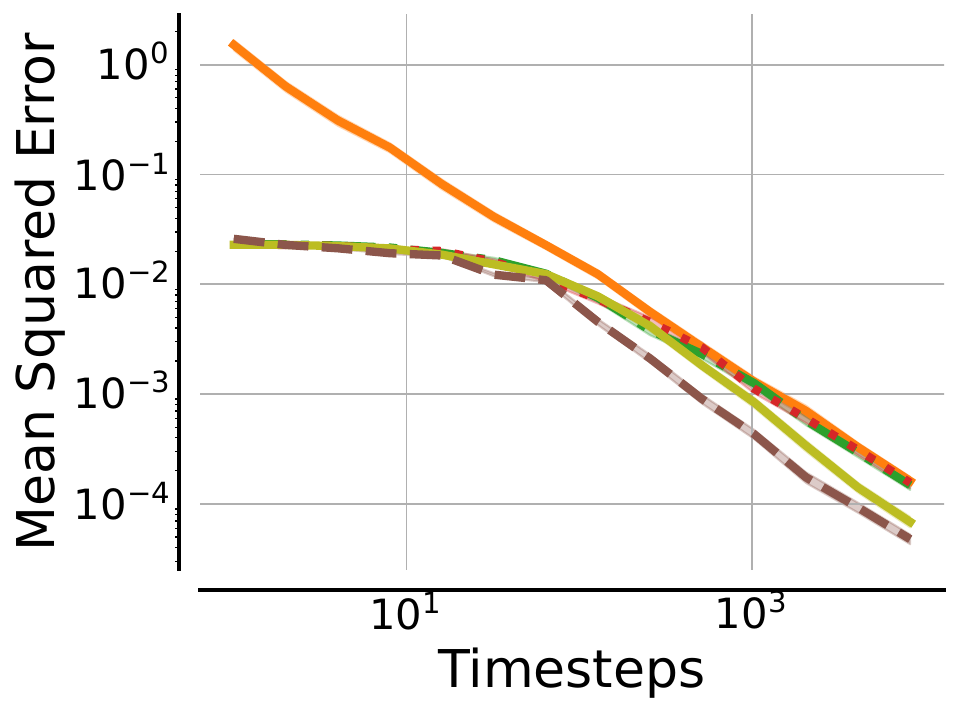}}
\subfigure[GridWorld]{\includegraphics[height=7.3em]{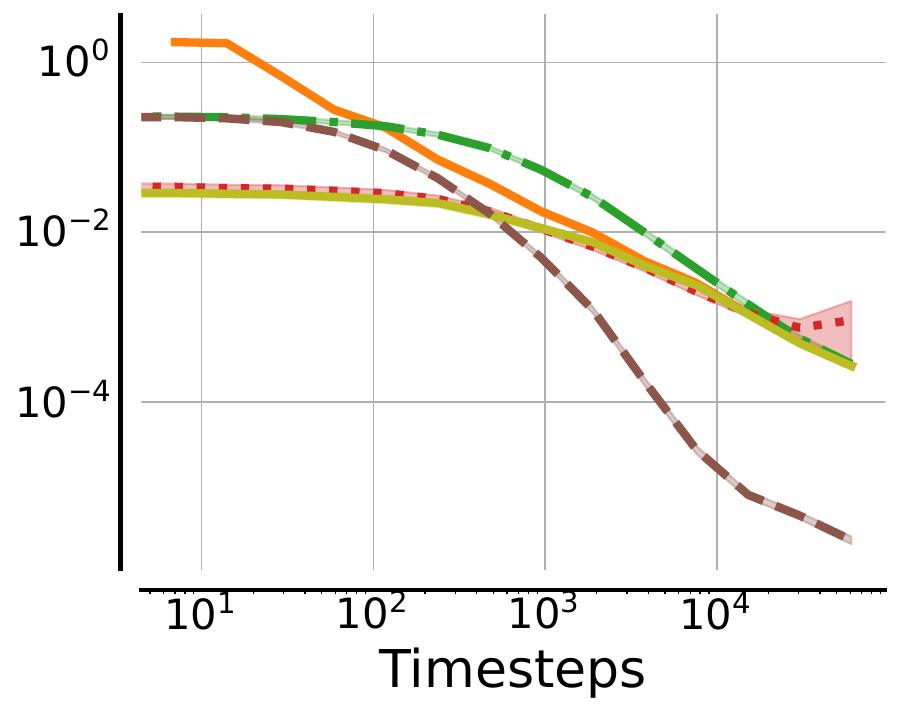}}
\subfigure[CartPole]{\includegraphics[height=7.3em]{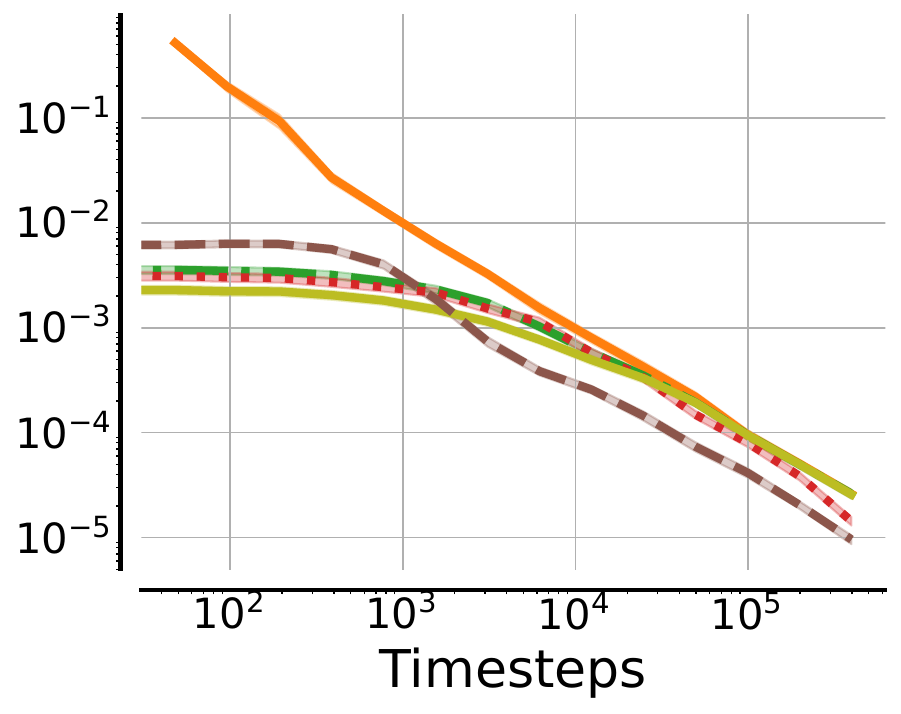}}
\subfigure[CartPoleContinuous]{\includegraphics[height=7.3em]{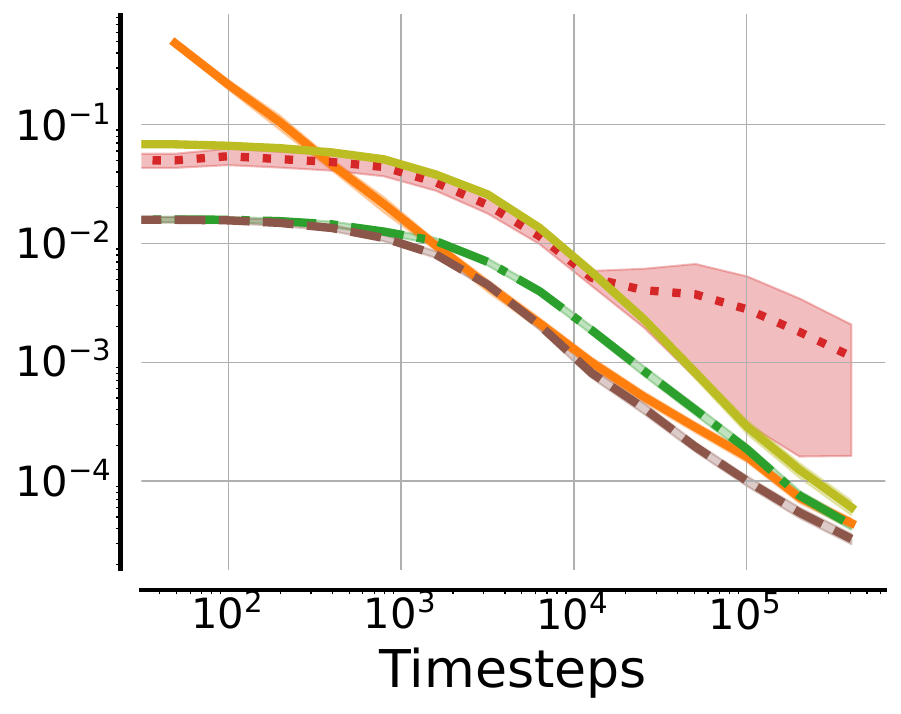}\label{fig:comb_mse_cpc}}
\subfigure{\includegraphics[width=0.66\textwidth]{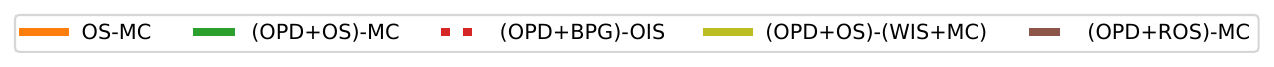}}
\vspace{-5pt}
\caption{Mean squared error (\MSE{}) of policy evaluation in the \textbf{with initial data} setting. Policy evaluation is conducted on the data collected from each strategy and a small set of initial data collected off-policy.
Axes and confidence intervals are the same as in Figure~\ref{fig:mse}.}
\label{fig:comb_mse}
\vspace{-10pt}
\end{figure}

Our next set of experiments considers a setting \textbf{with initial data}, in which a set of $100$ trajectories are already available and we wish to use these trajectories in our policy value estimate.
These trajectories are collected via i.i.d. \textit{off-policy} sampling with a behavior policy that is slightly different than $\pieval$.
This setting is intended to represent a setting where $\pieval$ has just been updated from an older policy and we would like to still use the off-policy data already collected from the older policy combined with the data to be collected.
In addition to the off-policy data (\OPD{}), we collect an additional $2^{13}\overline{T}$ steps of environment interaction with each  method.
\textit{We do not count the initial 100 trajectories towards the total collected data.}

For \ROS{}, we use the \OPD{} to initialize $\nabla_\btheta \loss$.
We expect to see that \ROS{} will collect data to combine with the \OPD{} such that the aggregate data set looks as if it had been collected with $\pieval$ to begin with.
We compare \ROS{} to the following baseline methods.
\textbf{(\OPD{} + \OS{})-\MC{}} collects additional data with \OS{} and uses the Monte Carlo estimator with the total data-set. \textbf{(\OPD{} + \OS{})-(\WIS{} + \MC{})} uses weighted importance sampling (\WIS{}) to compute an estimate from the \OPD{} combines the \WIS{} estimate with a Monte Carlo estimate using on-policy data. \textbf{(\OPD{} + \BPG{})-\OIS{}} collects additional data with \BPG{}{} and uses ordinary importance sampling as the estimator with all data. Finally, \textbf{(\OS{} - \MC{})} replaces the 100 initial trajectories with trajectories from \OS{}, then collects the remaining data with \OS{} and uses the Monte Carlo estimator.
In the case of \textbf{(\OS{} - \MC{})}, the 100 initial trajectories are counted towards the total data collected.

Figure \ref{fig:_kl_comb} shows that sampling error decreases fastest for \ROS{} as the additional data is collected.
For policy evaluation, we show \MSE{} for varying amounts of data in Figure \ref{fig:comb_mse} and provide numerical values for the final \MSE{} in Appendix \ref{app:stat}.
We observe that the initial data provides an immediate reduction in \MSE{} at the expense of injecting bias into the estimates.
\textbf{(\OPD{}+\OS{})-\MC{}} struggles to reduce this bias while \textbf{(\OPD{}+\ROS{})-\MC{}} is able to through data collection.
Overall, this result highlights that \ROS{} can collect additional data that reduces sampling error in the aggregate data set and produce lower \MSE{} estimates compared to other data collection methods.
Intuitively, \OS{} requires many more samples to dilute the bias brought on by using \OPD{} in the Monte Carlo estimator, while \ROS{} is able to correct the empirical off-policy distribution to the expected on-policy distribution and use the Monte Carlo estimator without any off-policy corrections.

The comparison to \OS{}-\MC{} demonstrates the potential of \ROS{} for correcting an off-policy empirical distribution to the expected on-policy distribution.
As noted above, \OS{}-\MC{} has 100 fewer trajectories than the other baselines.
However -- even when including the initial 100 off-policy trajectories in the data total for all methods -- \ROS{} eventually obtains lower \MSE{} compared to \OS{}-\MC{}.
In this sense, \ROS{} has taken an initially biased dataset and collected the right trajectories to make it look as-if the evaluation policy had collected all trajectories in the first place.

\subsection{Sensitivity Study}

\begin{figure*}
\centering
\subfigure[GridWorld]{\includegraphics[width=0.245\textwidth]{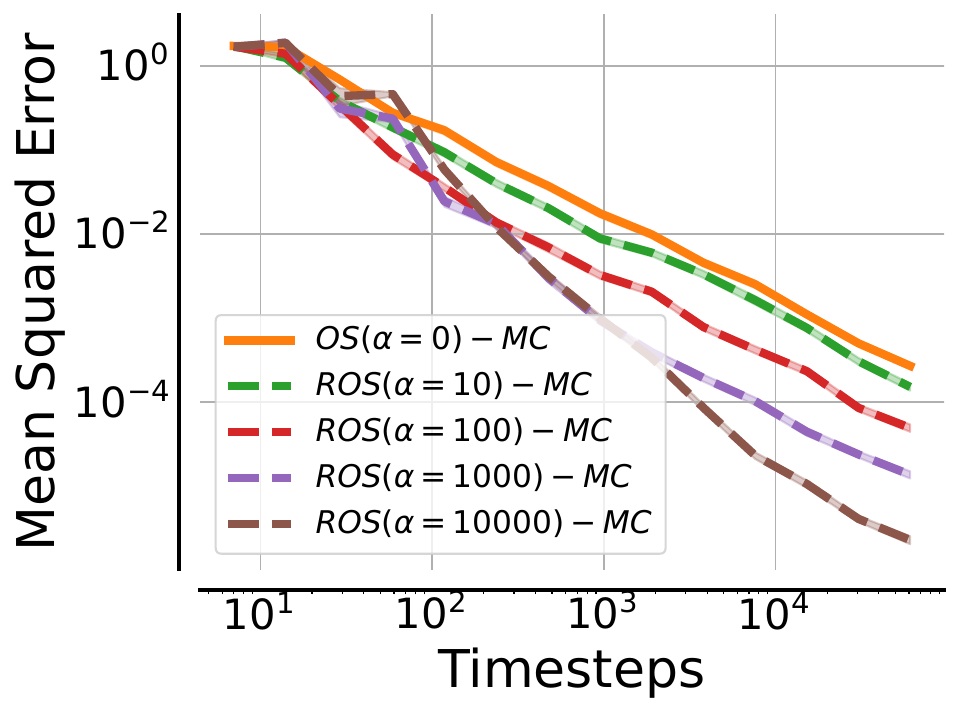}\label{fig:alpha_gw}}
\subfigure[CartPole]{\includegraphics[width=0.245\textwidth]{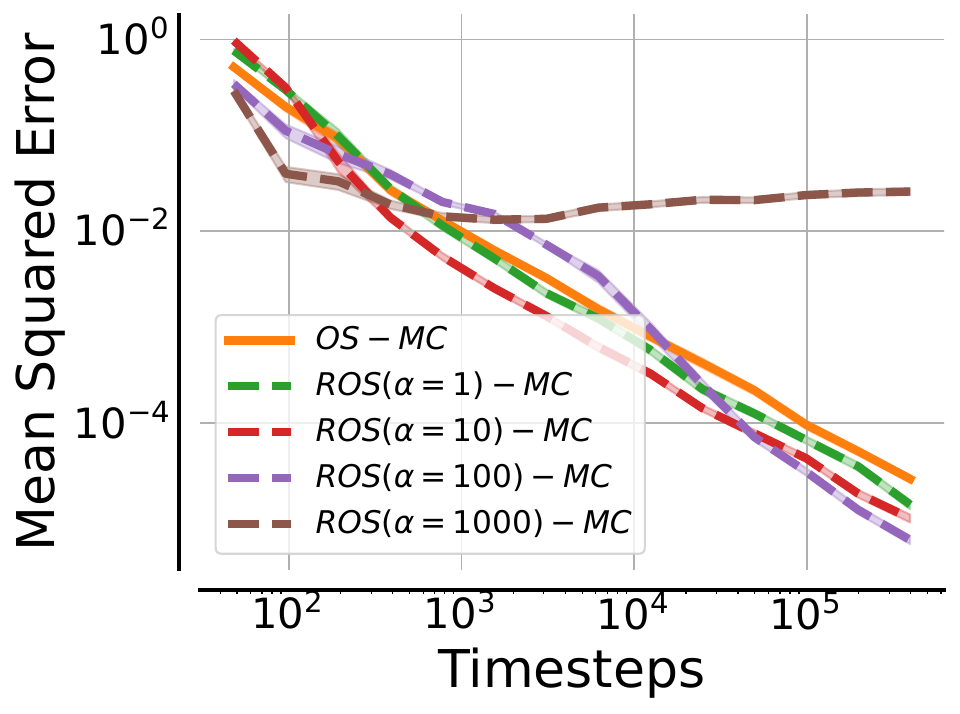}\label{fig:alpha_cp}}
\subfigure[Environment Noise]{\includegraphics[width=0.245\textwidth]{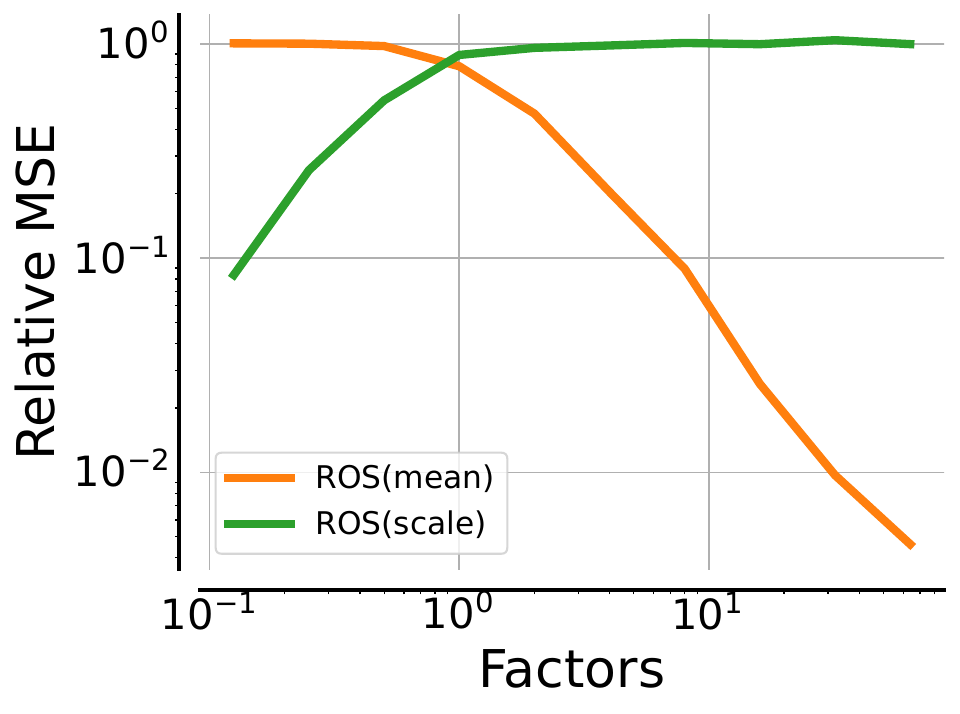}\label{fig:imp_env_mb}}
\subfigure[Policy Noise]{\includegraphics[width=0.245\textwidth]{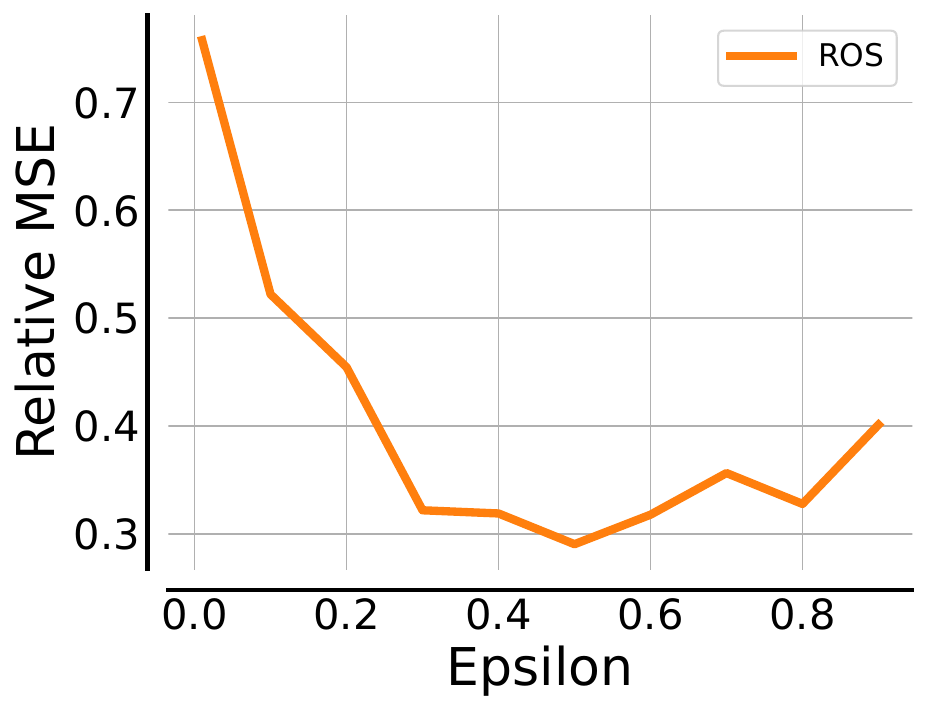}\label{fig:imp_pie_mb}}
\caption{\MSE{} of \ROS{} with different step-size, $\alpha$ (\ref{fig:alpha_gw} and \ref{fig:alpha_cp}). %
Relative improvement of \ROS{} in Bandit compared to \OS{} with different stochasticity in the environment (\ref{fig:imp_env_mb}) and policy (\ref{fig:imp_pie_mb}). The relative MSE is computed as the MSE of ROS divided by the MSE of OS. Results in these figures are averaged over $500$ trials.}
\label{fig:imp_mb}
\vspace{-10pt}
\end{figure*}

\looseness=-1
Finally, we evaluate the sensitivity of \ROS{} to hyper-parameter, environment, and policy settings.
\ROS{} requires setting a step size, $\alpha$, which controls how much \ROS{} updates the behavior policy away from $\pieval$.
We show \MSE{} curves for \ROS{} with different step size $\alpha$ on GridWorld and CartPole in Figures~\ref{fig:imp_mb} ($\alpha=0$ corresponds to \OS{}).
Figure~\ref{fig:alpha_gw} shows that, in GridWorld, \ROS{} with any tested step-size produces lower \MSE{} policy evaluation than \OS{} for any data set size.
As it collects more data, \ROS{} with larger $\alpha$ enables lower \MSE{} because the norm of $\nabla_\btheta \loss$ decreases as sampling error decreases, and thus a larger $\alpha$ is required to make significant updates.
A larger $\alpha$ value is also in line with our theoretical results which prescribe $\alpha\rightarrow\infty$.
However, in CartPole, (Figure~\ref{fig:alpha_cp}), \ROS{} with the largest tested $\alpha$ (1000) diverges and the second largest ($\alpha=100$) requires many steps before it improves upon \OS{}.
Thus, in domains with continuous state-spaces, more conservative $\alpha$ values may be preferred.

Our final set of experiments considers how the stochasticity of a domain and entropy of $\pieval$ affect the relative improvement that \ROS{} offers.
In this sub-section, we study these settings in the Bandit domain for its simplicty; similar experimental results in GridWorld can be found in Appendix~\ref{sec:imp_gw}. 
We choose $\alpha=1000$ for the following experiments.

To study domain stochasticity, we first create variants of the Bandit environment by multiplying either the mean or scale of the reward distribution of each action by a varying factor.
In each experimental trial, we use \ROS{} to collect $1000\overline{T}$ steps for the Monte Carlo estimator and compute the relative \MSE{} compared to the Monte Carlo estimator using \OS{} with the same number of steps.
Figure~\ref{fig:imp_env_mb} shows that as the factor on the mean increases, \ROS{} provides a greater reduction in \MSE{} as even small amounts of sampling error translate into large \MSE{} when the reward means are large.
On the other hand, as the scale factor increases, the \MSE{} is dominated by reward noise and the relative benefit of reducing sampling error disappears.

We also evaluate the relative improvement of \ROS{}  as a function of the entropy of $\pieval$.
For $\pieval$, we use $\epsilon$-greedy policies which select the optimal action in a state with probability $1-\epsilon$ and otherwise select an action uniformly at random.
Relative improvement in \MSE{} is shown in Figure~\ref{fig:imp_pie_mb}.
For all $\epsilon$, \ROS{} improves upon the \MSE{} of \OS{}. The improvement is generally larger for more stochastic $\pieval$ when sampling error in action selection will be highest.

\section{Discussion and Future Work}\label{sec:discussion}

This work has shown that off-policy non-i.i.d.\ sampling can produce data sets that more closely approximate the on-policy data distribution than on-policy i.i.d.\ sampling.
We considered the problem of policy evaluation and showed that more closely approximating the on-policy data distribution leads to more data efficient policy evaluation across several domains.
As far as we know, \ROS{} is the first data collection method for policy evaluation that uses off-policy sampling to produce more closely on-policy data than the data produced by on-policy sampling.

While \ROS{} is a first step towards off-policy algorithms that produce data matching a target distribution, we highlight a few limitations of the algorithm and our study.
In our view, the main limitations of the \ROS{} algorithm are the need to set a step-size parameter (in contrast to parameter-free on-policy sampling) and the need to update $\nabla_\btheta \loss$ at each action step.
For the former, future work should investigate robust methods for setting the step-size, particularly in settings where $\pitheta$ generalizes across the state-space.
For the latter limitation, a future study could consider only updating $\nabla_\btheta \loss$ at the end of each episode instead of after each action choice (assuming more computation can be done between episodes).
In terms of our study, for this paper we chose to study  many different facets of \ROS{} on a suite of simpler domains (see the appendices for additional ablations and extensions); a future study should assess the scalability of \ROS{} with more complex function approximators.
Finally, our theoretical results were conducted in the tabular setting; an important open question is at what rate \ROS{} converges when $\pitheta$ uses a function approximator that must generalize across states.
Beyond these minor technical limitations, our paper addresses fundamental research questions in \RL{} and thus we do not see obvious negative societal impacts that are unique to this work in comparison to other work in \RL{} and policy evaluation.

\looseness=-1
While we evaluated \ROS{} for policy evaluation, the long-term importance of this work may be in exploring the distinction between on-policy sampling and on-policy data.
On-policy \RL{} algorithms require on-policy data and our work suggests that adaptive off-policy sampling can produce on-policy data more efficiently than on-policy sampling.
In the future, we wish to study these insights for on-policy policy improvement algorithms (e.g., policy gradient methods \citep{reinforce,schulman2017proximal}) and to extend our convergence results to non-tabular settings. %

\section{Conclusion}
In this paper, we have introduced a novel data collection method for policy evaluation in reinforcement learning environments.
Our method -- Robust On-Policy Sampling (\ROS{}) -- considers previously collected data when selecting actions to reduce sampling error in the entire collected data set.
We show both in theory and in practice that data from \ROS{} converges faster to the on-policy data distribution compared to on-policy sampling.
Empirically, we find that faster convergence to the on-policy data distributions lowers the \MSE{}  of policy evaluation.

\begin{ack}
We thank Ishan Durugkar, Brahma Pavse, Subhojyoti Mukherjee, Elliot Fosong, and Filippos Christianos for their feedback which greatly strengthened the paper.
We also wish to acknowledge the anonymous reviewers for their comments and constructive criticisms.
Support for this research was provided by the Office of the Vice Chancellor for Research and Graduate Education at
the University of Wisconsin — Madison with funding from the Wisconsin Alumni Research Foundation.

\end{ack}

\bibliographystyle{plainnat}
\bibliography{ros-sources}

\begin{thebibliography}{34}
\providecommand{\natexlab}[1]{#1}
\providecommand{\url}[1]{\texttt{#1}}
\expandafter\ifx\csname urlstyle\endcsname\relax
  \providecommand{\doi}[1]{doi: #1}\else
  \providecommand{\doi}{doi: \begingroup \urlstyle{rm}\Url}\fi

\bibitem[Agarwal et~al.(2022)Agarwal, Jiang, Kakade, and
  Sun]{agarwal2022reinforcement}
Alekh Agarwal, Nan Jiang, Sham~M. Kakade, and Wen Sun.
\newblock \emph{Reinforcement learning: Theory and algorithms}.
\newblock 2022.

\bibitem[Alexanderian(2009)]{alexanderian2009some}
Alen Alexanderian.
\newblock Some notes on asymptotic theory in probability.
\newblock \emph{Notes. University of Maryland}, 2009.

\bibitem[Arnold et~al.(2022)Arnold, L'Ecuyer, Chen, Chen, and
  Sha]{arnold2022policy}
S\'ebastien M.~R. Arnold, Pierre L'Ecuyer, Liyu Chen, Yi-fan Chen, and Fei Sha.
\newblock Policy learning and evaluation with randomized quasi-monte carlo.
\newblock In \emph{International Conference on Artificial Intelligence and
  Statistics}, 2022.

\bibitem[Bouchard et~al.(2016)Bouchard, Trouillon, Perez, and
  Gaidon]{bouchard2016online}
Guillaume Bouchard, Th{\'e}o Trouillon, Julien Perez, and Adrien Gaidon.
\newblock Online learning to sample.
\newblock \emph{arXiv preprint arXiv:1506.09016}, 2016.

\bibitem[Brockman et~al.(2016)Brockman, Cheung, Pettersson, Schneider,
  Schulman, Tang, and Zaremba]{gym}
Greg Brockman, Vicki Cheung, Ludwig Pettersson, Jonas Schneider, John Schulman,
  Jie Tang, and Wojciech Zaremba.
\newblock {OpenAI} {Gym}.
\newblock \emph{arXiv preprint arXiv:1606.01540}, 2016.

\bibitem[Ciosek and Whiteson(2017)]{ciosek2017offer}
Kamil Ciosek and Shimon Whiteson.
\newblock O{F}{F}{E}{R}: Off-environment reinforcement learning.
\newblock In \emph{AAAI Conference on Artificial Intelligence}, 2017.

\bibitem[Farajtabar et~al.(2018)Farajtabar, Chow, and
  Ghavamzadeh]{farajtabar2018more}
Mehrdad Farajtabar, Yinlam Chow, and Mohammad Ghavamzadeh.
\newblock More robust doubly robust off-policy evaluation.
\newblock In \emph{International Conference on Machine Learning}, 2018.

\bibitem[Frank et~al.(2008)Frank, Mannor, and Precup]{frank2008reinforcement}
Jordan Frank, Shie Mannor, and Doina Precup.
\newblock Reinforcement learning in the presence of rare events.
\newblock In \emph{International Conference on Machine Learning}, 2008.

\bibitem[Hanna et~al.(2017)Hanna, Thomas, Stone, and
  Niekum]{hanna2017data-efficient}
Josiah~P. Hanna, Philip~S. Thomas, Peter Stone, and Scott Niekum.
\newblock Data-efficient policy evaluation through behavior policy search.
\newblock In \emph{International Conference on Machine Learning}, 2017.

\bibitem[Hanna et~al.(2021)Hanna, Niekum, and Stone]{hanna2021importance}
Josiah~P. Hanna, Scott Niekum, and Peter Stone.
\newblock Importance {Sampling} in {Reinforcement} {Learning} with an
  {Estimated} {Behavior} {Policy}.
\newblock \emph{Machine Learning}, 110\penalty0 (6):\penalty0 1267--1317, May
  2021.

\bibitem[Harris et~al.(2020)Harris, Millman, van~der Walt, Gommers, Virtanen,
  Cournapeau, Wieser, Taylor, Berg, Smith, Kern, Picus, Hoyer, van Kerkwijk,
  Brett, Haldane, del R{\'{i}}o, Wiebe, Peterson, G{\'{e}}rard-Marchant,
  Sheppard, Reddy, Weckesser, Abbasi, Gohlke, and Oliphant]{harris2020array}
Charles~R. Harris, K.~Jarrod Millman, St{\'{e}}fan~J. van~der Walt, Ralf
  Gommers, Pauli Virtanen, David Cournapeau, Eric Wieser, Julian Taylor,
  Sebastian Berg, Nathaniel~J. Smith, Robert Kern, Matti Picus, Stephan Hoyer,
  Marten~H. van Kerkwijk, Matthew Brett, Allan Haldane, Jaime~Fern{\'{a}}ndez
  del R{\'{i}}o, Mark Wiebe, Pearu Peterson, Pierre G{\'{e}}rard-Marchant,
  Kevin Sheppard, Tyler Reddy, Warren Weckesser, Hameer Abbasi, Christoph
  Gohlke, and Travis~E. Oliphant.
\newblock Array programming with {NumPy}.
\newblock \emph{Nature}, 585\penalty0 (7825):\penalty0 357--362, 2020.

\bibitem[Jiang and Li(2016)]{jiang2016doubly}
Nan Jiang and Lihong Li.
\newblock Doubly robust off-policy evaluation for reinforcement learning.
\newblock In \emph{International Conference on Machine Learning}, 2016.

\bibitem[Konyushova et~al.(2021)Konyushova, Chen, Paine, Gulcehre, Paduraru,
  Mankowitz, Denil, and de~Freitas]{konyushkova2021active}
Ksenia Konyushova, Yutian Chen, Thomas Paine, Caglar Gulcehre, Cosmin Paduraru,
  Daniel~J. Mankowitz, Misha Denil, and Nando de~Freitas.
\newblock Active offline policy selection.
\newblock In \emph{Advances in Neural Information Processing Systems}, 2021.

\bibitem[Le et~al.(2019)Le, Voloshin, and Yue]{le2019batch}
Hoang Le, Cameron Voloshin, and Yisong Yue.
\newblock Batch policy learning under constraints.
\newblock In \emph{International Conference on Machine Learning}, 2019.

\bibitem[Li et~al.(2015)Li, Munos, and Szepesv{\'a}ri]{li2015toward}
Lihong Li, R{\'e}mi Munos, and Csaba Szepesv{\'a}ri.
\newblock Toward minimax off-policy value estimation.
\newblock In \emph{International Conference on Artificial Intelligence and
  Statistics}, 2015.

\bibitem[Mardia et~al.(2019)Mardia, Jiao, Tánczos, Nowak, and
  Weissman]{10.1093/imaiai/iaz025}
Jay Mardia, Jiantao Jiao, Ervin Tánczos, Robert~D Nowak, and Tsachy Weissman.
\newblock Concentration inequalities for the empirical distribution of discrete
  distributions: beyond the method of types.
\newblock \emph{Information and Inference: A Journal of the IMA}, 9\penalty0
  (4):\penalty0 813--850, 2019.

\bibitem[Mukherjee et~al.(2022)Mukherjee, Hanna, and
  Nowak]{mukherjee_revar_2022}
Subhojyoti Mukherjee, Josiah~P. Hanna, and Robert Nowak.
\newblock {ReVar}: {Strengthening} {Policy} {Evaluation} via {Reduced}
  {Variance} {Sampling}.
\newblock In \emph{{International} {Conference} on {Uncertainty} in
  {Artificial} {Intelligence} ({UAI})}, August 2022.

\bibitem[Narita et~al.(2019)Narita, Yasui, and Yata]{narita2019efficient}
Yusuke Narita, Shota Yasui, and Kohei Yata.
\newblock Efficient counterfactual learning from bandit feedback.
\newblock In \emph{AAAI Conference on Artificial Intelligence}, 2019.

\bibitem[O'Hagan(1987)]{ohagan1987monte}
Anthony O'Hagan.
\newblock Monte carlo is fundamentally unsound.
\newblock \emph{The Statistician}, pages 247--249, 1987.

\bibitem[Oosterhuis and de~Rijke(2020)]{oosterhuis2020taking}
Harrie Oosterhuis and Maarten de~Rijke.
\newblock Taking the counterfactual online: Efficient and unbiased online
  evaluation for ranking.
\newblock In \emph{International Conference on Theory of Information
  Retrieval}, 2020.

\bibitem[Ostrovski et~al.(2017)Ostrovski, Bellemare, Oord, and
  Munos]{ostrovski2017count}
Georg Ostrovski, Marc~G. Bellemare, A{\"a}ron Oord, and R{\'e}mi Munos.
\newblock Count-based exploration with neural density models.
\newblock In \emph{International conference on machine learning}, 2017.

\bibitem[Paszke et~al.(2019)Paszke, Gross, Massa, Lerer, Bradbury, Chanan,
  Killeen, Lin, Gimelshein, Antiga, Desmaison, Kopf, Yang, DeVito, Raison,
  Tejani, Chilamkurthy, Steiner, Fang, Bai, and Chintala]{NEURIPS2019_9015}
Adam Paszke, Sam Gross, Francisco Massa, Adam Lerer, James Bradbury, Gregory
  Chanan, Trevor Killeen, Zeming Lin, Natalia Gimelshein, Luca Antiga, Alban
  Desmaison, Andreas Kopf, Edward Yang, Zachary DeVito, Martin Raison, Alykhan
  Tejani, Sasank Chilamkurthy, Benoit Steiner, Lu~Fang, Junjie Bai, and Soumith
  Chintala.
\newblock {PyTorch}: An imperative style, high-performance deep learning
  library.
\newblock In \emph{Advances in Neural Information Processing Systems}. 2019.

\bibitem[Pavse et~al.(2020)Pavse, Durugkar, Hanna, and
  Stone]{pavse2020reducing}
Brahma~S. Pavse, Ishan Durugkar, Josiah~P. Hanna, and Peter Stone.
\newblock Reducing sampling error in batch temporal difference learning.
\newblock In \emph{International Conference on Machine Learning}, 2020.

\bibitem[Puterman(2014)]{puterman2014markov}
Martin~L. Puterman.
\newblock \emph{Markov decision processes: discrete stochastic dynamic
  programming}.
\newblock John Wiley \& Sons, 2014.

\bibitem[Rasmussen and Ghahramani(2003)]{rasmussen2003bayesian}
Carl~Edward Rasmussen and Zoubin Ghahramani.
\newblock Bayesian monte carlo.
\newblock In \emph{Advances in Neural Information Processing Systems}, 2003.

\bibitem[Sch\"afer et~al.(2022)Sch\"afer, Christianos, Hanna, and
  Albrecht]{schaefer2022derl}
Lukas Sch\"afer, Filippos Christianos, Josiah~P. Hanna, and Stefano~V.
  Albrecht.
\newblock Decoupled reinforcement learning to stabilise intrinsically-motivated
  exploration.
\newblock In \emph{International Conference on Autonomous Agents and Multiagent
  Systems}, 2022.

\bibitem[Schulman et~al.(2017)Schulman, Wolski, Dhariwal, Radford, and
  Klimov]{schulman2017proximal}
John Schulman, Filip Wolski, Prafulla Dhariwal, Alec Radford, and Oleg Klimov.
\newblock Proximal policy optimization algorithms.
\newblock \emph{arXiv preprint arXiv:1707.06347}, 2017.

\bibitem[Sen and Singer(1993)]{sen1993large}
Pranab~K. Sen and Julio~M. Singer.
\newblock \emph{Large Sample Methods in Statistics: An Introduction with
  Applications}.
\newblock Chapman \& Hall, 1993.

\bibitem[Sutton and Barto(1998)]{sutton1998reinforcement}
Richard~S. Sutton and Andrew~G. Barto.
\newblock \emph{Reinforcement Learning: An Introduction}.
\newblock MIT Press, 1998.

\bibitem[Tang et~al.(2017)Tang, Houthooft, Foote, Stooke, Chen, Duan, Schulman,
  De~Turck, and Abbeel]{tang2017exploration}
Haoran Tang, Rein Houthooft, Davis Foote, Adam Stooke, Xi~Chen, Yan Duan, John
  Schulman, Filip De~Turck, and Pieter Abbeel.
\newblock \# exploration: A study of count-based exploration for deep
  reinforcement learning.
\newblock In \emph{Advances in neural information processing systems}, 2017.

\bibitem[Thomas and Brunskill(2016)]{thomas2016data-efficient}
Philip~S. Thomas and Emma Brunskill.
\newblock Data-efficient off-policy policy evaluation for reinforcement
  learning.
\newblock In \emph{International Conference on Machine Learning}, 2016.

\bibitem[Tucker and Joachims(2022)]{tucker2022variance-optimal}
Aaron~David Tucker and Thorsten Joachims.
\newblock Variance-optimal augmentation logging for counterfactual evaluation
  in contextual bandits.
\newblock \emph{arXiv preprint arXiv:2202.01721}, 2022.

\bibitem[Williams(1992)]{reinforce}
Ronald~J. Williams.
\newblock Simple statistical gradient-following algorithms for connectionist
  reinforcement learning.
\newblock \emph{Machine learning}, 8\penalty0 (3):\penalty0 229--256, 1992.

\bibitem[Zinkevich et~al.(2006)Zinkevich, Bowling, Bard, Kan, and
  Billings]{Zinkevich2006}
Martin Zinkevich, Michael Bowling, Nolan Bard, Morgan Kan, and Darse Billings.
\newblock Optimal unbiased estimators for evaluating agent performance.
\newblock In \emph{AAAI Conference on Artificial Intelligence}, 2006.

\end{thebibliography}

\section*{Checklist}

\begin{enumerate}

\item For all authors...
\begin{enumerate}
  \item Do the main claims made in the abstract and introduction accurately reflect the paper's contributions and scope?
    \answerYes{}
  \item Did you describe the limitations of your work?
    \answerYes{See Section \ref{sec:discussion}.}{}
  \item Did you discuss any potential negative societal impacts of your work?
    \answerYes{See Section \ref{sec:discussion}.}
  \item Have you read the ethics review guidelines and ensured that your paper conforms to them?
    \answerYes{}
\end{enumerate}

\item If you are including theoretical results...
\begin{enumerate}
  \item Did you state the full set of assumptions of all theoretical results?
    \answerYes{See Section \ref{sec:theory} and Appendix \ref{app:theory}.}
    \item Did you include complete proofs of all theoretical results?
    \answerYes{See Appendices \ref{app:prop-bias} and \ref{app:theory}.}
\end{enumerate}

\item If you ran experiments...
\begin{enumerate}
  \item Did you include the code, data, and instructions needed to reproduce the main experimental results (either in the supplemental material or as a U\RL{})?
    \answerYes{Included in supplementary material.}
  \item Did you specify all the training details (e.g., data splits, hyperparameters, how they were chosen)?
    \answerYes{See Section \ref{sec:empirical} and Appendices \ref{app:pi_e} and \ref{app:hparams}.}
        \item Did you report error bars (e.g., with respect to the random seed after running experiments multiple times)?
    \answerYes{}
    \item Did you include the total amount of compute and the type of resources used (e.g., type of GPUs, internal cluster, or cloud provider)?
    \answerNo{Experiments use \RL{} domains and algorithms that can be ran on a typical personal computer. Minimal compute resources required to reproduce any experiment in the paper.}
\end{enumerate}

\item If you are using existing assets (e.g., code, data, models) or curating/releasing new assets...
\begin{enumerate}
  \item If your work uses existing assets, did you cite the creators?
    \answerYes{}
  \item Did you mention the license of the assets?
    \answerNo{}
  \item Did you include any new assets either in the supplemental material or as a U\RL{}?
    \answerNo{}
  \item Did you discuss whether and how consent was obtained from people whose data you're using/curating?
    \answerNA{}
  \item Did you discuss whether the data you are using/curating contains personally identifiable information or offensive content?
    \answerNA{}
\end{enumerate}

\item If you used crowdsourcing or conducted research with human subjects...
\begin{enumerate}
  \item Did you include the full text of instructions given to participants and screenshots, if applicable?
    \answerNA{}
  \item Did you describe any potential participant risks, with links to Institutional Review Board (IRB) approvals, if applicable?
    \answerNA{}
  \item Did you include the estimated hourly wage paid to participants and the total amount spent on participant compensation?
    \answerNA{}
\end{enumerate}

\end{enumerate}

\newpage
\appendix

\section{Proof of Proposition 1}\label{app:prop-bias}

In this appendix we prove \Cref{proposition:bias} from Section 4.

\propositionbias*

\begin{proof}
In the following, we use $\textbf{E}_\pieval[\cdot]$ as shorthand for $\textbf{E}[\cdot | D \sim \pieval]$.
Let $\hat{v}_1 \coloneqq \frac{1}{n_{\dataset_1}} \sum_{i=1}^{n_{\dataset_1}} g(h_i) = \operatorname{MC}(\dataset_1)$. %
\begin{align}
    \textbf{E}_\pieval \biggl[\operatorname{MC}(\dataset_1 \cup D_2) \biggr] &=
    \textbf{E}_\pieval\left[\frac{1}{n} \sum_{i=1}^{n_{\dataset_1}} g(h_i) + \frac{1}{n} \sum_{i=1}^{n_{\data_2}} g(H_i)\right]\\
    &= \textbf{E}_\pieval\left[\frac{1}{n} \sum_{i=1}^{n_{\dataset_1}} g(h_i)\right] + \textbf{E}_\pieval\left[\frac{1}{n} \sum_{i=1}^{n_{\data_2}} g(H_i)\right] \\
    &\overset{(a)}{=} \frac{1}{n} \frac{n_{\dataset_1}}{n_{\dataset_1}} \sum_{i=1}^{n_{\dataset_1}} g(h_i) + \textbf{E}_\pieval\left[\frac{1}{n}\frac{n_{\data_2}}{n_{\data_2}} \sum_{i=1}^{n_{\data_2}} g(H_i)\right] \\
    &\overset{(b)}{=} \frac{n_{\dataset_1}}{n}\hat{v}_1 + \frac{n_{\data_2}}{n}\textbf{E}_\pieval\left[\underbrace{\frac{1}{n_{\data_2}} \sum_{i=1}^{n_{\data_2}} g(H_i)}_{\operatorname{MC}(D_2)}\right] \\
    &\overset{(c)}{=} \frac{n_{\dataset_1}}{n}\hat{v}_1 + \frac{n_{\data_2}}{n} v(\pieval) \\
    &\overset{}{=} \frac{n_{\dataset_1}}{n}(\hat{v}_1 - v(\pieval)) + \frac{n_{\data_2}}{n} v(\pieval) + \frac{n_{\dataset_1}}{n}v(\pieval)\\
    &\overset{}{=} \frac{n_{\dataset_1}}{n}(\hat{v}_1 - v(\pieval)) + v(\pieval) \label{eq:final}
\end{align}
where \textbf{(a)} uses the fact that no random variables appear inside the first expectation, \textbf{(b)} uses the definition of $\hat{v}$, and \textbf{(c)} uses the fact that $\operatorname{MC}(D_2)$ is an unbiased estimator of $v(\pieval)$.
The proposition follows by observing that equation \ref{eq:final} is equal to $v(\pieval)$ if and only if $n_{\dataset_1} = 0$ (i.e., $\dataset_1 = \emptyset$) or if $\hat{v} = v(\pieval)$.

Note that the latter case in which the data-conditioned Monte Carlo estimator is unbiased only occurs when the Monte Carlo estimate \textit{only using} $\dataset_1$ has non-zero error.
If $g(H)$ has non-zero variance under on-policy sampling then the probability that $\operatorname{MC}(\dataset_1)$ is exactly equal to $v(\pieval)$ with a finite-size $D_2$ is low.

\end{proof}

\section{Proofs of ROS Properties}\label{app:theory}

Before giving the proofs for Theorems \ref{theorem:converge} and \ref{theorem:rate} we re-state our key assumption.
\assumptionparams*

We next derive two lemmas that will be used in the proofs of our theorems.
\begin{restatable}[]{rlemma}{lemmaconvergence}
Let $S_m(a)$ denote the number of times that action $a$ was taken after visiting a particular state, $s$, $m$ times in $\dataset$ and let $k \coloneqq |\mathcal{A}|$.
Under Assumption \ref{ass:param} and \ROS{} collection of $\dataset$, we have that:
\begin{equation}
    \sup_{a \in \mathcal{A}}|S_m(a) - m \pieval(a|s)| \le k-1.
\end{equation}
\label{lemma:bound}
\end{restatable}
\begin{proof}

We first formally show that Assumption \ref{ass:param} implies that \ROS{} always taking the most under-sampled action at each time-step.
Because we only consider a single state, we suppress dependencies on the state in this proof, e.g., we write $\pieval(a)$ instead of $\pieval(a|s)$ and $\theta_a$ instead of $\theta_{s,a}$ for the softmax parameters.
We have $\loss(\pitheta) = \sum_{a} S_m(a) \theta_a - m \log(\sum_{b\in\mathcal{A}}e^{\theta_b})$ and $\nabla_{\theta_a} \loss(\pitheta)\vert_{\pitheta=\pieval} = S_m(a)(1 - \pieval(a)) - \pieval(a) (m - S_m(a)) = S_m(a) - m \pieval(a)$, which is exactly the number of times that action $a$ was over-sampled.
The softmax parameter for action $a$ after the \ROS{} update is given by:
\[
\theta_a^\prime = {\theta_{e,a}} - \alpha (S_m(a) - m\pieval(a))
\]
where $\theta_{e,a}$ is the softmax parameter for action $a$ under the evaluation policy.
Taking the limit as $\alpha \rightarrow \infty$, we see that the $\theta_{e,a}$ term is dominated by the $\alpha(S_m(a) - \pieval(a)m)$ term.
Thus the parameter for each action is:
\[
\theta_a^\prime = \alpha ((m\pieval(a) - S_m(a)).
\]
Interpreting $\alpha$ as the inverse of the softmax temperature, we can see that $\alpha \rightarrow \infty$ corresponds to a softmax temperature $\rightarrow 0$ which in turn corresponds to a hard max over the $(m\pieval(a) - S_m(a))$ values.
Thus, the updated behavior policy puts all probability mass on the action for which $\pieval(a)m - S_m(a)$ is largest.
Hence we select the most under-sampled action if we take $\alpha \rightarrow \infty$ in Algorithm \ref{alg:ros2}.

Now we show that under \ROS{} action selection that the amount of over- or under-sampling is bounded.
First, we observe that,
\[
    \sum_{a \in \mathcal{A}}(S_m(i)-m \pieval(i))= m - m = 0,
\]
and we claim that, 
\[
  S_m(a) - m \pieval(a) \leq 1  
\]
for any $a \in \mathcal{A}$. We prove this claim by contradiction.
Assume not: $S_m(j) - m \pieval(j) > 1$ for some $j \in \mathcal{A}$.
Let $X_m(j) \coloneqq 1$ if action $j$ was taken at step $m$ and $0$ otherwise.
If $X_m(j) = 1$, we have that:
\begin{align*}
    S_m(j) - m \pieval(i) &> 1 \implies \\
    S_{m-1}(j) + 1 - (m-1) \pieval(j) - \pieval(j) &> 1 \implies\\
    S_{m-1}(j) - (m - 1) \pieval(j) &> 1 - 1 + \pieval(j) > 0.
\end{align*}        
However, this results in a contradiction since we have that action $j$ was over-sampled at the previous step but action $j$ would not be selected by \ROS{} if it was over-sampled.
So, in order for $S_m(j) - m\pieval(j) > 1$, we must have that $X_m(j) = 0$.
Combining the fact that $X_m(j) = 0$ with our assumption that $S_m(j) - m \pieval(j) > 1$ tells us that:
\begin{align*}
    &S_{m}(j) - m \pieval(j) > 1 \\
    \overset{(a)}{\implies} &S_{m-1}(j) - m \pieval(j) > 1 \\
    \implies &S_{m-1}(j) - (m-1) \pieval(j) > 1
\end{align*}
where \textbf{(a)} is because $S_{m-1}(j)$ must be equal to $S_m(j)$ if $X_m(j)=0$.
By the same logic we get that $X_{m-1}(j)=x_{m-2}(j)=\cdots=X_1(j)=0$ which implies that $S_1(j) - \pieval(j) = 0 - \pieval > 1$ which is a contradiction.
Thus, we conclude that $S_m(j) - m \pieval(j) \leq 1$ for any $j$, i.e., any action can be \textit{over-sampled} by at most 1.
Combining this conclusion with the observation that $\sum_{a\in\mathcal{A}} S_m(a) - m \pieval(a) = m - m = 0$ tells us that any action can be \textit{under-sampled} by at most $k-1$.
Thus, the absolute difference $|S_m(a) - m \pieval(a)|$ is at most $k - 1$ for any action $a\in \mathcal{A}$ which completes the proof.

\end{proof}

\begin{restatable}[]{rlemma}{lemmaconverge}
Let $s$ be a state that we visit $m$ times. Under \ROS{} sampling, we have $\forall a \in \mathcal{A}$ that:
\[
\lim_{m\rightarrow\infty} \pi_D(a|s) = \pieval(a|s).
\]
\label{lemma:converge}
\end{restatable}
\begin{proof}
The proof follows from Lemma \ref{lemma:bound}.
As in the proof of Lemma \ref{lemma:bound}, we suppress the dependency on $s$ in our notation since we are only concerned with a fixed state.
Using the notation from the proof of Lemma \ref{lemma:bound}, we have that $\pi_D(a) = \frac{S_m(a)}{m}$.
\begin{align*}
    &|S_m(a) - m \pieval(a)| \leq k - 1 \\
    \implies& \frac{|S_m(a) - m \pieval(a)|}{m} \leq  \frac{k - 1}{m} \\
    \implies& \lim_{m\rightarrow\infty} \frac{|S_m(a) - m \pieval(a)|}{m} \leq \lim_{m \rightarrow \infty} \frac{k - 1}{m} \\
    \implies& \lim_{m\rightarrow\infty} |\pi_D(a) - \pieval(a)|  \leq 0 \\
    \implies& \lim_{m\rightarrow\infty} |\pi_D(a) - \pieval(a)|  = 0 \\
    \implies& \lim_{m\rightarrow\infty} \pi_D(a) = \pieval(a).
\end{align*}

\end{proof}

\theoremmdpgeneral*
\begin{proof}
The proof is by induction.
For the base case, note that $d^0_\pi(s) = d_0(s)$ and thus $\lim_{n \rightarrow \infty} d^0_n(s) = d^0_\pi(s)$ $\forall s$ with probability $1$ by the strong law of large numbers.

For the induction step, we assume $\lim_{n \rightarrow \infty} d^t_n(s)=d^t_\pi(s)$ $\forall s$ with probability $1$ for some episode step $t < \horizon$. 
We want to show that this implies that $\lim_{n\rightarrow\infty}d^{t+1}_n(s)=d^{t+1}_\pi(s)$ with probability $1$ for all $s$.
Let $P_n$ denote the empirical state transition function, i.e., $P_n(s' | s,a) \coloneqq \frac{c_n(s,a,s')}{c_n(s,a)}$ where $c_n(s,a,s')$ is the number of times that $(s,a,s')$ occurred in $D$ and similarly for $c_n(s,a)$.
Note that $d^{t+1}_n(s) = \sum_{\tilde s} \sum_a d^t_n(\tilde s)  \pi_D(a|\tilde s) P_n(s | \tilde s, a)$ and $d^{t+1}_\pi(s) = \sum_{\tilde s} \sum_a d^t_\pi(\tilde s) \pi(a|\tilde s) P(s | \tilde s, a)$.
The former claim follows as a consequence of the finite-horizon MDP setting in which the state implicitly must depend on the current time-step. In this setting $P_n(s'|s,a)$ and $\pidata(a|s)$ are only computed with samples from a particular time-step (or pair of subsequent time-steps in the case of $P_n$). 
Then we have that:
\begin{align*}
    \lim_{n\rightarrow\infty}d^{t+1}_n(s) =& \lim_{n\rightarrow\infty}\sum_{\tilde s} \sum_a d^t_n(\tilde s) \pi_D(a|\tilde s) P_n(s | \tilde s, a )\\
    =& \sum_{\tilde{s}} \sum_a \lim_{n\rightarrow\infty} d^t_n(\tilde s) \pi_D(a | \tilde s) P_n(s | \tilde s, a) \\
    =& \sum_{\tilde{s}} \sum_a \lim_{n\rightarrow\infty} d^t_n(\tilde s) \cdot \lim_{n\rightarrow\infty} \pi_D(a|\tilde s) \cdot \lim_{n\rightarrow \infty}  P_n(s|\tilde s, a)\\
    =& \sum_{\tilde{s}} \sum_a \lim_{n\rightarrow\infty} d^t_n(\tilde s) \cdot \lim_{n\rightarrow\infty} \pi_D(a|\tilde s) \cdot \lim_{n\rightarrow \infty}  P_n(s|\tilde s, a)\\
    \overset{(a)}{=}& \sum_{\tilde{s}} \sum_a d^t_\pi(\tilde s) \pi(a | \tilde s) P(s|\tilde s, a) \\
    =& d^{t+1}_\pi(s)
\end{align*}
where \textbf{(a)} follows from the induction hypothesis, Lemma \ref{lemma:converge}, and $\lim_{n\rightarrow \infty} P_n(s'|s,a) = P(s'|s,a)$ follows from the strong law of large numbers.
This completes the proof.

\end{proof}

\theoremfaster*
\begin{proof}
This proof has two parts.
Since we only consider a particular state, we suppress the state-dependency in policies throughout this proof, i.e., we write $\pi(a)$ instead of $\pi(a|s)$.
The on-policy sampling rate is adapted from Theorem 1 of \citet{10.1093/imaiai/iaz025} which implies that, the KL-divergence between the empirical distribution of a discrete distribution and that discrete distribution itself is $O_p(\frac{1}{m})$.
In our case, $\pieval$ is the discrete distribution and $\pi_D$ is the empirical distribution.
Then we have from \citet{10.1093/imaiai/iaz025} that $2m D_\mathtt{KL}(\pidata||\pieval) \xrightarrow[]{d} \chi_{k-1}^2$ where $k$ is the number of actions.
So we have that $D_{KL}(\pidata||\pieval)= O_p(\frac{1}{m})$ from Lemma 5.3 of \citet{alexanderian2009some} since convergence in distribution implies bounded in probability.

Second, we obtain the \ROS{} rate.
Define the vectors $\tpieval \coloneqq (\pieval(1),\cdots,\pieval(k-1))$ and $\tpiD \coloneqq (\pidata(1),\cdots,\pidata(k-1))$.
We write the KL-divergence between $\pieval$ and $\pi_D$ as a function of these two vectors:
\[
D_{KL}(\tpiD||\tpieval) = \sum_{i=1}^{k-1}\tpiD(i) \log(\frac{\tpiD(i)}{\tpieval(i)}) + (1-\sum_{j=1}^{k-1}\tpiD(j))\log(\frac{1-\sum_{j=1}^{k-1}\tpiD(j)}{1-\sum_{j=1}^{k-1}\tpieval(j)}). 
\]
Let $\mathbf{g}$ denote the gradient of $D_{KL}(\tpiD||\tpieval)$ with respect to $\tilde\pi_{D}$, evaluated at point $\tilde\pi_{e}$ and $\mathbf{H}$ denotes the Hessian matrix of $D_{KL}(\pidata||\pieval)$ evaluated at point $\tilde\pi_{e}$.
Clearly, $\mathbf{g}=\mathbf{0}$ because setting $\pidata$ to $\pieval$ will minimize $D_\mathtt{KL}(\pi_D||\pieval)$.
We take the Hessian from \citet{10.1093/imaiai/iaz025}:
\[ \mathbf{H}(i,j) = \begin{cases} 
      \frac{1}{\pieval(k)} & i \neq j \\
      \frac{1}{\pieval(i)} + \frac{1}{\pieval(k)} & i = j
   \end{cases}
\]
Alternatively, we can write $\mathbf{H} = \frac{1}{\pieval(k)}\mathbf{1}\mathbf{1}^\top + \mathtt{diag}({{\tilde \pi}^{-1}}_e)$ where $\mathtt{diag}({\tilde\pi_{e}}^{-1})$ is the $(k-1 \times k-1)$ matrix with $\frac{1}{\tilde\pi_{e}}$ on its diagonal.
Using a Taylor Series expansion for $D_{\mathtt{KL}}$, we have:
$$
    m^2 D_{\mathtt{KL}}(\pidata||\pieval) = m^2 D_{\mathtt{KL}}(\pieval||\pieval) +m^2 \mathbf{g}^T(\tilde\pi_{D}-\tilde\pi_{e})+
    \underbrace{\frac{1}{2}m(\tilde\pi_{D} - \tilde\pi_{e})^T\mathbf{H} m(\tilde\pi_{D} - \tilde\pi_{e})}_\text{Quadratic Term} + Q_m
$$
where $Q_m$ consists of the higher order terms. 
Note that since $D_{\mathtt{KL}}(\pieval||\pieval)=0$, the first two terms are both 0 so we only need to bound the quadratic term and higher terms.
By  Lemma \ref{lemma:bound} every component of $m(\tilde\pi_{D} - \tilde\pi_{e})$ is bounded by $k-1$, so the quadratic term is bounded by $\frac{1}{2}\sum_{i,j}\mathbf{H}(i,j)(k-1)^2$, which is of  order $O(1)$ since $\mathbf{H}$ is a constant when evaluated at ${\tilde \pi}_e$. 
The higher order terms $Q_m$ contain higher powers of $({\tilde \pi}_D - {\tilde \pi}_e)$ multiplied by $m^2$.
Using Lemma \ref{lemma:bound}, we can bound all components of any $m(\tilde\pi_{D} - \tilde\pi_{e})$ by $k-1$ to replace each $m$ with a constant (trivially, all components of $(\tilde\pi_{D} - \tilde\pi_{e})$ can also be bounded by $k-1$) and higher-order derivatives of $D_\mathtt{KL}$ evaluated at $\pieval$ are also constant with respect to $m$.
Thus higher order terms are also $O(1)$ which gives us that $m^2 D_{\mathtt{KL}}(\pi_D||\pieval) = O(1)$ and thus that $D_{\mathtt{KL}}(\pi_D||\pieval) = O(\frac{1}{m^2})$.
Deterministic convergence to zero implies probabilistic convergence and thus, under \ROS{} action selection, $D_{\mathtt{KL}}(\pi_D||\pieval) = O_p(\frac{1}{m^2})$.
\end{proof}

\theoremerror*
\begin{proof}
First, we introduce some additional notation.
We define $\mu_n^t(s,a) \coloneqq \frac{n_t(s,a)}{n}$ as the empirical distribution of a state-action pair at time $t$, where $n_t(s,a)$ is the number of times that state $s$ and action $a$ occur jointly at time $t$ across trajectories in $\dataset$.
Let $\mu_\pieval^t(s,a)$ be the probability of state $s$ and action $a$ occurring jointly at time $t$ while following $\pieval$.
Note that $\mu_n^t(s,a) = d_n^t(s) \pidata(a|s)$ and $\mu_\pieval^t(s,a) = d_\pieval^t(s) \pieval(a|s)$.
Finally, we define $n(s_0,a_0,\dots,s_{\horizon-1},a_{\horizon-1})$ as the number of times that state-action trajectory $s_0,a_0,\dots,s_{\horizon-1},a_{\horizon-1}$ occurs in $\dataset$.

Observe that the Monte Carlo estimate can be re-written as:
\begin{align*}
    \operatorname{MC}(\data) 
    =& \frac{1}{n} \sum_{i=1}^n \sum_{t=0}^{\horizon-1} \gamma^t R(s_{i,t}, a_{i,t}) \\
    =& \frac{1}{n} \sum_{s_0} \sum_{a_0} \cdots \sum_{s_{\horizon-1}} \sum_{a_{\horizon-1}} n(s_0,a_0,\dots,s_{\horizon-1},a_{\horizon-1})\sum_{t=0}^{\horizon-1} \gamma^t R(s_{t}, a_{t}) \\
    =& \frac{1}{n} \sum_{t=0}^{\horizon-1} \gamma^t \sum_{s_0} \sum_{a_0} \cdots \sum_{s_{\horizon-1}} \sum_{a_{\horizon-1}} n(s_0,a_0,\dots,s_{\horizon-1},a_{\horizon-1}) R(s_{t}, a_{t}) \\
    =& \frac{1}{n} \sum_{t=0}^{\horizon-1} \gamma^t \sum_{s} \sum_{a} n_t(s, a) R(s,a) \\   
    =& \sum_{t=0}^{\horizon-1} \gamma^t \sum_{s} \sum_{a} \mu_n^t(s, a) R(s,a).
\end{align*}

Similarly, the true value of the evaluation policy can be re-written in terms of the state-action distribution at each time step under policy $\pieval$:
\begin{align*}
    v(\pieval) 
    =& \sum_h \Pr(h | \pieval) \sum_{t=0}^{\horizon-1} \gamma^t R(s_{h,t}, a_{h,t}) \\
    =& \sum_{s_0} \sum_{a_0} \cdots \sum_{s_{\horizon-1}} \sum_{a_{\horizon-1}} \underbrace{d_0(s_0) \pieval(a_0|s_0) \prod_{i=1}^{\horizon-1} \pieval(a_i|s_i) P(s_i|s_{i-1},a_{i-1})}_{=\Pr(h = (s_0,a_0,\dots,s_{\horizon-1},a_{\horizon-1})|\pieval)} \sum_{t=0}^{\horizon-1} \gamma^t R(s_t, a_t) \\
    =& \sum_{t=0}^{\horizon-1} \gamma^t \sum_{s_0} \sum_{a_0} \cdots \sum_{s_{\horizon-1}} \sum_{a_{\horizon-1}} d_0(s_0) \pieval(a_0|s_0) \cdots P(s_{\horizon-1}|s_{\horizon-2},a_{\horizon-2})  R(s_t, a_t) \\
    =& \sum_{t=0}^{\horizon-1} \gamma^t \sum_s \sum_a \mu_\pieval^t(s,a) R(s,a)
\end{align*}

Now, we use these alternative formulations to bound the squared error between the true value and the Monte Carlo estimate computed with a fixed data-set $\data$.
\begin{align*}
    (v(\pieval) - \operatorname{MC}(\data))^2
    &= \left(\sum_{t=0}^{\horizon-1} \gamma^t \sum_s \sum_a R(s,a) (\mu_n^t(s,a) - \mu_\pieval^t(s,a)) \right)^2 \\
    &\overset{(a)}{\leq} \sum_{t=0}^{\horizon-1} \gamma^{2t} \sum_s \sum_a \left( R(s,a) (\mu_n^t(s,a) - \mu_\pieval^t(s,a)) \right)^2 \\
    &\overset{(b)}{\leq} \sum_{t=0}^{\horizon-1} \gamma^{2t} R_\mathtt{max}^2 \sum_s \sum_a \left( \mu_n^t(s,a) - \mu_\pieval^t(s,a) \right)^2 \\
    &\overset{(c)}{\leq} \sum_{t=0}^{\horizon-1} \gamma^{2t} R_\mathtt{max}^2 \sum_s \sum_a \left | \mu_n^t(s,a) - \mu_\pieval^t(s,a) \right | \\
    &\overset{(d)}{\leq} \sum_{t=0}^{\horizon-1} \gamma^{2t} R_\mathtt{max}^2 \sqrt{ 2D_\mathtt{KL}(\mu_n^t||\mu_\pieval^t) }
\end{align*}
where (a) uses Jensen's inequality, (b) replaces $R(s,a)$ with the constant $r_\mathtt{max}$, (c) notes that $-1 \geq \mu_n^t(s,a) - \mu_\pieval^t(s,a) \leq 1$ so $|\mu_n^t(s,a) - \mu_\pieval^t(s,a)| \geq (\mu_n^t(s,a) - \mu_\pieval^t(s,a))^2$, and (d) uses Pinsker's inequality.
All that remains is to use the definition of the KL-divergence and properties of logarithms and expectations to obtain the final form of the bound:
\begin{align*}
\sum_{t=0}^{\horizon-1} \gamma^{2t} R_\mathtt{max}^2 \sqrt{2 D_\mathtt{KL}(\mu_n^t||\mu_\pieval^t) }
    &= \sum_{t=0}^{\horizon-1} \gamma^{2t} R_\mathtt{max}^2 \sqrt{2E_{S \sim d_n^t, A \sim \pi_\data}[\log \frac{d_n^t(S)\pidata(A|S)}{d_\pieval^t(S)\pieval(A|S)}]}\\
    &= \sum_{t=0}^{\horizon-1} \gamma^{2t} R_\mathtt{max}^2 \sqrt{2E_{S \sim d_n^t, A \sim \pi_\data}[\log \frac{d_n^t(S)}{d_\pieval^t(S)} + \log \frac{\pidata(A|S)}{\pieval(A|S)}]}\\
    &= \sum_{t=0}^{\horizon-1} \gamma^{2t} R_\mathtt{max}^2 \sqrt{2D_\mathtt{KL}(d_n^t||d_\pieval^t) + 2\mathbf{E}_{S \sim d_n^t} [D_\mathtt{KL}(\pi_D(\cdot|S)||\pieval(\cdot|S)]}.
\end{align*}

\end{proof}

\section{Experiment Domains}
\label{sec:env}

This appendix provides additional details on our experimental set-up.

\subsection{Extended Domain Descriptions}

\begin{figure}[H]
\centering
\subfigure[Bandit]{\includegraphics[width=0.32\textwidth]{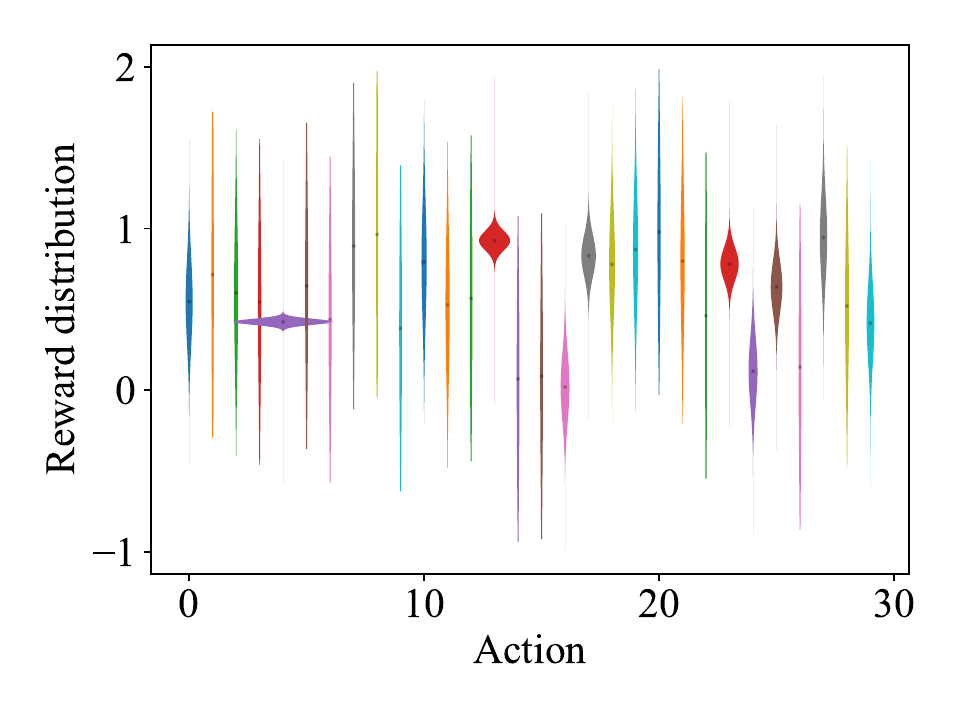}\label{fig:env_mb}}
\subfigure[GridWorld]{\includegraphics[width=0.32\textwidth]{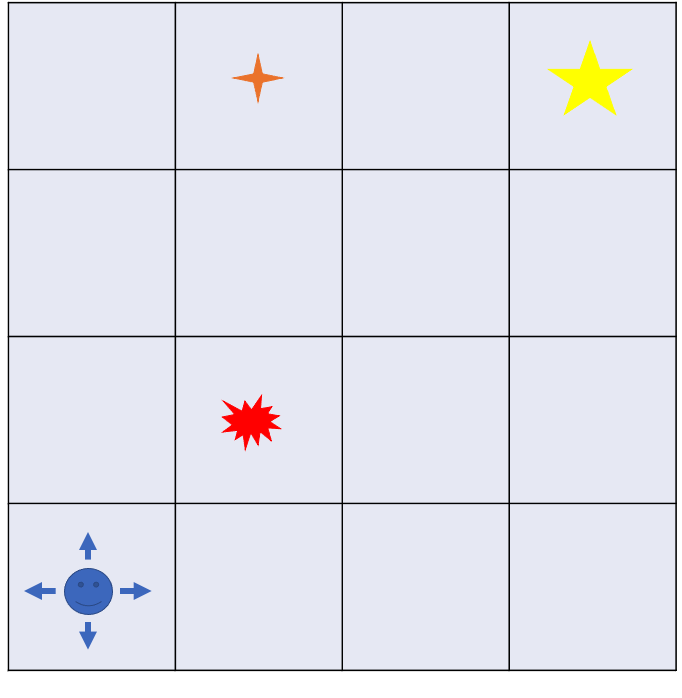}\label{fig:env_gw}}
\subfigure[CartPole]{\includegraphics[width=0.32\textwidth]{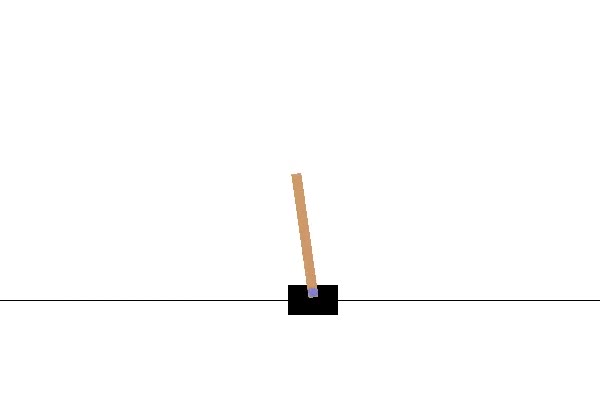}\label{fig:env_cp}}
\caption{Experimental Domains. Figure \ref{fig:env_mb} shows the distribution of rewards from each action in a $30$-armed bandit problem, with black dots indicating the mean rewards and shading indicating the scale of the distribution. Figure \ref{fig:env_gw} shows a $4 \times 4$ GridWorld with the agent starting from the bottom-left corner and the goal state in the top-right corner. Figure \ref{fig:env_cp} shows the CartPole environment in which the goal is to keep the pole from falling over. ContinuousCartPole is the same as CartPole except with a modified action space.}
\label{fig:domains}
\end{figure}

This section describes the domains used in our empirical evaluation.
Figure \ref{fig:domains} illustrates each domain.

Our first domain, Bandit, is a 30-armed bandit problem modelled after the 10-armed bandit in \citet[Chapter 2]{sutton1998reinforcement}.
This domain has a single state, 30 actions, and episodes terminate after the first action is taken.
The reward following each action is normally distributed with a randomly generated mean and scale parameter.
The reward distribution parameters are sampled from a uniform distribution on $[0,1]$ at the start of each experimental trial.

Our second domain, GridWorld, is a discrete state and action domain that has been used in prior policy evaluation research (e.g., \citep{thomas2016data-efficient,farajtabar2018more}).
The domain (shown in Figure~\ref{fig:env_gw}) has $4 \times 4$ states. The agent starts from $(0,0)$ and has the action space \{left, right, up, down\}. The reward is $-1$ in all non-terminal states except $(1,1)$ where the reward is $-10$ and $(1,3)$ where the reward is $+1$. The agent receives $+10$ in terminal state $(3,3)$. The maximum number of steps is $100$ and $\gamma=1$ in this domain.

Our last two domains are based on the CartPole problem from OpenAI Gym~\citep{gym}.
In CartPole, the agent tries to balance a pole, mounted on a cart, by moving the cart left and right.
States are given as vectors that describe the cart position and velocity and pole angle and angular velocity.
Our third domain is the standard variant in which the agent controls the cart by choosing either a constant leftward force of $-1$, no force, or a constant rightward force of $+1$.
Our fourth domain is a continuous control variant in which the agent can control the exact force ranging from $-1$ to $1$.
In either case, the agent receives a reward of $+1$ for each time-step until termination when the pole falls or the cart moves out of bounds. The maximum number of steps is $200$ and $\gamma=0.99$ in this domain.

\subsection{Creation of Pre-trained Evaluation Policy}\label{app:pi_e}

Each domain requires creation of an evaluation policy to serve as $\pieval$.
In the three domains with a discrete action space we use softmax policies of the form:
\begin{equation}
    \pitheta(a|s) \propto \frac{e^{w_a^\top\mathbf{\phi}(s)}}{ \sum_{b \in \mathcal{A}}e^{w_b^\top\mathbf{\phi}(s)}} %
    \label{eq:policy_dis}
\end{equation}
where $\mathbf{\phi}$ is a one-hot encoding operator for domains with discrete state space (Bandit and GridWorld) and a feed-forward neural network for continuous state spaces (CartPole).
For ContinuousCartPole,
\begin{equation}
    \pitheta(a|s) \coloneqq \mathcal{N} \left(a;  \mathbf{w}_\mu^\top\mathbf{\phi}(s), \left( \mathbf{w}_\sigma^\top\mathbf{\phi}(s) \right)^2 \right),
    \label{eq:policy_con}
\end{equation}
where $\mathbf{\phi}$ is a function of the state given by a feed-forward neural network.
The policy parameters, $\btheta$, denote all policy parameters. For Bandit and GridWorld, this is just the vectors $\mathbf{w_a}$ and for CartPole and ContinuousCartPole, $\btheta$ also includes the neural network weights and biases.
When $\mathbf{\phi}$ is a neural network, it is constructed with one batch normalization layer as the first layer, followed by two hidden layers, both of which have 64 hidden states and use ReLU as the activation function.
We use PyTorch for neural network implementations \citep{NEURIPS2019_9015} and NumPy for linear algebra computations \citep{harris2020array}.

For all domains, we use REINFORCE~\citep{reinforce} to train the policy model, and choose a policy snapshot during training as the evaluation policy, which has higher returns than the uniformly random policy, but is still far from convergence. To obtain $v(\pieval)$, we use on-policy sampling to collect $10^6$ trajectories and compute the Monte Carlo estimate of $v(\pieval)$.

\subsection{Off-policy Data for \textbf{With Initial Data} Experiments}
\label{sec:opd}

To create an initial data set of slightly off-policy data for each domain, we create the behavior policy $\pib$ based on the evaluation policy $\pi_e$.
For domains with discrete action space, the off-policy behavior policy is built as:
$$ \pib(a|s) = (1-\delta)\pieval(a|s) + \delta \frac{1}{\lvert \mathcal{A} \rvert} $$
where $\delta \in (0,1]$ controls the probability of randomly choosing an action from the action space, and otherwise sampling an action from the evaluation policy $\pieval$.
For ContinuousCartPole, the behavior policy is built as:
$$ \pib(a|s) = \mathcal{N} \left(a; \mu_e(s), \left(\left( 1 + \delta \right)\sigma_e(s)\right)^2 \right) $$ where $\mu_e(s)$ and $\sigma_e(s)$ are the output mean and standard deviation of the evaluation policy $\pi_e$ in state $s$, and $\delta$ (for $\delta > 0$) increases the standard deviation. 
In all domains, we use $\delta=0.1$ and collect 100 trajectories to create the initial off-policy data set.

\section{Measuring Sampling Error}\label{app:measure-sampling-error}

One of our central claims is that \ROS{} reduces sampling error and reducing sampling error corresponds to a reduction in \MSE{} for policy evaluation.
To evaluate this claim, we must define a metric for measuring sampling error.
In this appendix, we describe two possible metrics and demonstrate how these metrics change as \OS{} and \ROS{} collect data for policy evaluation.

\subsection{KL-divergence of Data Collection}
\label{sec:kl}

Our first metric (also used in the main paper), is to measure sampling error in the collected data $D$ by using KL-divergence of the empirical policy $\pi_D$ and the evaluation policy $\pi_e$.
The KL-divergence between $\pi_D$ and $\pieval$ in a particular state $s$ is defined as:
\[
D_\mathtt{KL}(\pi_D, \pieval) \coloneqq \textbf{E}\biggl[\log \frac{\pi_D(A|s)}{\pieval(A|s)} \biggm| A \sim \pi_D \biggr].
\]
To obtain an explicit representation of $\pi_D$, we use a parametric estimate by maximizing the log-likelihood function over a parametric policy class.
Specifically, we use:
\[
\hat{\btheta} = \arg\max_{\btheta'} \sum_{(s,a) \in \dataset_1} \log \pi_{\btheta'} (a|s)
\]
for a policy class parameterized by $\btheta$.
We then use $\pi_{\hat{\btheta}}$ in place of $\pi_D$ when computing the KL-divergence.
Our final sampling error metric for a data set $\dataset_1$ is:
\[
D_\mathtt{KL}(\pieval, D) \coloneqq \sum_{(s,a) \in \dataset_1} \log \pi_{\hat{\btheta'}}(a|s) - \log \pieval(a|s).
\]

\begin{figure}[H]
\centering
\subfigure[Bandit]{\includegraphics[height=7.3em]{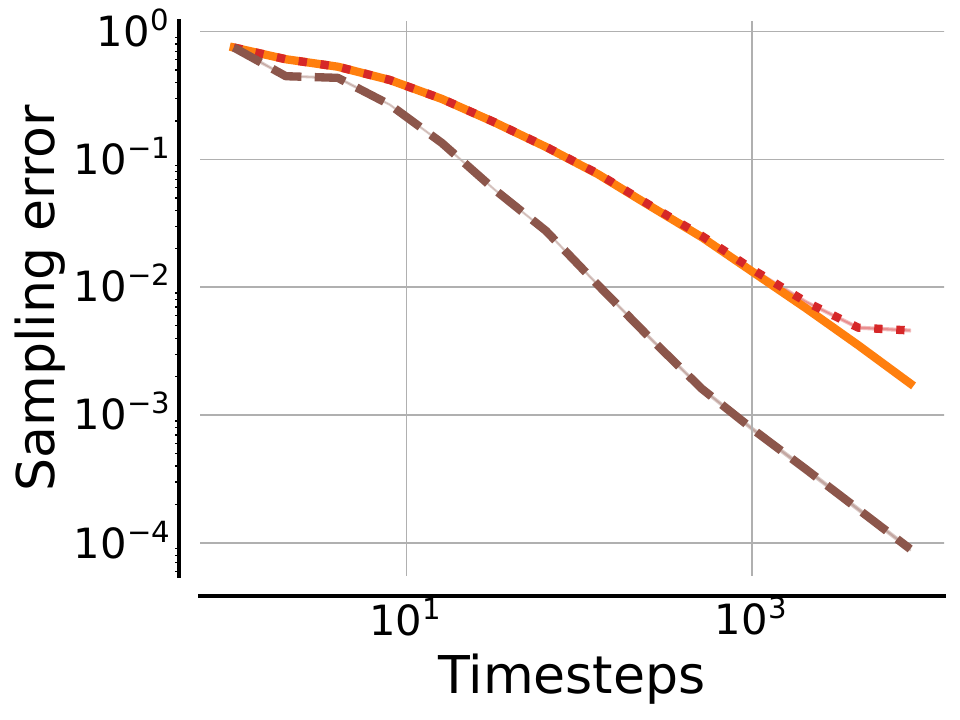}}
\subfigure[GridWorld]{\includegraphics[height=7.3em]{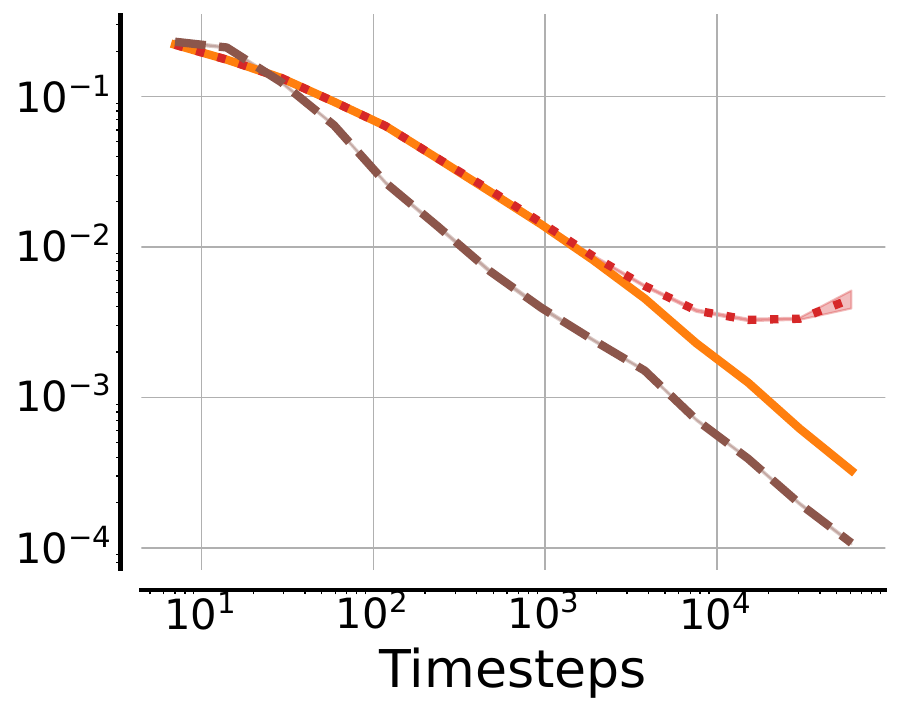}}
\subfigure[CartPole]{\includegraphics[height=7.3em]{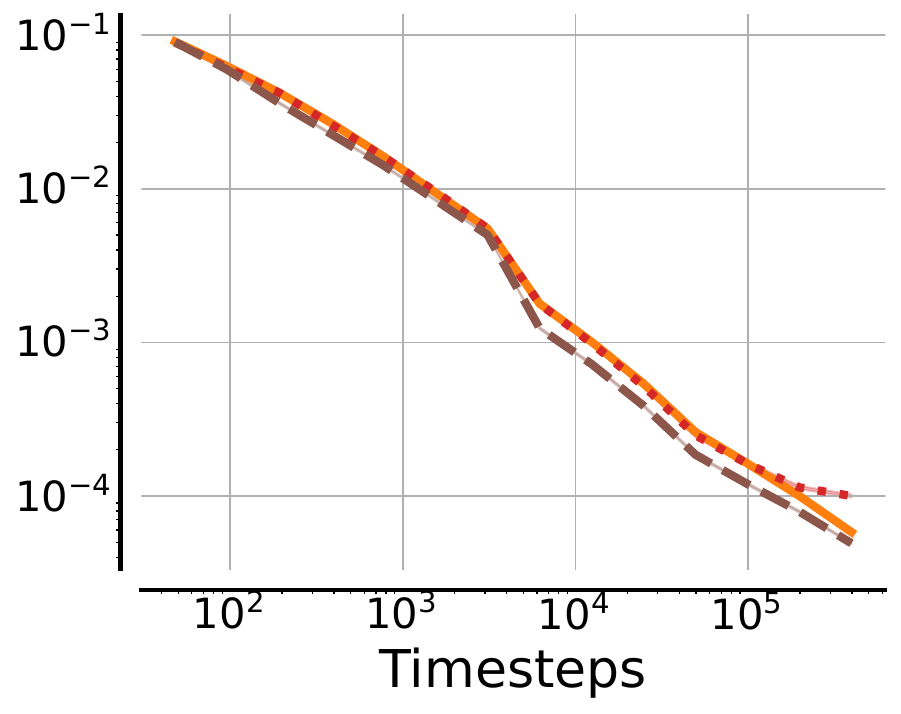}}
\subfigure[ContinuousCartPole]{\includegraphics[height=7.3em]{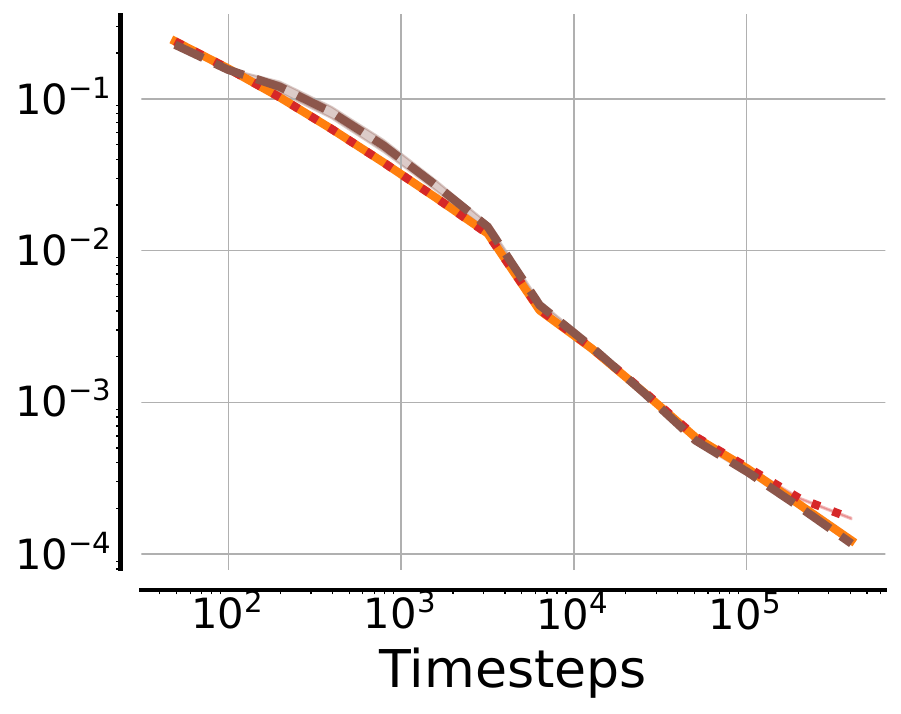}}
\subfigure{\includegraphics[width=0.31\textwidth]{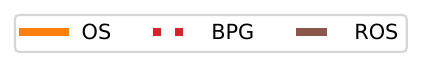}}
\caption{Sampling error (KL) of data collection \textbf{without initial data}. Each strategy is followed to collect data with $2^{13}\overline{T}$ steps, and all results are averaged over $200$ trials with shading indicating one standard error intervals. Axes in these figures are log-scaled.}
\label{fig:kl}
\end{figure}

\begin{figure}[H]
\centering
\subfigure[Bandit]{\includegraphics[height=7.3em]{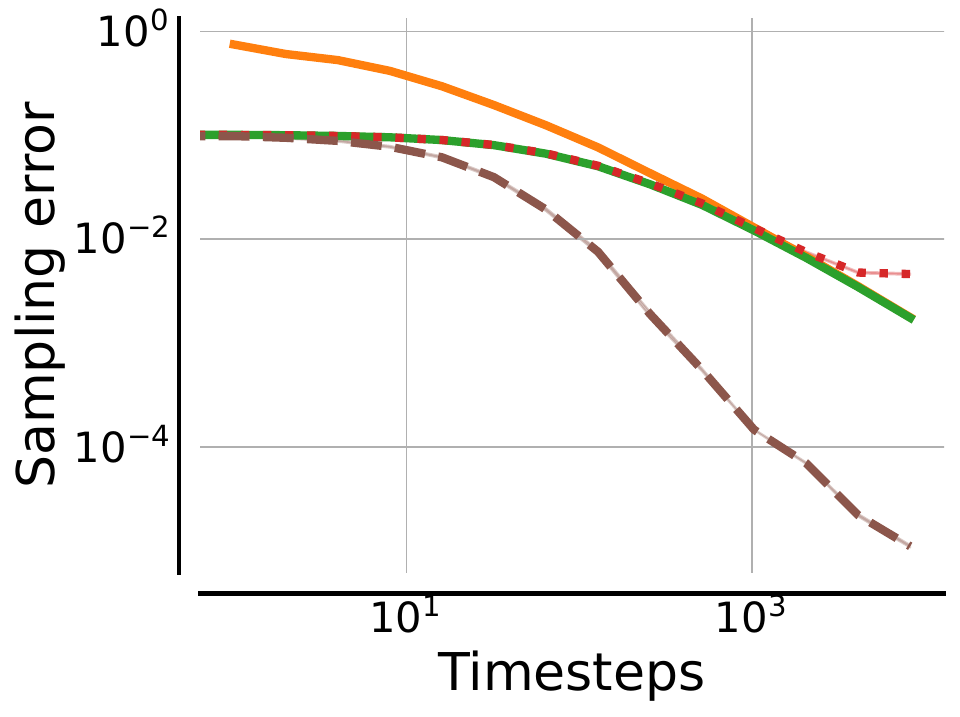}}
\subfigure[GridWorld]{\includegraphics[height=7.3em]{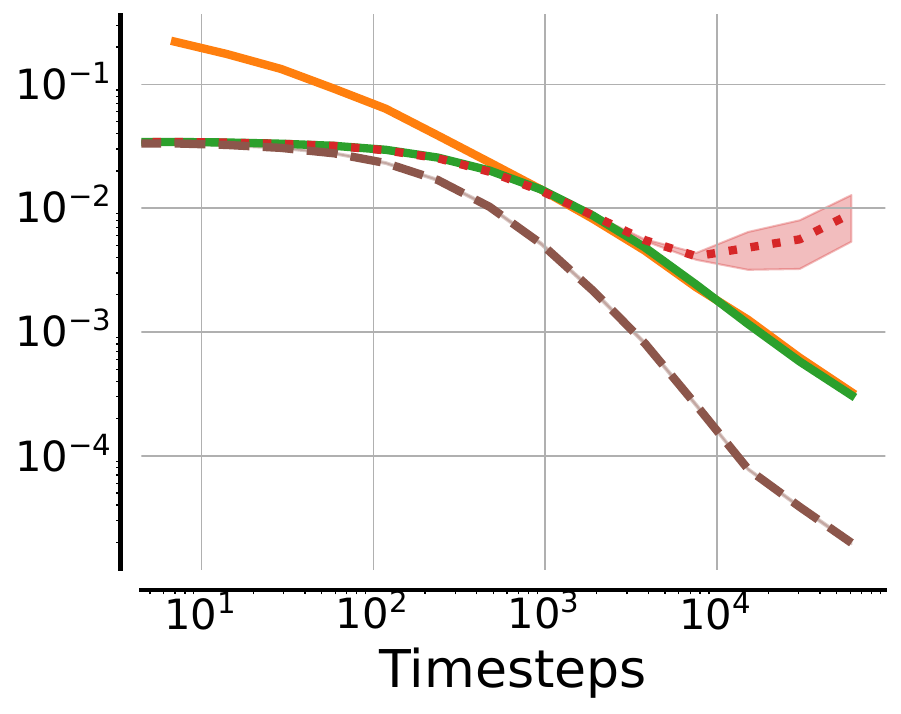}}
\subfigure[CartPole]{\includegraphics[height=7.3em]{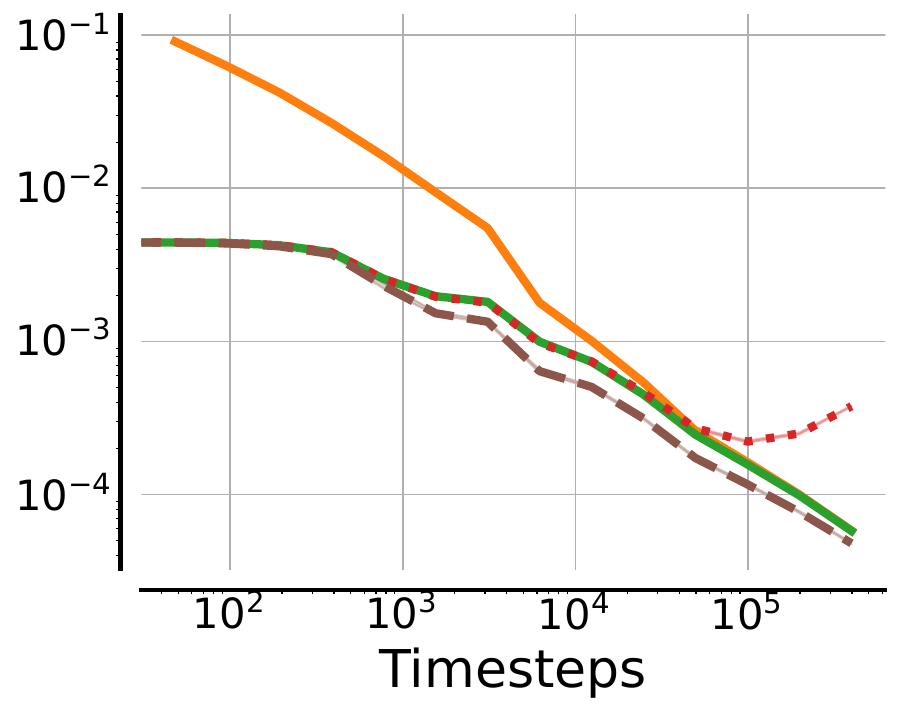}}
\subfigure[ContinuousCartPole]{\includegraphics[height=7.3em]{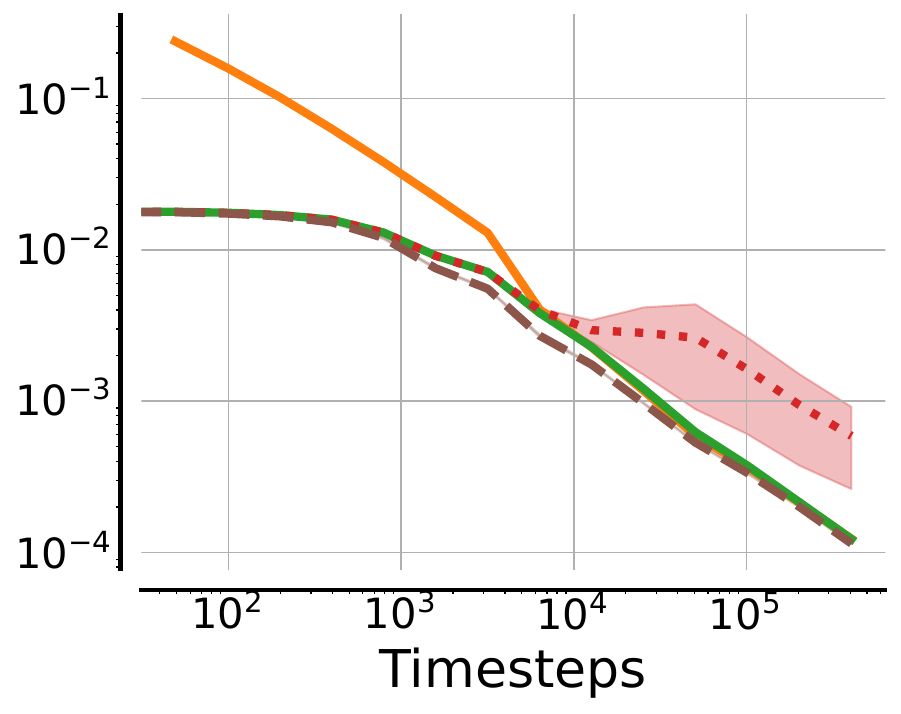}}
\subfigure{\includegraphics[width=0.66\textwidth]{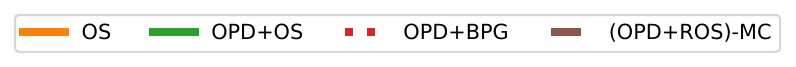}}

\caption{Sampling error (KL) of data collection \textbf{with initial data}. Steps, axes, trials and intervals are the same as Figure~\ref{fig:kl}.}
\label{fig:comb_kl}
\end{figure}

The sampling error curves of data collection \textbf{without} and \textbf{with initial data} are shown in Figure~\ref{fig:kl} and~\ref{fig:comb_kl}, respectively.
We observe from these experiments that \ROS{} can collect data with lower sampling error than \OS{} and \BPG{}.

\subsection{l1-Norm of the Log-likelihood Gradient}
\label{sec:grad}

\begin{figure}[H]
\centering
\subfigure[Bandit]{\includegraphics[height=7.3em]{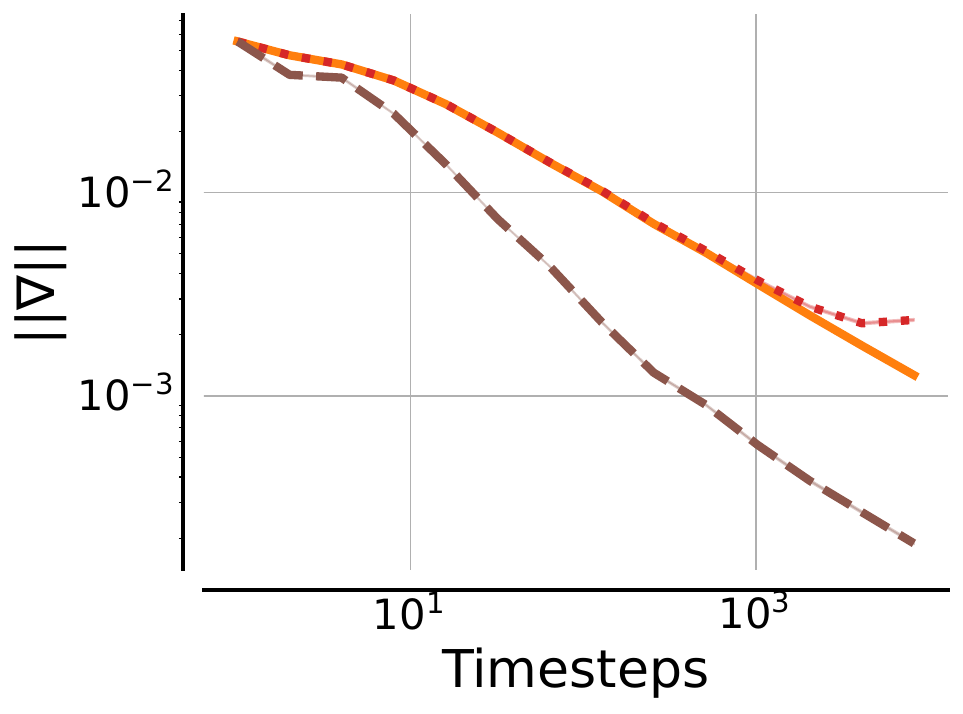}}
\subfigure[GridWorld]{\includegraphics[height=7.3em]{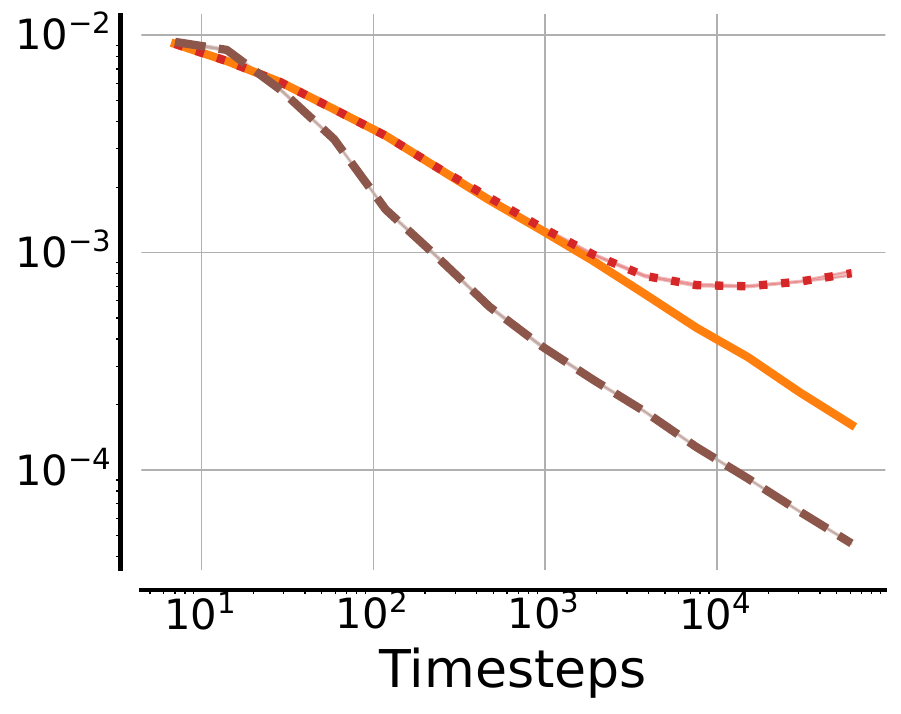}}
\subfigure[CartPole]{\includegraphics[height=7.3em]{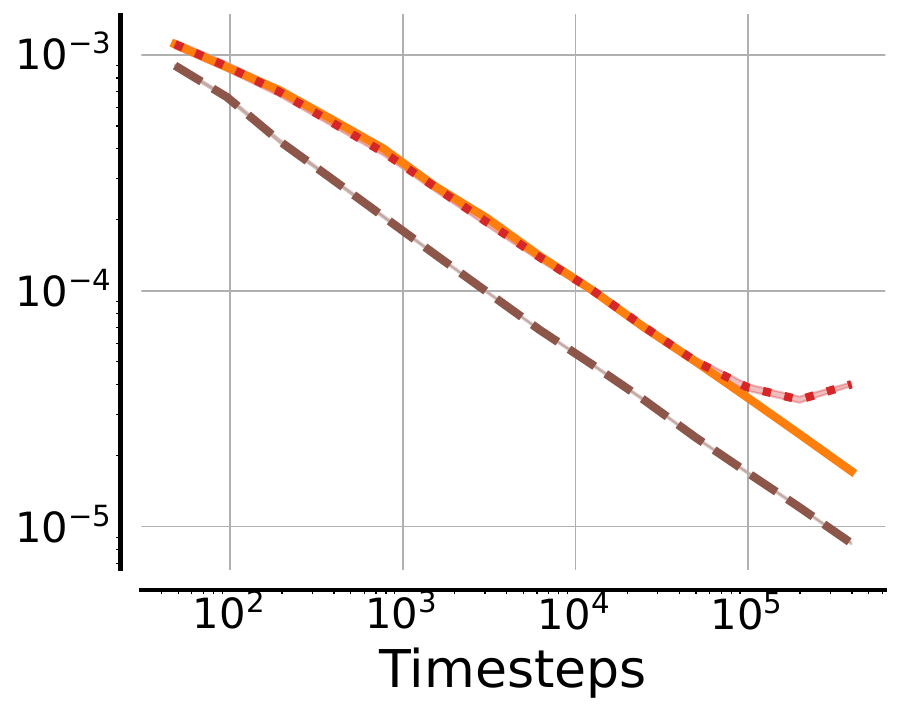}}
\subfigure[ContinuousCartPole]{\includegraphics[height=7.3em]{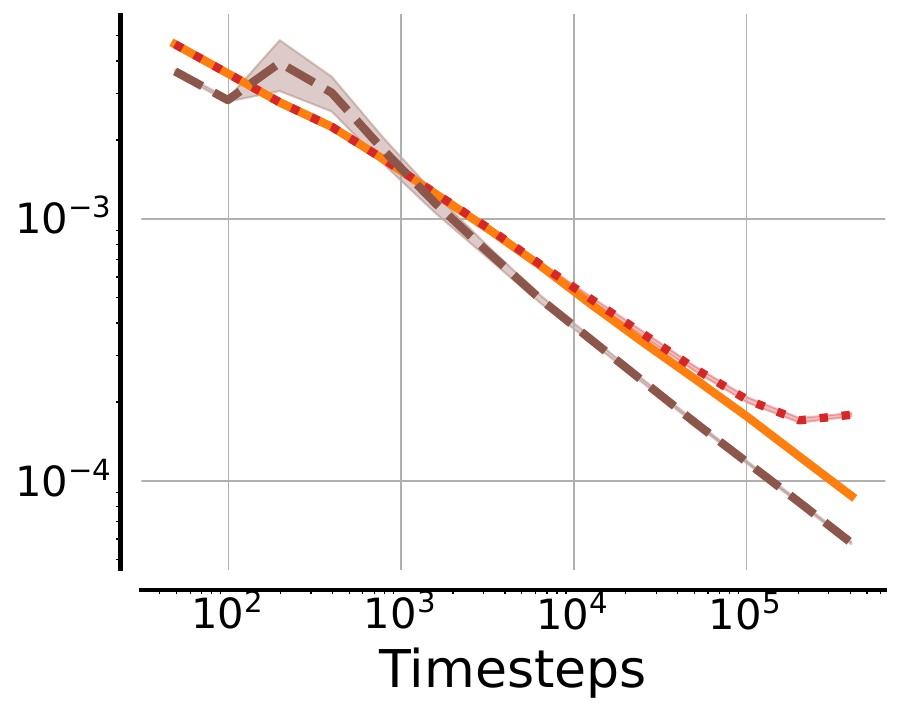}}
\subfigure{\includegraphics[width=0.31\textwidth]{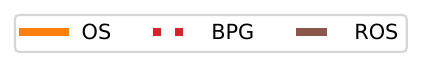}}

\caption{Log-likelihood gradient of the data collection \textbf{without initial data}. Steps, axes, trials and intervals are the same as Figure~\ref{fig:kl}.}
\label{fig:grad}
\end{figure}

\begin{figure}[H]
\centering
\subfigure[Bandit]{\includegraphics[height=7.3em]{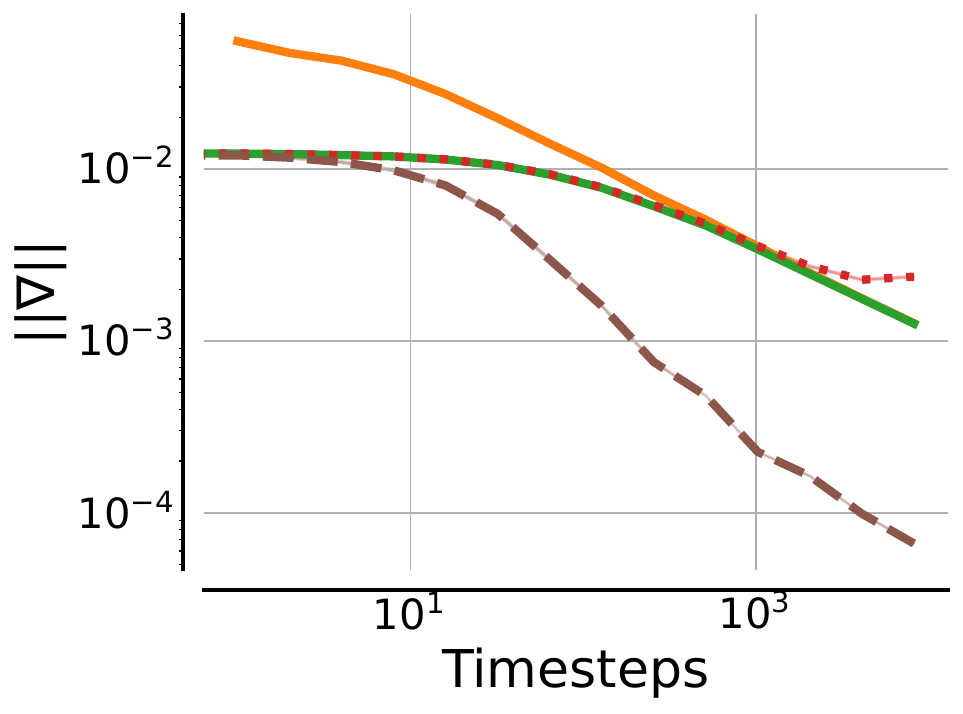}}
\subfigure[GridWorld]{\includegraphics[height=7.3em]{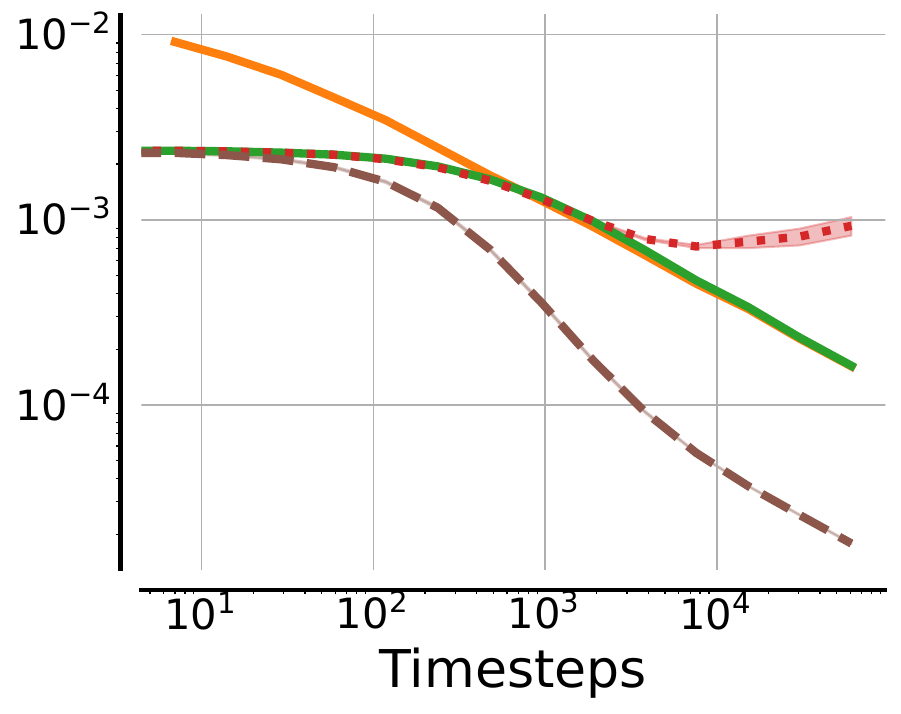}}
\subfigure[CartPole]{\includegraphics[height=7.3em]{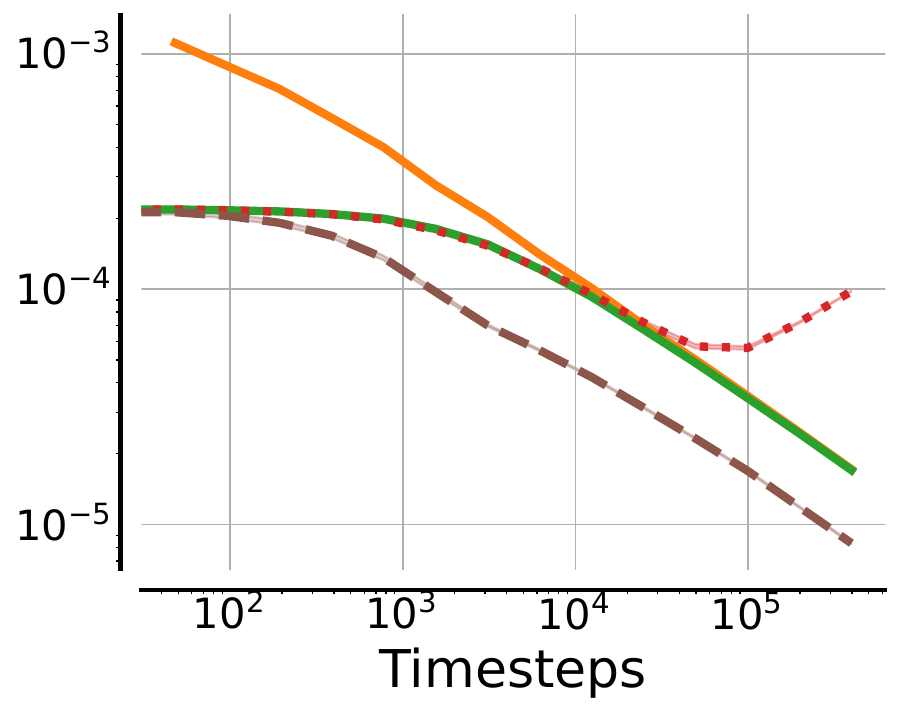}}
\subfigure[ContinuousCartPole]{\includegraphics[height=7.3em]{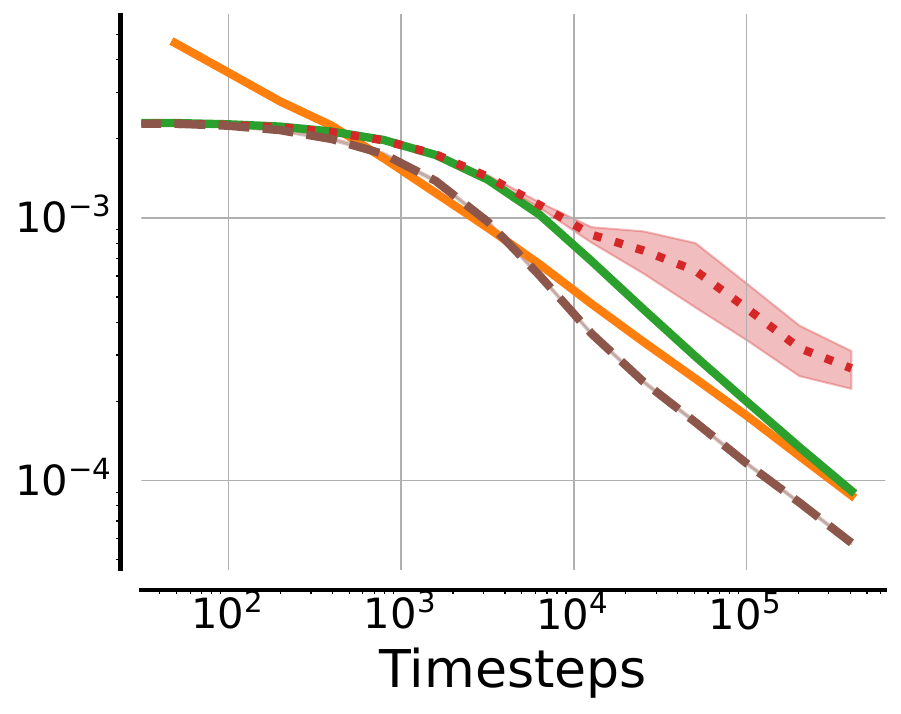}}
\subfigure{\includegraphics[width=0.66\textwidth]{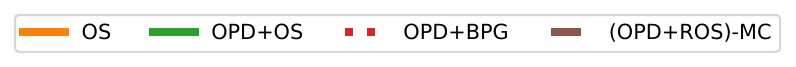}}

\caption{Log-likelihood gradient of the data collection with initial data. Steps, axes, trials and intervals are the same as Figure~\ref{fig:kl}.}
\label{fig:comb_grad}
\end{figure}

In this section, we propose an alternative sampling error measure based on the norm of the log-likelihood gradient evaluated at $\pieval$.
If $\pieval$ is the maximum likelihood policy under $\dataset$ then the norm of the gradient evaluated at $\btheta_e$ will be zero when sampling error is zero.
A non-zero norm indicates the parameters must change from $\btheta_e$ to maximize the log-likelihood under the observed data.
We show empirically that this measure of sampling error roughly corresponds to using the KL-divergence by computing the gradient norm of log-likelihood with respect to the data collection without and with initial data, shown in Figure~\ref{fig:grad} and~\ref{fig:comb_grad}, respectively.
These curves are generally consistent with their corresponding sampling error curves in Appendix~\ref{sec:kl}.

\section{Hyper-parameter Configurations}
\label{app:hparams}
This appendix gives the hyper-parameter settings for the policy evaluation experiments.
Settings for the \textbf{without initial data} experiments are given in Table~\ref{tab:hp} and settings for the \textbf{with initial data} experiments are given in Table~\ref{tab:comb_hp}.
For \BPG{}, $k$ denotes the batch-size and $\alpha$ the step-size.

\begin{table}[H]
\centering
\begin{tabular}{cccc}
\hline
Domain & \BPG{} - $k$ & \BPG{} - $\alpha$ & \ROS{} - $\alpha$  \\
\hline
Bandit & 10 & 0.01 & 10000.0\\
GridWorld & 10 & 0.01 & 1000.0\\
CartPole & 10 & 5e-05 & 10.0\\
CartPoleContinuous & 10 & 1e-06 & 0.1 \\
\hline
\end{tabular}
\caption{Hyper-parameters for experiments \textbf{without initial data}.}
\label{tab:hp}
\end{table}

\begin{table}[H]
\centering
\begin{tabular}{cccc}
\hline
Domain & \BPG{} - $k$ & \BPG{} - $\alpha$ & \ROS{} - $\alpha$ \\
\hline
Bandit & 10 & 0.01 & 10000.0  \\
GridWorld & 10 & 0.01 & 10000.0  \\
CartPole & 10 & 5e-05 & 10.0  \\
CartPoleContinuous & 10 & 1e-06 & 0.1  \\
\hline
\end{tabular}
\caption{Hyper-parameters for experiments \textbf{with initial data}.}
\label{tab:comb_hp}
\end{table}

\section{Numerical Results of Policy Evaluation with Initial Data}
\label{app:stat}

This appendix provides the numerical final values for the \MSE{} of policy evaluation with each data collection method.
Table~\ref{tab:mse} gives these values for the \textbf{without initial data} experiments and Table \ref{tab:comb_mse} gives these values for the \textbf{with initial data} experiments.
These tables provide the numerical value corresponding to the final \MSE{} value for each method-domain pair shown in Figures \ref{fig:mse} and \ref{fig:comb_mse}.

\begin{table}[H]
\centering
\scriptsize
\begin{tabular}{ccccc}
\hline
Policy Evaluation & Bandit & GridWorld & CartPole & CartPoleContinuous\\
\hline
OS - MC & 2.06e-04 $\pm$ 2.12e-05 & 2.68e-04 $\pm$ 2.42e-05 & 2.61e-05 $\pm$ 2.40e-06 & 4.44e-05 $\pm$ 4.01e-06\\
\BPG{} - OIS & 9.52e-05 $\pm$ 8.83e-06 & 2.81e-04 $\pm$ 2.63e-05 & 1.20e-05 $\pm$ 1.11e-06 & 3.62e-05 $\pm$ 3.54e-06\\
\ROS{} - MC & 6.17e-05 $\pm$ 5.77e-06 & 1.36e-05 $\pm$ 1.33e-06 & 1.01e-05 $\pm$ 9.79e-07 & 3.29e-05 $\pm$ 3.01e-06\\
\hline
\end{tabular}
\caption{Final \MSE{} of policy evaluation \textbf{without initial data}. These results give the \MSE{} for policy evaluation at the end of data collection, averaged over 200 trials $\pm$ one standard error.}
\label{tab:mse}
\end{table}

\begin{table}[H]
\centering
\scriptsize
\begin{tabular}{ccccc}
\hline
Policy Evaluation & Bandit & GridWorld & CartPole & CartPoleContinuous\\
\hline
OS - MC & 1.62e-04 $\pm$ 1.45e-05 & 2.68e-04 $\pm$ 2.42e-05 & 2.61e-05 $\pm$ 2.40e-06 & 4.44e-05 $\pm$ 4.01e-06\\
(OPD + OS) - MC & 1.48e-04 $\pm$ 1.41e-05 & 2.80e-04 $\pm$ 2.82e-05 & 2.69e-05 $\pm$ 2.70e-06 & 4.27e-05 $\pm$ 3.72e-06\\
(OPD + OS) - (WIS + MC) &
1.56e-04 $\pm$ 1.70e-05 & 2.65e-04 $\pm$ 2.40e-05 & 2.59e-05 $\pm$ 2.51e-06 & 6.02e-05 $\pm$ 8.51e-06\\
(OPD + \BPG{}) - OIS &
6.86e-05 $\pm$ 7.46e-06 & 9.20e-04 $\pm$ 6.27e-04 & 1.52e-05 $\pm$ 1.53e-06 & 1.12e-03 $\pm$ 9.58e-04\\
(OPD + \ROS{}) - MC & \textbf{4.78e-05 $\pm$ 4.81e-06} & 2.35e-06 $\pm$ 2.42e-07 & 9.60e-06 $\pm$ 9.35e-07 & 3.27e-05 $\pm$ 3.21e-06\\
\hline
\end{tabular}
\caption{Final \MSE{} of policy evaluation \textbf{with initial data}. These results give the \MSE{} for policy evaluation at the end of data collection, averaged over 200 trials $\pm$ one standard error.}
\label{tab:comb_mse}
\end{table}

\section{Median and Inter-quartile Range of Policy Evaluation}
\label{app:median}

In this appendix, we present the computed median and interquartile range for the squared error of policy evaluation both \textbf{with} and \textbf{without initial data} across all four domains.
In the main paper we present the mean squared error and standard error as is typical in the policy evaluation literature.
Here, we include the median and interquartile ranges as they are more robust statistics.
We give these results for completeness; qualitatively, they leave the conclusions from the main paper unchanged.

\begin{figure}[H]
\centering
\subfigure[Bandit]{\includegraphics[height=7.3em]{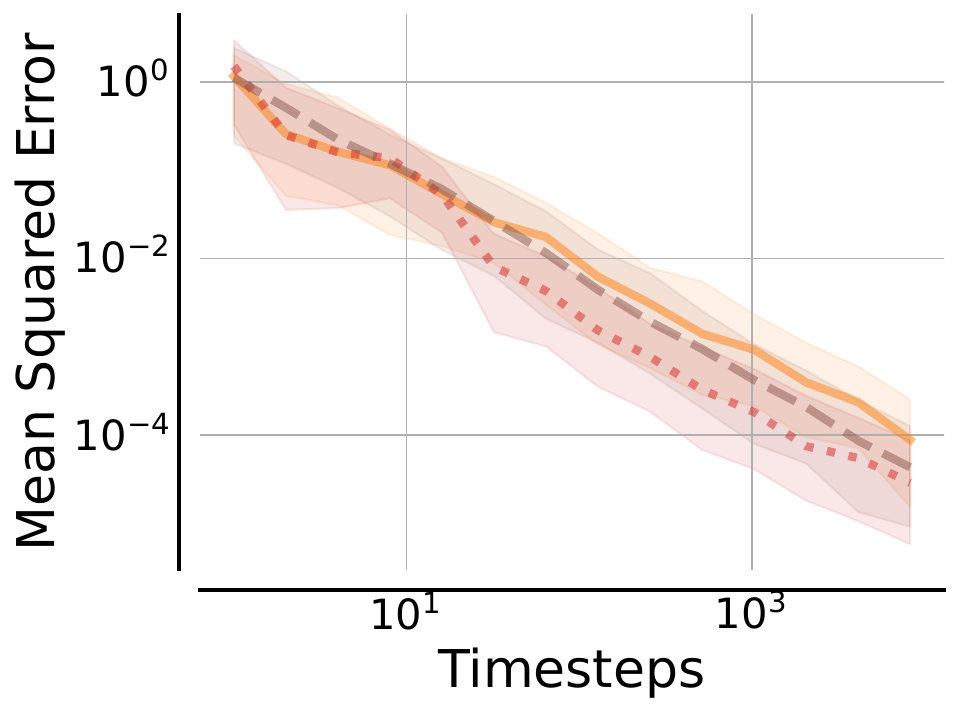}}
\subfigure[GridWorld]{\includegraphics[height=7.3em]{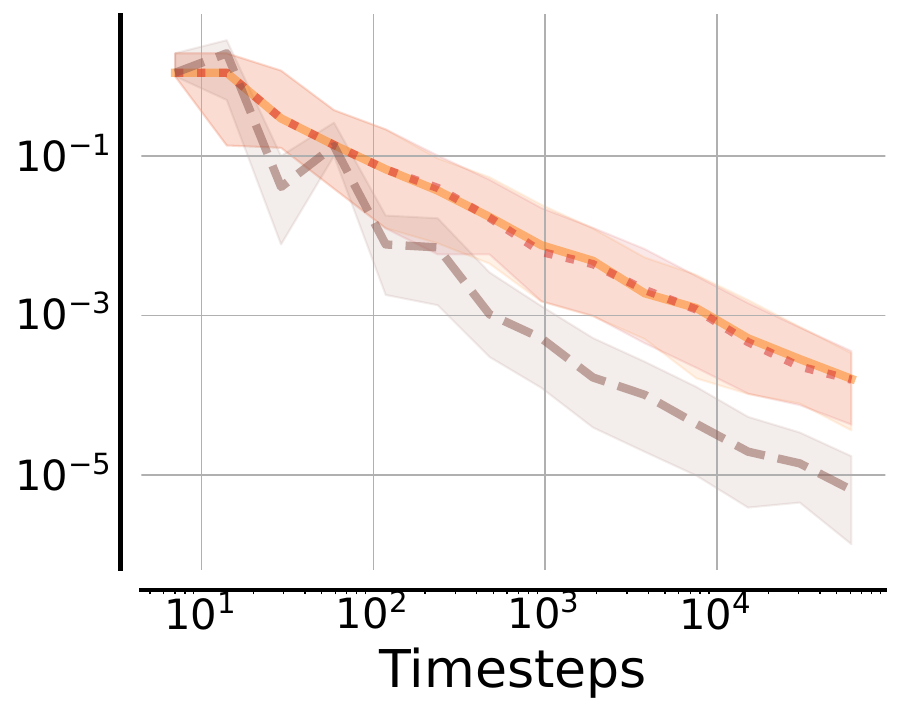}}
\subfigure[CartPole]{\includegraphics[height=7.3em]{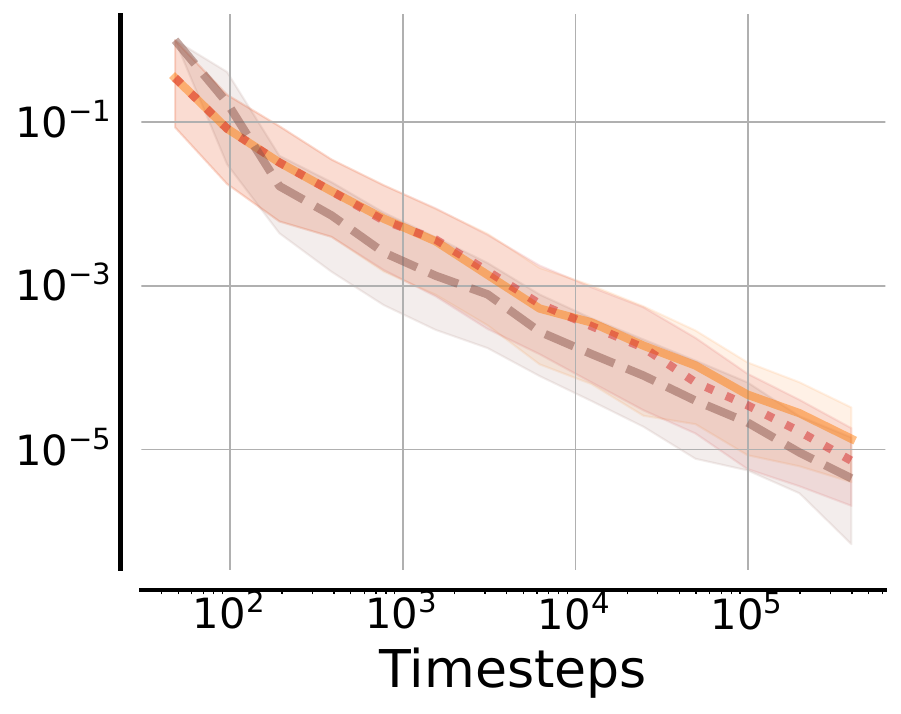}}
\subfigure[ContinuousCartPole]{\includegraphics[height=7.3em]{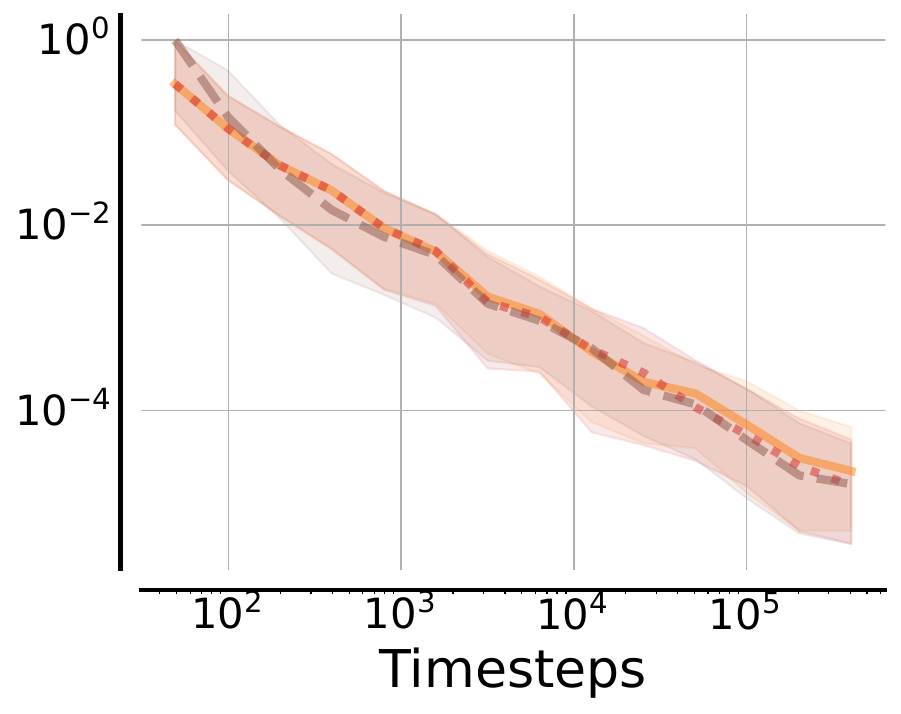}}
\subfigure{\includegraphics[width=0.31\textwidth,clip=true]{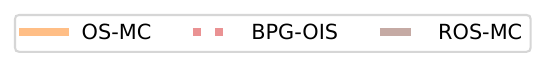}}
\caption{Median and inter-quartile range of squared error (SE) of policy evaluation \textbf{without initial data}. The lines in these figures denote the median of squared error over 200 trials, and the shading indicates the interquartile range. Axes in these figures are log-scaled.}
\label{fig:mq}
\end{figure}

\begin{figure}[H]
\centering
\subfigure[Bandit]{\includegraphics[height=7.3em]{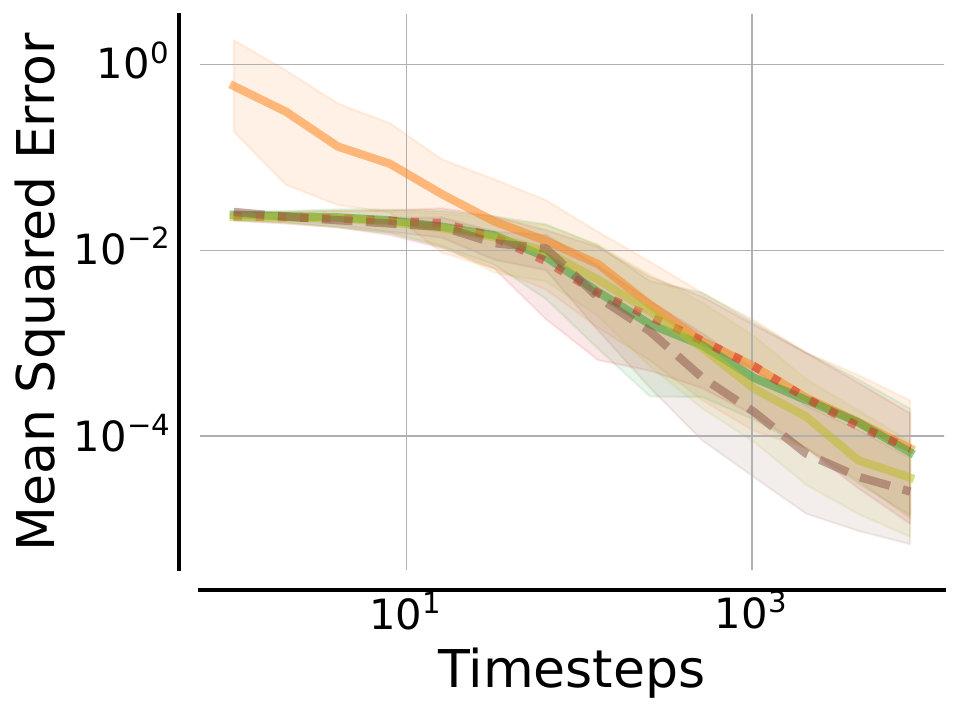}}
\subfigure[GridWorld]{\includegraphics[height=7.3em]{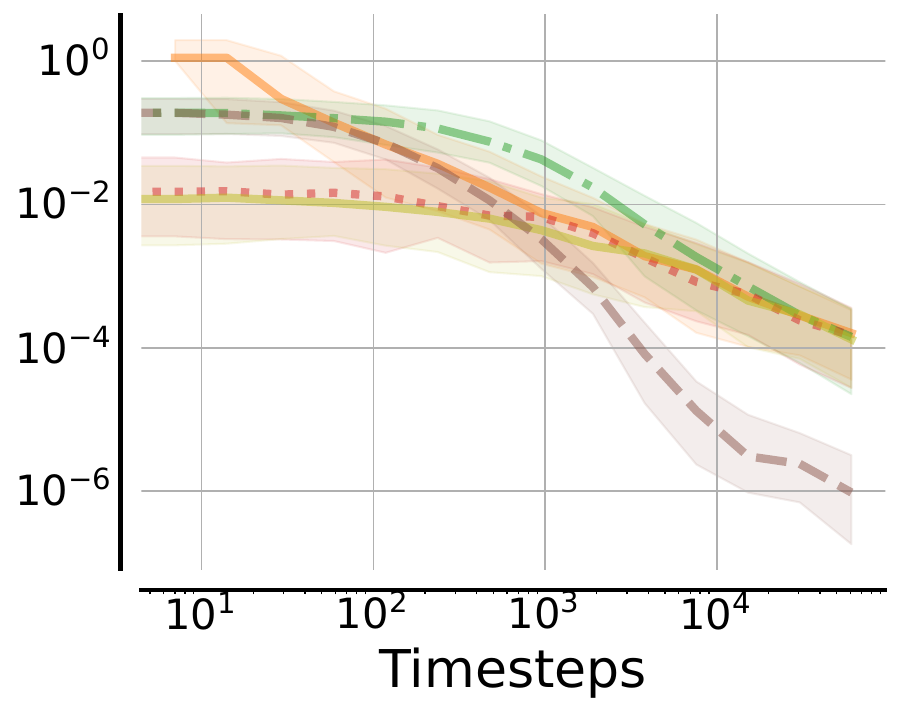}}
\subfigure[CartPole]{\includegraphics[height=7.3em]{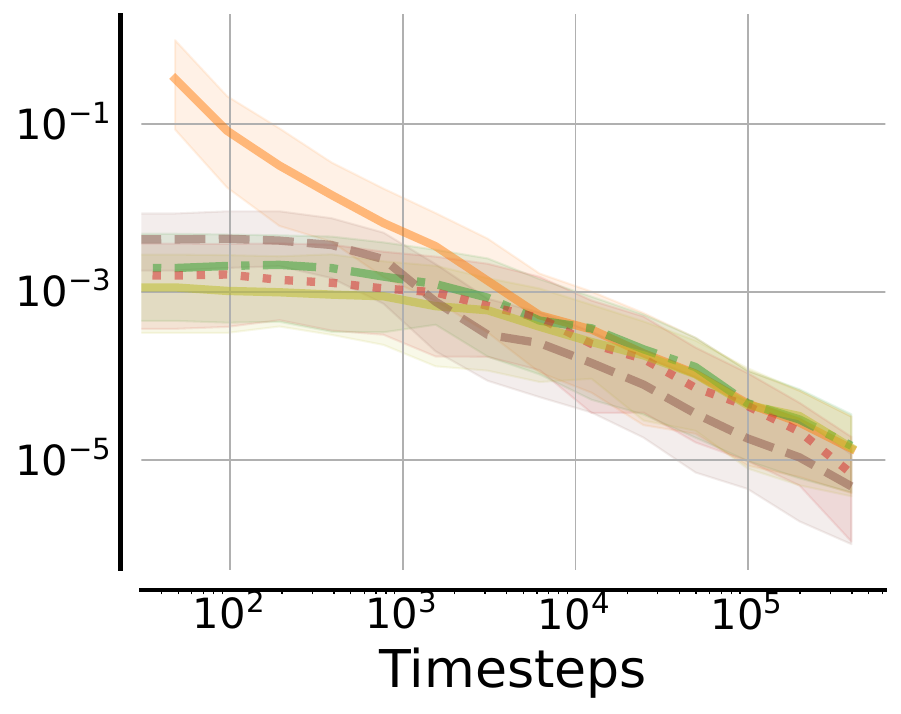}}
\subfigure[ContinuousCartPole]{\includegraphics[height=7.3em]{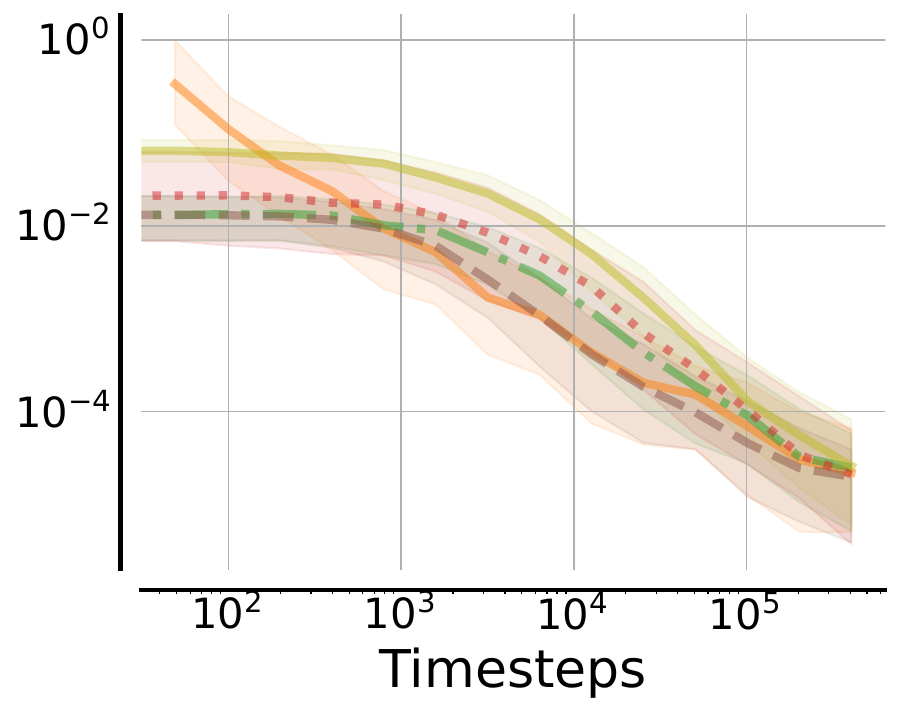}}
\subfigure{\includegraphics[width=0.66\textwidth]{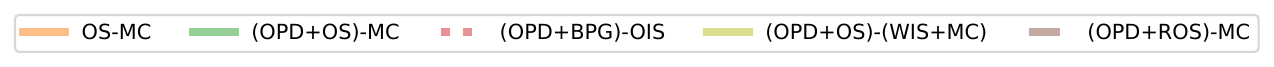}}
\vspace{-5pt}
\caption{Median and inter-quartile range of squared error (SE) of policy evaluation \textbf{with initial data}. Axes, trials and intervals are the same as Figure~\ref{fig:mq}.}
\label{fig:comb_mq}
\end{figure}

Figure \ref{fig:mq} shows the median and interquartile ranges for policy evaluation \textbf{without initial data}.
Figure \ref{fig:comb_mq} gives the same for policy evaluation \textbf{with initial data}.
From these figures, we can observe that all data collection methords have a similar inter-quartile range and \ROS{} lowers the median of the squared error compared to \OS{}.

\section{Environment and Policy Sensitivity in GridWorld}
\label{sec:imp_gw}

\begin{figure}[H]
\subfigure[Environment Settings]{\includegraphics[width=0.45\textwidth]{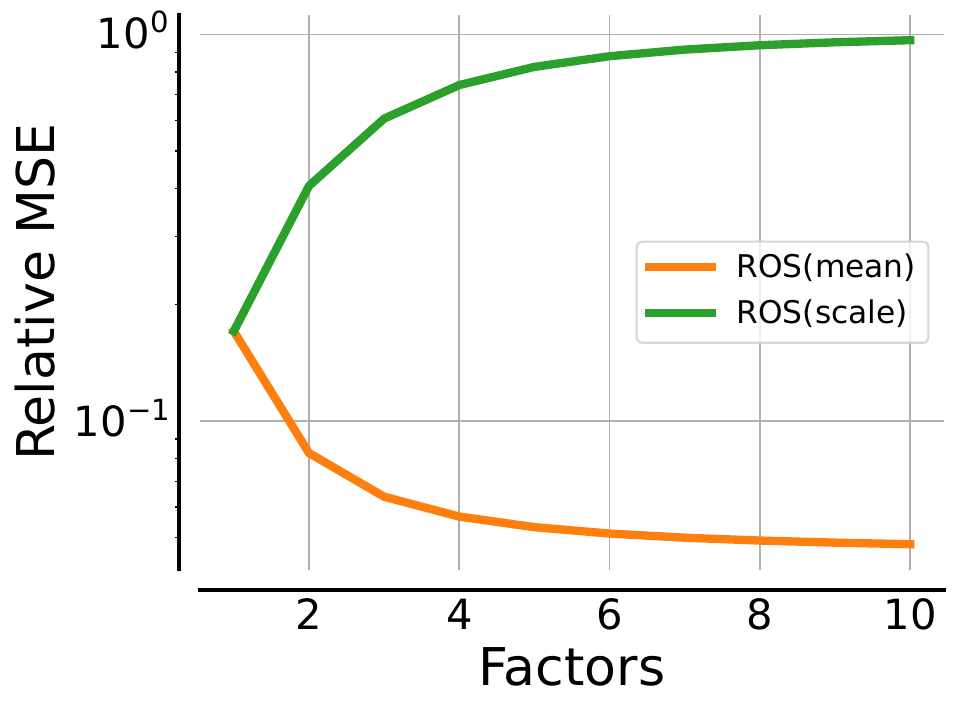}\label{fig:imp_env_gw}}
\subfigure[Policy Settings]{\includegraphics[width=0.45\textwidth]{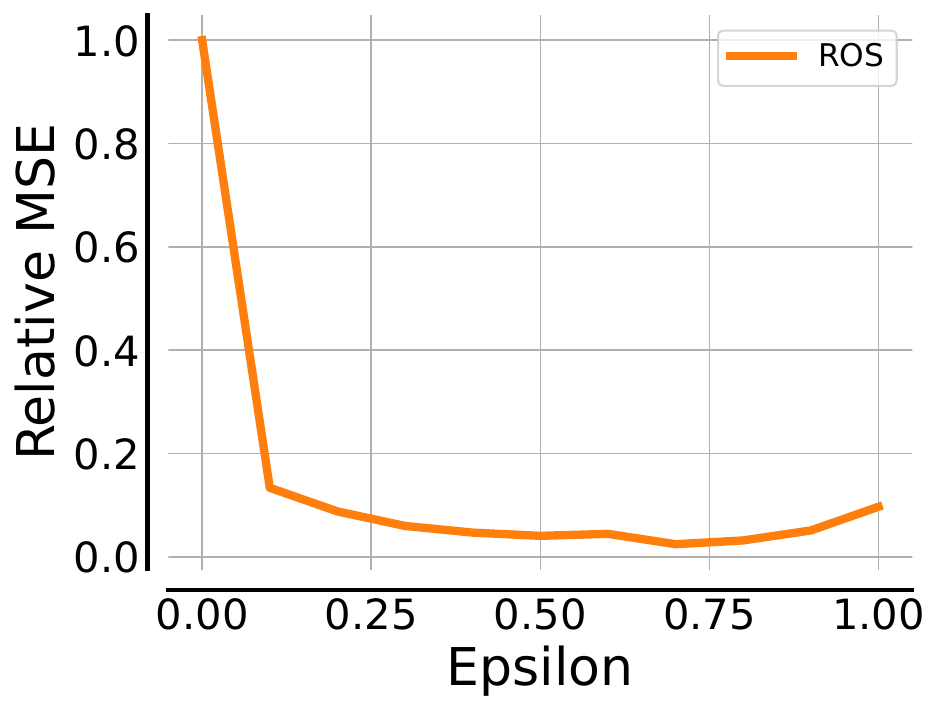}\label{fig:imp_pie_gw}}
\caption{Improvement of \ROS{} compared to \OS{} with different settings in GridWorld. Axes and trials are the same as Figure~\ref{fig:imp_mb}.}
\label{fig:imp_gw}
\end{figure}

In the main paper, we evaluated the relative \MSE{} of \ROS{}  compared to \OS{} in the Bandit environment under different settings of the reward scale and variance and the stochasticity of $\pieval$.
In this appendix, we repeat the same study in the GridWorld environment.
The original GridWorld domain has fixed rewards for each state. We vary these by either 1) multiplying by a fixed factor (referred to as the mean factor) or 2) replacing the deterministic reward with a reward sampled uniformly from $[-f_\text{scale}, f_\text{scale}]$ where $f_\text{scale}$ determines the reward variance.
We then follow \OS{} or \ROS{} ($\alpha=1000$) to collect $1000$ trajectories, and perform Monte Carlo estimation. The relative \MSE{}s between \OS{} and \ROS{} are shown in Figure~\ref{fig:imp_env_gw}.
Results show an identical trend to the Bandit environment:
a larger reward scale increases the amount of improvement because small amounts of sampling error can lead to larger amounts of error;
larger reward variance decreases the amount of improvement because the \MSE{} becomes dominated by variance in the reward.

We also create different evaluation policies using $\epsilon$-greedy policies with $\epsilon$ ranging from $0$ to $1$. 
We then perform the same policy evaluation as above and compute the relative \MSE{}s, shown in Figure~\ref{fig:imp_pie_gw}.
As with the Bandit domain, we see that greater entropy in $\pieval$ generally leads to a wider margin of improvement between \ROS{} and \OS{}.

\section{Mean Squared Error of Policy Evaluation with WRIS and FQE}
\label{app:ope}

\begin{figure}[H]
\centering
\subfigure[Bandit]{\includegraphics[height=7.3em]{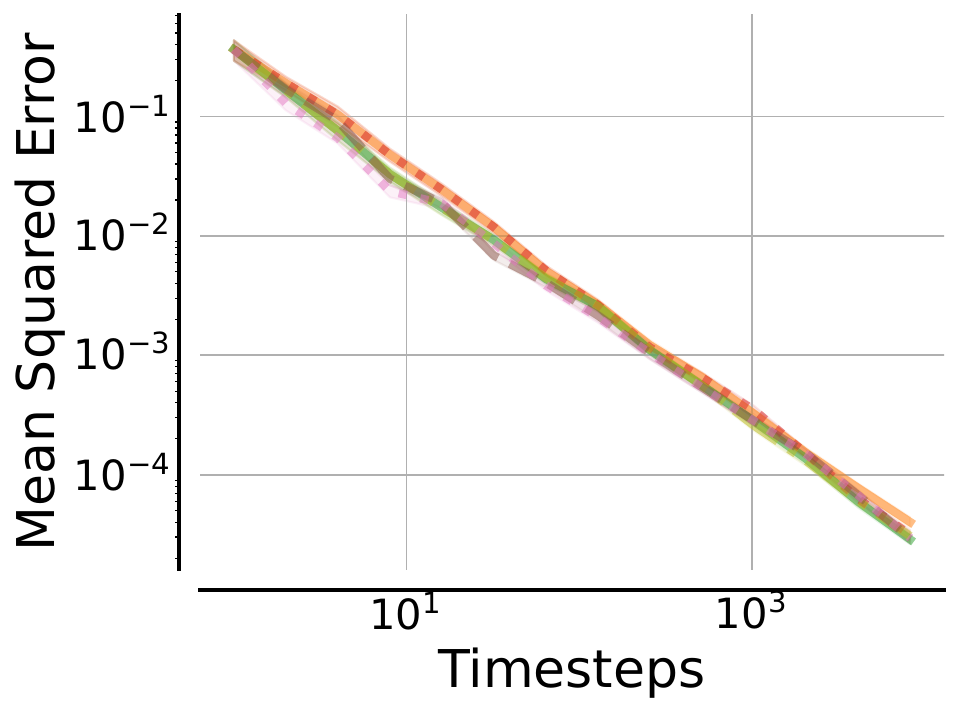}}
\subfigure[GridWorld]{\includegraphics[height=7.3em]{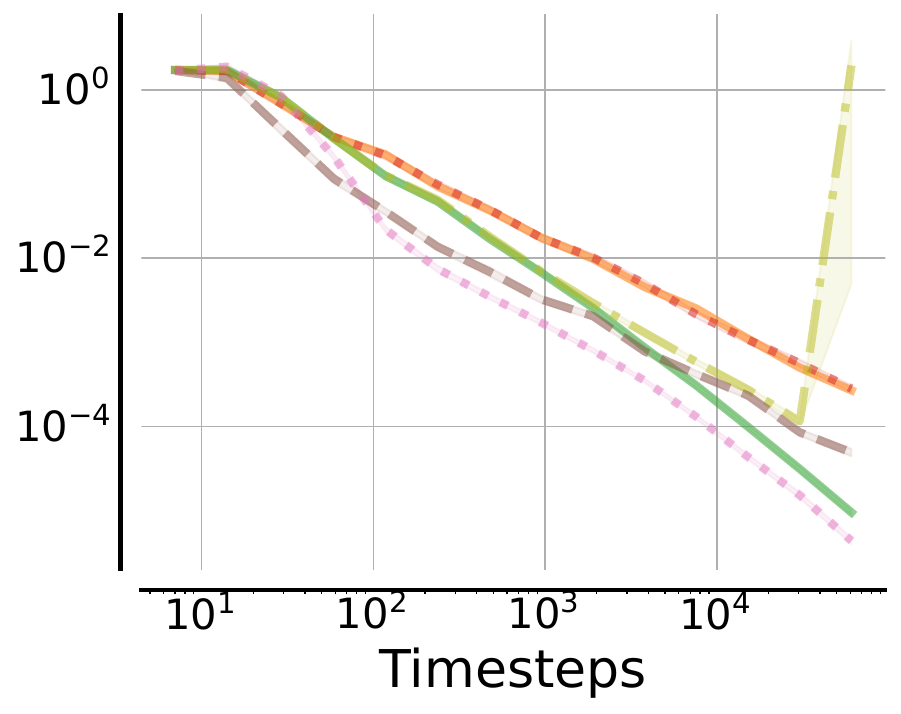}}
\subfigure[CartPole]{\includegraphics[height=7.3em]{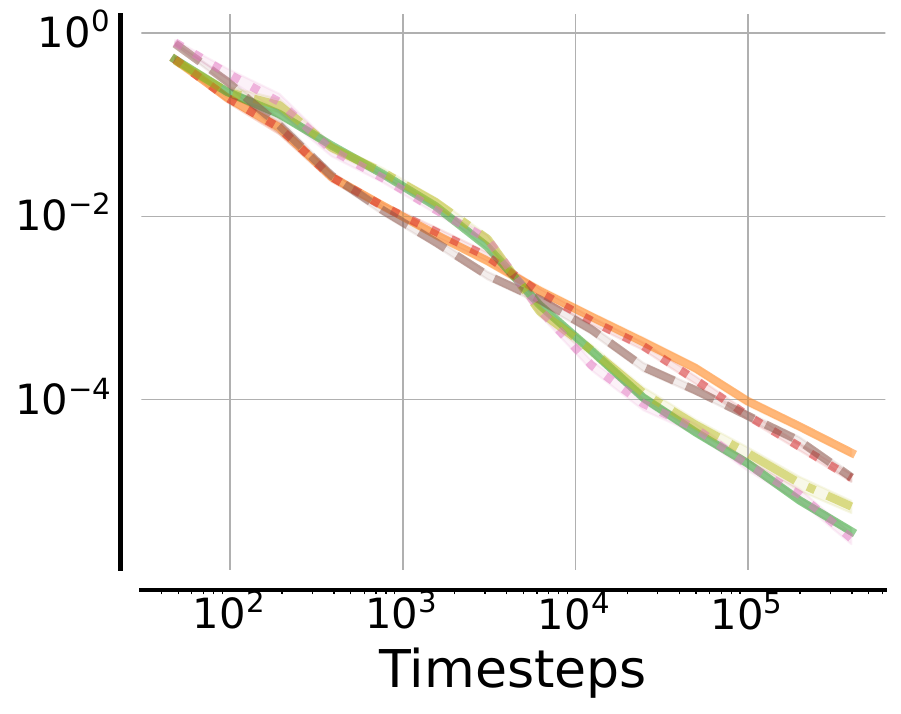}}
\subfigure[ContinuousCartPole]{\includegraphics[height=7.3em]{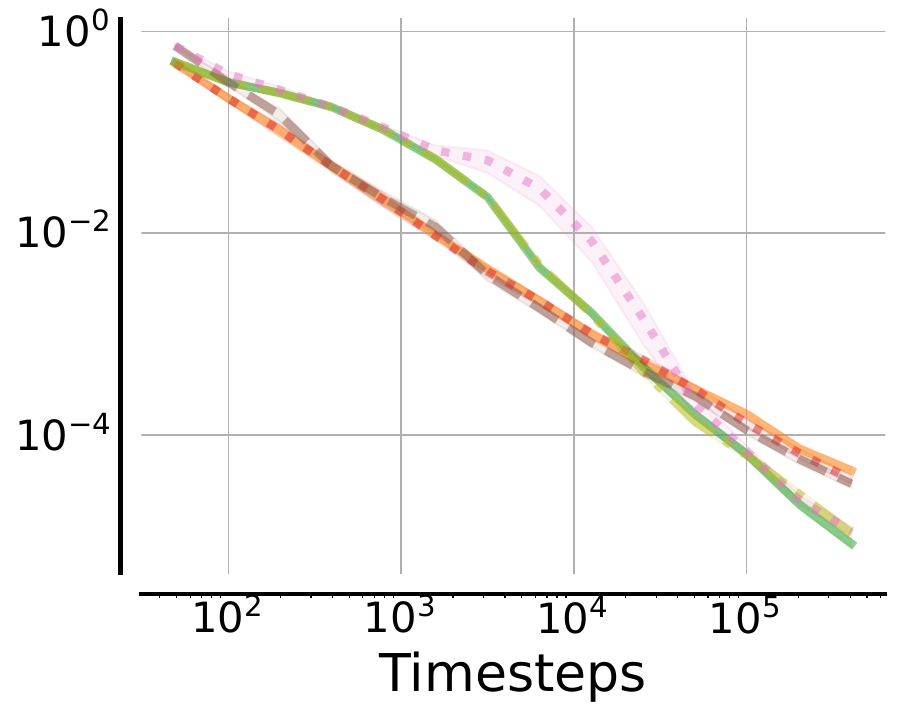}}
\subfigure{\includegraphics[width=0.66\textwidth]{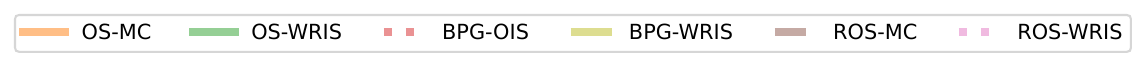}}
\caption{\MSE{} of WRIS \textbf{without initial data}. Steps, axes, trials and intervals are the same as Figure~\ref{fig:mse}.}
\label{fig:mse_wris}
\end{figure}

\begin{figure}[H]
\centering
\subfigure[Bandit]{\includegraphics[height=7.3em]{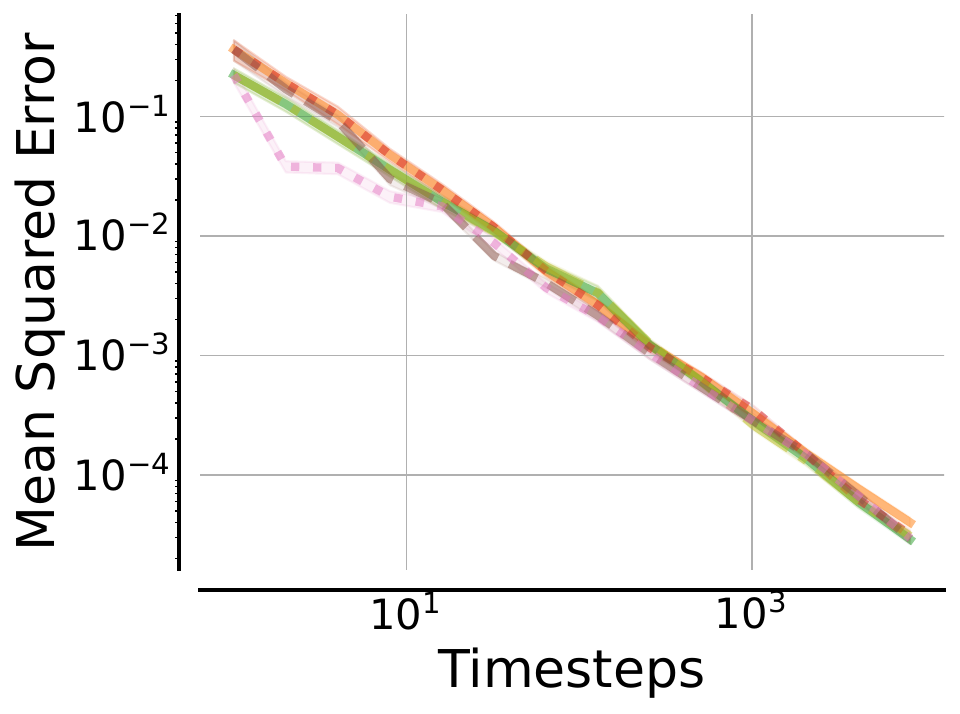}}
\subfigure[GridWorld]{\includegraphics[height=7.3em]{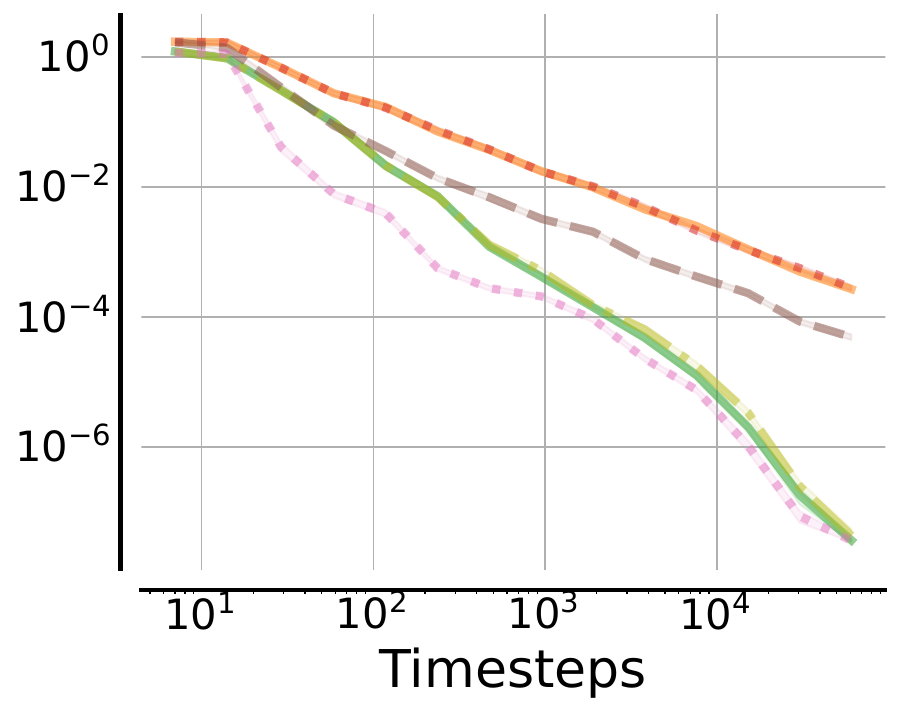}}
\subfigure[CartPole]{\includegraphics[height=7.3em]{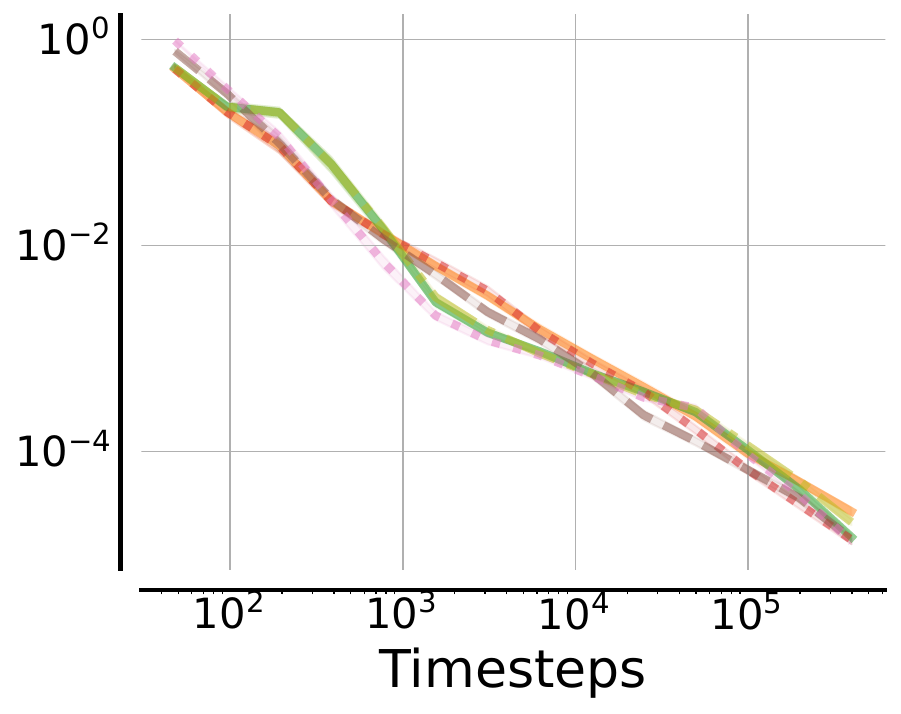}}
\subfigure[ContinuousCartPole]{\includegraphics[height=7.3em]{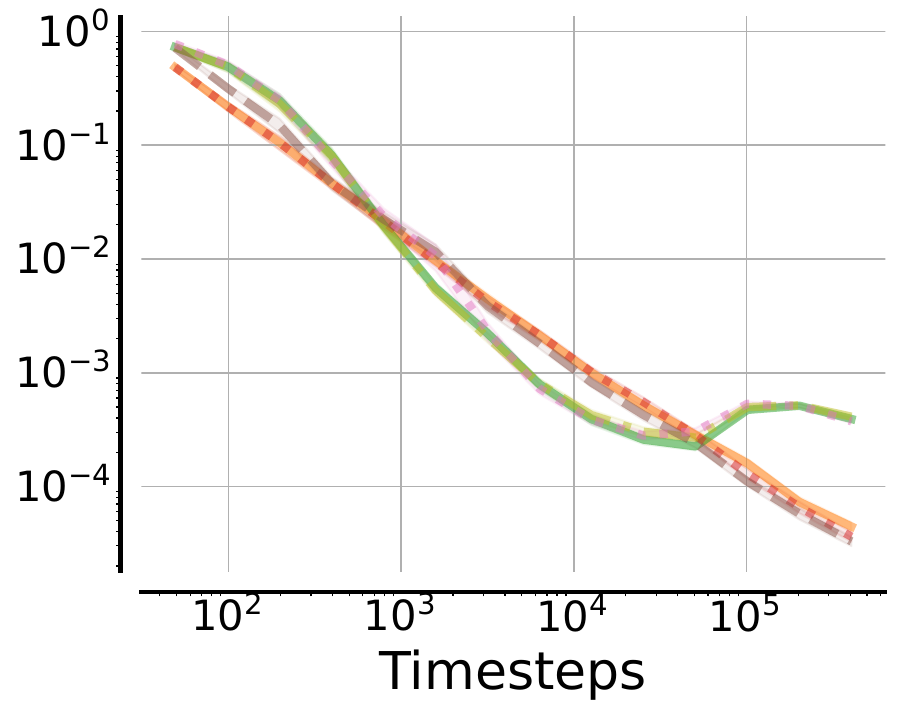}}
\subfigure{\includegraphics[width=0.66\textwidth]{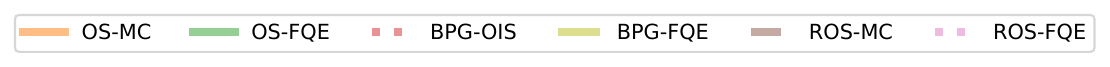}}
\caption{\MSE{} of FQE \textbf{without initial data}. Steps, axes, trials and intervals are the same as Figure~\ref{fig:mse}.}
\label{fig:mse_fqe}
\end{figure}

This appendix presents preliminary results on combining \ROS{} with estimators specifically designed for the off-policy setting instead of the Monte Carlo estimator.
Specifically, we used weighted regression importance sampling (\WRIS{}) \citep{hanna2021importance} and fitted q-evaluation (\FQE{}) \citep{le2019batch}.
We study these estimators in the \textbf{without initial data} setting and show their \MSE{} with varying amounts of data.
Figure \ref{fig:mse_wris} shows the \MSE{} of different data collection methods for \WRIS{} across the different domains.
Figure \ref{fig:mse_fqe} shows the same except with \FQE{} as the estimator.

We observe in tabular domains that \OS{} cannot obtain the same level \MSE{} as \ROS{}, even with the help of \WRIS{}, which is designed to correct sampling error during the value estimation stage. 
This shows the importance of reducing sampling error during the data collection stage. 
The same pattern can also be observed when using \FQE{}. In non-tabular domains, the performances of off-policy evaluation methods with different data collection methods are very similar. However, \WRIS{} usually requires larger sizes of data to make accurate estimation, and can only achieve similar \MSE{} as \ROS{} when collecting a large amount of data. When using \FQE{} in non-tabular domains, data from \ROS{} can generally enable lower \MSE{} than \OS{}, although this improvement is very small.

\end{document}